\documentclass{article}

\usepackage[T1]{fontenc}
\usepackage[utf8]{inputenc}
\usepackage[english]{babel}

\usepackage{amsmath, amssymb, amsfonts, amsthm}
\usepackage{mathtools}
\newcommand{\norm}[1]{\left\lVert #1 \right\rVert}

\usepackage{graphicx}
\usepackage{booktabs}
\usepackage{nicefrac}
\usepackage{microtype}
\usepackage{xcolor}
\usepackage{enumitem}
\usepackage{algorithm}
\usepackage{algorithmic}

\usepackage[numbers,sort&compress]{natbib} % <- si usa \citep, \citet
\usepackage{hyperref}
\usepackage{url}
\usepackage{cleveref}

\newif\ifanon
\anonfalse        % ponga \anontrue para versión anónima

\usepackage{titlesec}
\titlespacing*{\section}{0pt}{0pt}{0pt}
\titlespacing*{\subsection}{0pt}{0pt}{0pt}
\titlespacing*{\subsubsection}{0pt}{0pt}{0pt}

\newtheorem{proposition}{Proposition}
\newtheorem{theorem}{Theorem}

\newtheorem{remark}{Remark}

 \title{Continuous Temporal Learning of Probability Distributions via Neural ODEs with Applications in Continuous Glucose Monitoring Data
}
\author{%
Antonio Álvarez-López\footnotemark[1]\\
Universidad Autónoma de Madrid \\
\texttt{antonio.alvarezl@uam.es}\footnotemark[1]
\and
Marcos Matabuena\footnotemark[1] \\
Harvard University \\
\texttt{mmatabuena@hsph.harvard.edu}\footnote{AÁ and MM share the first position.}
}

\begin{document}

\maketitle

\begin{abstract}
  Modeling the dynamics of probability distributions from time-dependent data samples is a fundamental problem in many fields, including digital health. The goal is to analyze how the distribution of a biomarker, such as glucose, changes over time and how these changes may reflect the progression of chronic diseases such as diabetes. We introduce a probabilistic model based on a Gaussian mixture that captures the evolution of a continuous-time stochastic process. Our approach combines a nonparametric estimate of the distribution, obtained with Maximum Mean Discrepancy (MMD), and a Neural Ordinary Differential Equation (Neural ODE) that governs the temporal evolution of the mixture weights. The model is highly interpretable, detects subtle distribution shifts, and remains computationally efficient. We illustrate the broad utility of our approach in a 26-week clinical trial that treats all continuous glucose monitoring (CGM) time series as the primary outcome. This method enables rigorous longitudinal comparisons between the treatment and control arms and yields characterizations that conventional summary-based clinical trials analytical methods typically do not capture.

\end{abstract}

%\newpage

%\tableofcontents
\newpage

\section{Introduction}

Characterizing the distribution function of a random variable $X$ is a long-standing problem in statistics \cite{silverman2018density} and machine learning \cite{bengio2017deep}. Today, it remains a central challenge in the era of generative AI. In large language models \cite{mesko2023imperative}, for example, accurately modeling the probability distribution of textual data is essential for automatic text and report generation.

Beyond natural language processing, distribution estimation is equally important in clinical settings. In digital health, estimating the distribution of physiological time series data over specific time periods enables the construction of patient representations that capture their underlying physiological processes with high precision \cite{matabuena2021glucodensities, matabuena2023distributional, ghosal2023distributional}. Recent studies show that, when used properly, such representations can reveal clinically relevant patterns that traditional (non-digital) biomarkers do not detect \cite{katta2024interpretable, matabuena2024glucodensity, park2025beyond}.

We address the problem of estimating a CDF \emph{continuously} over time: random samples are observed consecutively, and the goal is to learn how the dynamics of the underlying distribution evolve.  Navely extending classical CDF or kernel density estimators (KDEs) \cite{chacon2018multivariate,papamakarios2021normalizing} with a temporal dimension, or adopting normalized gradient flow techniques, often proves unsatisfactory--KDEs suffer from tuning parameter sensitivity and the curse of dimensionality, while gradient flow methods tend to lack interpretability. Statistical semiparametric approaches provide a partial solution to this problem. For example, time-varying models such as Generalized Additive Models for Location, Scale, and Shape (GAMLSS) alleviate some issues \cite{rigby2005generalized}, yet most implementations are constrained to scalar responses and can impose rigid functional forms. Recent multilevel functional frameworks based on functional-quantile representations \cite{matabuena2024multilevel} offer interpretability but rely on linear dynamics, which limits their ability to capture complex non-linear relationships and multivariate results.

To overcome these limitations, we propose a time--dependent Gaussian mixture model in which a Neural ODE governs the distribution shift smoothly over time, see \cite{chen_neural_2018}.

\subsection*{Notation and Problem Definition}

Let $\mathcal{T}\subset \mathbb{R}_{\ge 0}$ be a continuous-time index set. For each $t \in \mathcal{T}$, the random variable $X(t) \in \mathbb{R}^d$ denotes the outcome of interest at time $t$. Its (cumulative) distribution function is
\begin{equation}\label{eqn:eq1}
  F(x,t)
  \;=\;
  \mathbb{P}\bigl(X(t)\le x\bigr)
  \;=\;
  \int_{\prod_{i=1}^{d}(-\infty,x_i]} f(r,t)\,\mathrm{d}r,
\end{equation}
\noindent for $x=(x_1,\dots,x_d)^{\top} \in \mathbb{R}^d$, where the inequality is taken component-wise for $d>1$. Although our target object is $F(\cdot,t)$, in practice we estimate the time–varying density $f(\cdot,t)$ and recover $F$ via \eqref{eqn:eq1}. We parameterize the density as a \(K\)-component Gaussian mixture
\[
  f(x,t)\;=\;\sum_{s=1}^{K}\alpha_s(t)\,\mathcal{N}\bigl(x\mid m_s,\Sigma_s\bigr),
\]
\noindent where $m_s\in\mathbb{R}^d$ and $\Sigma_s\in\mathbb{R}^{d\times d}$ are the mean vector and positive definite covariance matrix of the $s$-th Gaussian component, and the weight vector $\alpha(t)=[\alpha_1(t),\dots,\alpha_K(t)]$ lies in the probability simplex:
\begin{equation}\label{eq:simp}
\alpha(t)\in\Delta^{K-1}
  \coloneqq \bigl\{\,w\in\mathbb{R}^K:\; w_s\ge0,\ \sum_{s=1}^{K}w_s=1\,\bigr\}.\end{equation}

  \noindent We let $\alpha(t)$ vary continuously in time; its evolution is governed by a Neural ODE, endowing the model with universal approximation power while preserving interpretability.

Our model is motivated by the need to track the evolution of glucose distribution in a longitudinal diabetes trial \cite{battelino2023continuous}, where glucose is continuously recorded using continuous glucose monitoring (CGM). In free-living conditions, individual glucose time series cannot be aligned directly, making the raw temporal stochastic process difficult to compare between participants \cite{ghosal2024multivariate,matabuena2021glucodensities,matabuena2024glucodensity}. Analyzing the time-varying probability distribution therefore provides a more natural and realistic biomarker to characterize the evolution of glucose metabolism \cite{katta2024interpretable,park2025beyond,matabuena2020glucodensities}.

We emphasize that this approach conveys richer information than conventional CGM summary statistics ----such as mean glucose level or time-in-range metrics ----because it captures the full spectrum of low, moderate and high glucose values simultaneously, which is not possible with traditional summaries \cite{matabuena2021glucodensities, katta2024interpretable}.

\subsection*{Contributions}
The main contributions of this paper are:
\begin{enumerate}[left=0pt]
  \item We propose a general framework for modeling the dynamics of multivariate continuous probability distributions in continuous time. Our algorithms scale to large multivariate settings and exploit the universal approximation power of mixture models driven by Neural ODEs.
  
  \item We introduce a novel parameter estimation scheme for mixture models based on a Maximum Mean Discrepancy objective \cite{gretton2012kernel}. This approach avoids the strong structural assumptions and serial correlation issues that affect maximum likelihood methods in this context. Moreover, defining the estimator as a V-statistic \cite{serfling2009approximation} improves both the robustness and the statistical efficiency. To reduce computational cost, we derive a closed-form expression of the loss function.

\item We provide theoretical guarantees for the algorithm prior to the smoothing step. Specifically, we establish statistical consistency and convergence rates by empirical process analysis of the associated estimators \cite{kosorok2007introduction}.  We also analyze the computational complexity of the post-smoothing stage, which explains the acceleration achieved by our implementation, based on massive univariate fitting followed by smoothing, relative to a joint optimization strategy.

%  \item Based on our theoretical developments, we propose a bootstrap-based uncertainty-quantification procedure to construct confidence bands for the estimated distributions.

  \item We demonstrate the practical value of our methods in biomedical applications by analyzing continuous glucose monitoring data from longitudinal diabetes trials \cite{battelino2023continuous}. Our approach offers a novel characterization of the glucose distributional dynamics that existing models cannot capture.
\end{enumerate}
%\subsection{Paper Outline}

%The remainder of the paper is organised as follows.
%Section \ref{sec:2} surveys related work and introduces the prerequisites needed to understand our mathematical framework.
%Section \ref{sec:3} presents the proposed temporal-distribution estimators, their optimisation procedures, and the %accompanying theoretical guarantees.
%Section \ref{sec:4} investigates the finite-sample properties of the estimators through a comprehensive simulation study.
%Section \ref{sec:5} illustrates the broader applicability of the methodology by analysing data from a longitudinal diabetes clinical trial.
%Finally, Section \ref{sec:6} concludes with a discussion of our findings and outlines directions for future research.

\section{Related Work and Preliminaries}
\label{sec:2}
From a modeling perspective, the main novelty of our paper is the integration of mixture models with Neural ODEs to create a highly interpretable estimator for the dynamics of probability distributions generated by temporal stochastic processes. Our estimator relies on \emph{Maximum Mean Discrepancy} as its divergence measure, with the aim of improving performance over classical likelihood-based methods. In what follows, we outline the core components of the framework and place it in the context of existing literature.

\paragraph{Maximum Mean Discrepancy (MMD)}
%--- Maximum Mean Discrepancy definition ------------------------------------

To quantify the discrepancy between two probability distributions
$P$ and $Q$ defined on a common measurable space $\mathcal{X}$, we use \emph{Maximum Mean Discrepancy} (MMD) \cite{gretton2012kernel,muandet2017kernel}.
MMD is a semimetric that equals zero if and only if $P = Q$ provided
the kernel is characteristic \cite{sriperumbudur2011universality,sejdinovic2013equivalence}.

Let $\mathcal{H}$ be a reproducing kernel Hilbert space (RKHS) with a positive definite kernel
$k \colon \mathcal{X}\times\mathcal{X} \to \mathbb{R}$.
Assume that $X \sim P$ and $Y \sim Q$ satisfy
\(
  \mathbb{E}\bigl[k(X,X)\bigr] < \infty
  \ \text{and}\ 
  \mathbb{E}\bigl[k(Y,Y)\bigr] < \infty,
\)
so that all the expectations below are finite.
Define \emph{kernel mean embeddings} \cite{muandet2017kernel} by
\(
  \mu_P \coloneqq \mathbb{E}_{X}\!\bigl[k(\,\cdot\,,X)\bigr] \in~\mathcal{H}\) and \(
  \mu_Q \coloneqq \mathbb{E}_{Y}\!\bigl[k(\,\cdot\,,Y)\bigr] \in \mathcal{H}.
\)
The squared MMD is then
\[
  \operatorname{MMD}^{2}(P,Q)
  =
  \lVert \mu_P - \mu_Q \rVert_{\mathcal{H}}^{2}
  =
  \mathbb{E}_{X,X'}\!\bigl[k(X,X')\bigr]
  +
  \mathbb{E}_{Y,Y'}\!\bigl[k(Y,Y')\bigr]
  -
  2\,\mathbb{E}_{X,Y}\!\bigl[k(X,Y)\bigr]
  \ge 0,
\]
\noindent where $X'$ (resp.\ $Y'$) is an independent copy of $X$ (resp.\ $Y$).

%If the map $P \mapsto \mu_P$ is injective---equivalently, if the kernel $k$ is \emph{characteristic}---then $\operatorname{MMD}^{2}(P,Q)$ is an \emph{omnibus} statistic: it can detect any fixed alternative $P \neq Q$.

\paragraph{Kernel choice.}
In our work, we employ the \emph{Gaussian kernel}, 
\begin{equation}\label{eq:Gkernel}
  k(x,y) = \exp\left(-\frac{\norm{ x - y}^{2}_{2}}{2\sigma^2}\right),
\end{equation}
\noindent where $\sigma > 0$ is a bandwidth parameter that controls smoothness. The Gaussian kernel is \emph{characteristic}; that is, the map $P \mapsto \mu_P$ is injective, so the kernel mean embedding uniquely characterizes the distribution $P$. Consequently, \(\operatorname{MMD}^2(P,Q) = 0\) if and only if \(P = Q\).

%A kernel \(k\) is said to be \emph{characteristic} if the kernel mean embedding \(\mu_P \coloneqq \mathbb{E}_{X \sim P}[k(X,\cdot)]\) uniquely characterizes the distribution \(P\), i.e., \(\mu_P = \mu_Q \Leftrightarrow P = Q\). This ensures that the Maximum Mean Discrepancy satisfies \(\operatorname{MMD}^2(P,Q) = 0\) if and only if \(P = Q\).

In practice, the choice of $\sigma$ strongly impacts the sensitivity of MMD. A standard heuristic sets $\sigma$ to \emph{median} of all pairwise Euclidean distances between samples---known as the median heuristic \cite{garreau2017large}---which balances sensitivity to both local and global structure in the data.
\\\texttt{Intuition.}
MMD measures how far apart $P$ and $Q$ are in the feature space
induced by $k$: it is the squared distance between their kernel mean
embeddings, $\mu_P$ and $\mu_Q$.  In our setting, we use MMD to
compare the empirically evolving time distribution of $X(t)$ with a
parametric (or target) distribution, thereby tracking distributional
changes over time.
%--------------------------------------------------------------------------- 

%--- Maximum Mean Discrepancy definition ------------------------------------

\paragraph{Neural Ordinary Differential Equations (Neural ODEs)}

Introduced in \cite{e_proposal_2017,Haber_2018,chen_neural_2018}, neural ODEs replace discrete layers by the continuous evolution of a hidden state $\dot z(t)=f_\phi\!\bigl(z(t),t\bigr),\;z(t_0)=z_0$, where $f_\phi:\mathbb{R}^{d}\times[t_0,t_1]\to\mathbb{R}^d$ is a learnable vector field usually parameterized by a multilayer perceptron. The value $z(t_1)=\textsf{ODESolve}(z_0,t_0,t_1,f_\phi)$ is then calculated using any standard numerical solver. Gradients with respect to $\phi$ are then obtained by the adjoint method, allowing constant-memory backpropagation \cite{NEURIPS2020_massaroli}.

The continuous framework offers four key benefits:
(i) it adapts computation to local dynamics;
(ii) it is parameter efficient (a single vector field encodes arbitrary depth);
(iii) it induces an (under mild conditions) invertible flow map; and
(iv) it yields smooth latent trajectories. In our model, $\alpha(t)$ denotes the vector of mixture weights, so learning $f_\phi$ captures how the probabilities of the components evolve while respecting the simplex constraint.

%{\color{blue}Some references:}

%For irregularly sampled time series, ODE-RNN architectures \cite{NEURIPS2019_rubanova} combine traditional recurrent updates (e.g., GRUs) at discrete observations with Neural ODE-based hidden state evolution between observations, enabling interpolation and accurate forecasting regardless of sampling irregularities. Similarly, GRU-ODE-Bayes \cite{NEURIPS2019_debrouwer} integrates continuous-time GRU dynamics for hidden-state evolution between observations and Bayesian updates at observation times. This model effectively encodes uncertainty and continuity priors, excelling in medical and climate data scenarios characterized by sparse, irregular measurements. Controlled Neural Differential Equations (Neural CDEs) \cite{kidger2020neural} generalize Neural ODEs by explicitly conditioning dynamics on continuous input streams. Neural CDEs model complex temporal dependencies robustly, offering significant improvements in tasks with irregular sampling and long-range correlations. \cite{NEURIPS2019_njumpsdes} adds an stochastic process term that models discrete events, resulting in a piecewise-continuous latent trajectory.
Neural ODEs have been applied to distribution learning for generative modeling \cite{chen_neural_2018}, and recent work has established approximation and stability guarantees \cite{Marzouk2024,alvarezlopez2025}. Their framework has also been extended to forecasting by modeling latent states as continuous-time trajectories \cite{kidger2020neural,NEURIPS2019_rubanova,NEURIPS2019_njumpsdes}, a strategy that has been shown to be especially effective in biomedical applications \cite{qian2021integrating}.

\paragraph{Distributional Data Analysis}
Distributional data analysis \cite{ghosal2023distributional,szabo2016learning} is an emerging field that treats or aggregates probability distributions as random objects for unsupervised and supervised learning, for example, to predict clinical outcomes \cite{matabuena2020glucodensities}. Biomedical applications are its most prominent use case. In digital health, measurements collected by continuous glucose monitoring devices, accelerometers, or medical imaging modalities such as functional magnetic resonance imaging (fMRI) are now commonly represented through their empirical distributions, which serve as latent descriptions of the underlying physiological processes \cite{ghosal2025distributional,ghosal2024multivariate,matabuena2024glucodensity,matabuena2022physical}. In recent years, several regression frameworks have been proposed to represent predictors, responses, or both as probability density functions \cite{ghosal2025distributional,ghosal2024multivariate}. Another research strand embeds probability distributions as random objects in metric spaces, for which dedicated statistical procedures have been developed (see, e.g. \cite{lugosi2024uncertainty}). Despite this growing body of work, to the best of our knowledge, there is still no comprehensive analytical framework that handles moderate- to high-dimensional distributions both flexibly and robustly. The methods introduced here aim to fill that gap and provide a practical toolkit for distributional data analysis.
\section{Mathematical models}
\label{sec:3}

In this section, we introduce our approach to learning probability distributions that evolve over time. %Let \(f(\cdot,t)\), with \(t\in\mathcal{T}\subset\mathbb{R}\), be a family of {\color{red}absolutely continuous} density functions on \(\mathcal{X}=\mathbb{R}^{d}\) that vary smoothly with \(t\).Define
Let \(\mathcal{T}\subset\mathbb{R}\) and consider the family of probability density functions defined in \(\mathcal{X}=\mathbb{R}^{d}\) that evolve in \(\mathcal{T}\), 
\[
  \mathcal{D}
  \;\coloneqq\;
  \bigl\{\,f\colon (x,t)\in\mathcal{X}\times\mathcal{T}\longmapsto f(x,t)\in\mathbb{R}_{\ge0}\;\big|\;
 \int_{\mathcal{X}}f(x,t)\mathrm{d}x=1\; \forall\,t\in\mathcal{T}\bigr\},
\]

\noindent equipped with the norm
\[
  \|f\|_{L^1,\mathcal{T}}
  \coloneqq \sup_{t\in\mathcal{T}}\|f(\cdot,t)\|_{L^1(\mathcal{X})}.
\]

\noindent For practical analytical purposes, we restrict our attention to a subset of
$\mathcal{D}$ consisting of elements that vary smoothly with $t$, such as differentiable density functions.

 A classical result (Wiener–Tauberian Theorem; \cite{wiener1932}) guarantees that the class of finite Gaussian mixtures,
\[
  \mathcal{M}
  \;=\;
  \Bigl\{
    x \mapsto \sum_{s=1}^K \alpha_s\,\mathcal{N}(x, m_s, \Sigma_s)
    \;\mid\;
    K\in\mathbb{N},\ \alpha_s\ge0,\ \sum^{K}_{s=1}\alpha_s=1
  \Bigr\},
\]
\noindent is dense in \(\bigl(L^1(\mathcal{X}),\|\cdot\|_{1}\bigr)\).  Consequently, for any  \(\varepsilon>0\) and \(f\in\mathcal{D}\)   with some additional mild continuity/tightness assumptions on $t\mapsto f(\cdot,t)$ (see \cref{prop:uniform_shared_dictionary}), there exists
\[
  g(x,t)
  = \sum_{s=1}^K \alpha_s(t)\,\mathcal{N}(x\mid m_s, \Sigma_s)
  \;\in\;\mathcal{M},
\]
with fixed means \(\{m_s\}^{K}_{s=1}\) and covariances \(\{\Sigma_s\}^{K}_{s=1}\) but time‑varying weights \(\{\alpha_s(t)\}^{K}_{s=1}\), such that
\[
  \sup_{t\in\mathcal{T}}
  \bigl\|f(\cdot,t) - g(\cdot,t)\bigr\|_{L^1(\mathcal{X})}
  < \varepsilon.
\]
% If you wish to call this out as a “main theorem,” include it here with a theorem environment.
Only the local mixing weights \(\alpha_s(t)\) vary in \(t\), whereas \(m_s\) and \(\Sigma_s\) are shared globally.  However, in practice we fix a moderate $K$ for interpretability and select it by the application (cf.  \cref{sec:5}). 
Let $$\theta(t)= \bigl(\alpha_1(t),\dots,\alpha_K(t),\,
           m_1,\dots,m_K,\,
           \Sigma_1,\dots,\Sigma_K\bigr)\in\Delta^{K-1}\times\mathbb{R}^{Kd}\times\mathbb{R}^{\frac{Kd(d+1)}{2}}\subset\mathbb{R}^p$$
      and thus $p = K\Bigl(1 + d + \tfrac{d(d+1)}2\Bigr)$. Assume the evolution of \(\alpha(t)=(\alpha_1(t),\dots,\alpha_K(t))\) follows the Neural ODE
\[
  \dot\alpha(t) = f_\phi\bigl(\alpha(t),t\bigr),
  \qquad
  \alpha(0)=\alpha_0,
\]
where \(f_\phi:\Delta^{K-1}\times[0,1]\to\Delta^{K-1}\) is globally Lipschitz in \(\alpha\).

Denote the set of all valid parameters by
\[
  \Theta
  = \bigl\{(\alpha_s,m_s,\Sigma_s)_{s=1}^K\;\big|\;
      \alpha_s\in\Delta^{K-1},\;
      m_s\in\mathbb{R}^d,\ \Sigma_s\succ0
  \bigr\}\subset\mathbb{R}^p.
\]
% If you need \(\Theta\) compact for your ODE theory, impose bounds \(\|m_s\|\le M\) and eigenvalues of \(\Sigma_s\) in \([\lambda_{\min},\lambda_{\max}]\).

From an empirical standpoint, we observe data at the discrete time grid \(\tau=\{t_0,\ldots,t_m\}\subset\mathcal{T}\) (\(|\tau|=m+1\)).  For each \(t_i\in\tau\), we record \(n_{i}\ge 1\) observations \(X_{t_i,1},\ldots,X_{t_i,n_{i}}\) drawn from the target distribution \(F(\cdot,t_i)\).  Two regimes are relevant: (i) a \emph{cross‐sectional} setting in which the full collection 
$\left\{ X_{t_i, j} \;:\;0 \leq i \leq m,\; 1 \leq j \leq n_i\right\}$
is i.i.d.; (ii) a \emph{longitudinal/time‐series} setting in which temporal dependence may exist within or across the blocks indexed by \(t_i\).  In the second regime, maximum--likelihood estimation can be cumbersome and often lacks robustness to model misspecification \cite{alquier2024mmd}. This motivates the minimum--MMD optimization approach introduced here.

\subsubsection*{Discrete-time MMD fitting}

For each time $t_i\in \tau,$ consider the empirical measure

\[
F_{t_i,n_i} = \frac{1}{n_{i}}\sum_{j=1}^{n_{i}} \delta_{X_{t_i,j}},\hspace{1cm} i=0,\dots,m
\]

%\footnote{\color{red}por qué $P_{t_i}$? pero si acabamos de introducir $F_{t_i,n_i}$. Creo que no está muy claro}

\noindent where $\delta_{X_{t_i,j}}$ denotes the Dirac measure at the observation $X_{t_i,j}$. The discrete-time optimization minimizes the squared MMD between $F_{t_i}$   and the kernel mean embedding of the Gaussian mixture density
$$
f_i(x) =\sum_{s=1}^{K} w_{s}\,\mathcal{N}(x \mid m_s, \Sigma_s).
$$

\noindent At each $t_i\in\tau$, we use a Gaussian kernel $k_i$ such as \eqref{eq:Gkernel}, with 
 $\sigma_i^2 \approx (\underset{j\neq k}{\operatorname{median}}\|X_{t_i,j}-X_{t_i,k}\|)^2$ the median heuristic separately at each $t_i$. Thus,
\[
 \mathrm{MMD}^2(F_{t_i,n_i},Q) 
=\sum_{s=1}^K\sum_{r=1}^K w_{s}w_{r}\, I_{i,s,r}
-\frac{2}{n_i}\sum_{s=1}^K\sum_{j=1}^{n_i}w_{s}\,J_{i,s,j}
+\frac{1}{n_i^2}\sum_{j=1}^{n_i}\sum_{\ell=1}^{n_i} k_i(X_{t_i,j}, X_{t_i,\ell}).
\]

The first two terms admit closed-form expressions:
\begin{align*}
 I_{i,s,r} &= \frac{(\sigma_i^2)^{d/2}}{\sqrt{\det(\Sigma_s + \Sigma_r+\sigma_i^2 \mathsf{Id})}}
\exp\!\Bigl(-\frac{1}{2} (m_s-m_r)^\top (\Sigma_s+\Sigma_r+\sigma_i^2 \mathsf{Id})^{-1} (m_s-m_r)\Bigr),\\
J_{i,s,j} &= \frac{(\sigma_i^2)^{d/2}}{\sqrt{\det(\Sigma_s+\sigma_i^2 \mathsf{Id})}}
\exp\!\Bigl(-\frac{1}{2} (X_{t_i,j}-m_s)^\top (\Sigma_s+\sigma_i^2 \mathsf{Id})^{-1}(X_{t_i,j}-m_s)\Bigr),
\end{align*}
\noindent while the last term is computed directly from the data. This MMD objective is minimized as follows.

\textbf{Initialization.} Run $k-$means clustering \cite{jain2010data}
on $\bigcup_{i,j} X_{t_i,j}$ to obtain the initial means $\{m_s\}_{s=1}^{K}$ as cluster centers; the initial covariances $\{\Sigma_s\}_{s=1}^{K}$  as the empirical covariances of each cluster; and the initial weights as $\{c_s/n\}_{s=1}^K$, where $c_s$ is the number of points in the group $s$ and $n=\sum_{i=0}^m n_i$.

\textbf{Local update (E-step).} For each $t_i$, update the weight vectors $\alpha_{i} = [\alpha_1(t_i), \dots, \alpha_K(t_i)]^\top\in\mathbb{R}^{K}$ as
\begin{equation}
    \alpha_{i} = \underset{w \in \Delta^{K-1}}{\operatorname{argmin}} \;\{ w^\top I_i \, w - 2\, w^\top J_i +  \sum_{s=1}^{K} \lambda _s\,w_s^2\},
    \label{eq:local_qp}
\end{equation}
\noindent where $I_i=(I_{i,s,r})_{s,r}\in \mathbb{R}^{K \times K}$;  $J_i = (\sum_{j=1}^nJ_{i,s,j})_{s}\in \mathbb{R}^{K}$; and 
   $\lambda=(\lambda_s)_{s=1}^K\in\mathbb{R}^K$ are individual hyperparameters of ridge penalty.\\
\textbf{Global update (M-step).} Update $m_s$ and $\Sigma_s$ iteratively via (Adam) gradient descent on the objective.\\
\textbf{Ridge penalty.} We note that the loss function in \eqref{eq:local_qp} was equipped with an individual ridge penalty for each model weight, motivated by the favorable properties this type of penalization has shown in sparse variable selection \cite{tucker2023variable, bertsimas2020sparse}.

%{\color{red}Esto me suena muy raro. La penalización que promociona sparsity siempre fue $\ell^1$. No me tiene mucho sentido que el motivo de introducir ridge regularization sea la sparsity, para eso deberíamos penalizar $\ell^1$. El motivo de usar $\ell^2$ normalmente es to promote conditioning and stability}

%{\color{red}Propongo:}
%\textbf{Ridge penalty.}
%We include an \(\ell^2\) (ridge) penalty \(\sum_s\lambda_s w_s^2\) in \eqref{eq:local_qp} to improve numerical conditioning of the quadratic program, stabilize the solution when components are nearly collinear in feature space, and mitigate overfitting.

\subsubsection*{Continuous-Time Weight Evolution via Neural ODEs}

Once all discrete-time local weights $\alpha_{i}$ have been fitted, the next step is to learn a continuous-time model for their evolution. We posit a Neural ODE 
\begin{equation}\label{eq:node}
   \frac{d\alpha(t)}{dt} \;=\; f_{\phi}\bigl(\alpha(t),t\bigr)\hspace{1cm}
   t \in \mathcal{T},  
\end{equation}
which produces a trajectory $\alpha(t)\in\mathbb{R}^K$. We parameterize the field $f_\phi$ with a multilayer perceptron.

%\footnote{\color{red}The exact choices for the architecture (layers, width, activation, integration interval) and the numerical solver (tolerances, discretization parameters) are provided in \cref{tab:exp-hyperparams-compact}}. 

To ensure that $\alpha(t)\in\Delta^{K-1}$ for all $t\in\mathcal{T}$, i.e. satisfying $\sum_{s=1}^K \alpha_s(t) = 1$ and $\alpha_s(t) \ge 0$, 
 we project 
\begin{equation*}
 \alpha(t)\longmapsto\alpha(t) = \frac{\alpha(t)}{\sum_{s=1}^K \alpha_s(t)}.   
\end{equation*}
The ODE parameters $\phi$ are then optimized by minimizing the discrepancy between the integrated trajectory and the previously fitted weights $\alpha_i$: \begin{equation}\label{eq:ode_loss} \mathcal{L}_{\text{NODE}}(\phi) = \sum_{i=0}^{m}\norm{\alpha(t_i;\phi) - \alpha_i}_{2}^2 + \nu\|\phi\|_2^2, \end{equation} where $\alpha(t_i)$ denotes the solution of \eqref{eq:node} evaluated at time $t_i$, and $\nu\ge0$ is a ridge hyperparameter.

             % \begin{itemize}
                %  \item \emph{Heuristic Update:} Update via weighted moment matching based on the aggregated local weights across all time steps.
                 % \item \emph{Gradient Update:} Compute gradients of the MMD objective with respect to the parameters and apply a gradient descent update.
             % \end{itemize}

\paragraph{Permutation symmetry.}
The permutation symmetry of Gaussian mixtures is solved in our workflow by the MMD fitting step. Because \((m_s,\Sigma_s)\) are shared over time and kept fixed after the global fit, component labels are anchored: ``component \(s\)'' denotes at every time (and, in the clinical study, across participants) the one centered at \(m_s\) with shape \(\Sigma_s\). The NODE stage only evolves \(\alpha_s(t)\), so trajectories cannot exchange labels, removing the permutation ambiguity without additional constraints.

%\begin{remark}[Identifiability up to permutation]\label{rem:permute}
%Parameters are unique only up to relabeling; $|\hat\theta-\theta|$ is minimized over permutations of components.
%\end{remark}

\paragraph{Why MMD?}
(i) With a Gaussian kernel, the MMD between an empirical sample and a Gaussian mixture admits closed-form terms, delivering numerically stable and fast updates; (ii) the RKHS geometry gives well-posed objectives under characteristic kernels; and (iii) empirical studies show robustness to misspecification and adverse settings \cite{CheriefAbdellatifAlquier2022,alquier2024mmd,gao2021mmd,alquier2023copulammd}. These properties make MMD a pragmatic choice for time-resolved density fitting.

\paragraph{Why neural ODEs?}
Our object of interest is the continuous-time distribution of CGM in free-living environments, where the measurements are irregular and not aligned between participants. Modeling the weight trajectories $\alpha(t)$ with an ODE yields a smooth and subject-invariant evolution that mitigates sensor noise and avoids grid alignment. Discrete sequence models \cite{wang2024timemixer,wu2023timesnet} are effective in forecasting raw traces but do not directly furnish the continuous-time distributional dynamics central to our aims.

\subsection{Theory}\label{sec:the}
We begin by introducing theoretical guarantees for our estimator of the model parameters based on the MMD before the smoothing step. Proofs are present in the last section.

First, we establish the universality of our model.

\begin{proposition}[Universality]
\label{prop:uniform_shared_dictionary}
Let $\{f(\cdot,t)\}_{t\in\mathcal T}\subset L^1(\mathbb R^d)$ be probability densities. Assume:
\begin{enumerate}
\item For every $\eta>0$ there exists $R<\infty$ such that $\int_{\|x\|>R} f(x,t)\,\mathrm{d}x<\eta$ for all $t\in\mathcal T$;
\item  $\displaystyle \lim_{|h|\to0}\,\sup_{t\in\mathcal T}\|f(\cdot+h,t)-f(\cdot,t)\|_{L^1}=0$.
\end{enumerate}
Then, for every $\varepsilon>0$ there exist $\sigma^2>0$ and a finite set of centres $\{\mu_s\}_{s=1}^K\subset\mathbb R^d$ such that, for each $t\in\mathcal T$, one can choose weights $\alpha(t)=(\alpha_1(t),\ldots,\alpha_K(t))\in\Delta^{K-1}$ with
\[
  \bigg\|\,f(\cdot,t)\;-\;\sum_{s=1}^K \alpha_s(t)\,\varphi_{\sigma}(\,\cdot-\mu_s)\,\bigg\|_{L^1}
  \;<\;\varepsilon,
\]
where $\varphi_{\sigma}(x)=(2\pi\sigma^2)^{-d/2}\exp\!\big(-\|x\|^2/(2\sigma^2)\big)$. If, in addition, $t\mapsto f(\cdot,t)$ is uniformly $L^1$–continuous, the map $t\mapsto\alpha(t)$ can be chosen continuous.
\end{proposition}

Now we derive uniform convergence rates for the initial nonparametric fitting of the MMD model.

\begin{theorem}[Uniform MMD Convergence Rate]
\label{thm:MMD_rate}
Let \(\{F^{\theta}_t\colon t \in \mathcal{T}\}\) be a time-indexed family of mixture distributions on \(\mathbb{R}^d\), each with \(K\) components parameterized by \(\theta(t) \in \Theta \subset \mathbb{R}^p\). Assume that at each time \(t\), we observe an i.i.d. sample of size \(n_t\), and define \(n_* = \min_{t \in \mathcal{T}} n_t\). Let \(F^{\theta}_{t,n_t}\) denote the empirical estimator  at time \(t\) based on the \(n_t\) samples of \(\{F^{\theta}_t\colon t \in \mathcal{T}\}\), and define the minimum component separation at time \(t\) as
\[
  \Delta(\theta(t)) \;=\; \min_{j \ne k} \|\theta_j(t) - \theta_k(t)\|.
\]
Then:
\begin{enumerate}
  \item \emph{Regular regime:} If there exists \(\Delta > 0\) such that
  \(\inf_{t \in \mathcal{T}} \Delta(\theta(t)) \ge \Delta\), then
  \[
    \sup_{t \in \mathcal{T}} \mathrm{MMD}\bigl(F^{\theta}_{t,n_t},\,F^{\theta}_t\bigr)
    = O_p\!\bigl(\Delta^{-2} n_*^{-1/2} \sqrt{\log n_*} \bigr).
  \]

  \item \emph{Singular regime:} If there exists \(t \in \mathcal{T}\) such that \(\Delta(\theta(t)) = 0\),
  then
  \[
    \sup_{t \in \mathcal{T}} \mathrm{MMD}\bigl(F^{\theta}_{t,n_t},\,F^{\theta}_t\bigr)
    = O_p\!\bigl(n_*^{-1/4} \sqrt{\log n_*} \bigr).
  \]
\end{enumerate}

In both regimes, the convergence rates are optimal up to logarithmic factors. An additional multiplicative constant of the order \(\sqrt{p}\) may appear depending on the dimension \(p\) of the parameter space.
\end{theorem}

 Finally, we compare the computational cost of a ``massive univariate'' fitting procedure---where we independently fit at each time point \(t_i \in \tau\)---with that of a global strategy that simultaneously optimizes the model at \(m+1\) time points.  These results justify using the massive univariate fitting approach.

\begin{proposition} [Comparison with Per-Time MMD Evaluation]\label{prop:comparison}
Let \( \bar{n} = \max_i n_{t_i} \), and assume all \( n_{t_i} \approx \bar{n} \), and that \( n = \bar{n} \) in the joint formulation. Then:

\begin{itemize}
  \item \textbf{Per-time cost:}
  \[
  \mathcal{O}\left((m+1)\left[ K^2 d^3 + \bar{n} K d^2 + \bar{n}^2 d \right] \right)
  \]
  \item \textbf{Joint cost:}
  \[
  \mathcal{O}\left(K^2 d^3 (m+1)^3 + \bar{n} K d^2 (m+1)^2 + \bar{n}^2 d (m+1) \right)
  \]
\end{itemize}

\end{proposition}

\begin{remark}[Scope of statistical guarantees]
Our finite-sample rates quantify the error of the population minimizer and its empirical MMD counterpart in Step~1, under characteristic kernels and standard regularity. The alternating optimization and the subsequent ODE smoothing (Step~2) are used to enforce temporal coherence and interpretability; a full convergence analysis of EM-like routines and post-smoothing is beyond our scope. Thus, our rates refer to the oracle MMD estimator; the practical two-step procedure is designed so that smoothing acts as a stable refinement of a consistent first-stage estimate.
\end{remark}

\paragraph{Parameter estimation via MMD.}
Let $Q_\theta$ denote a $K$-component Gaussian mixture with parameter $\theta=(\alpha,m,\Sigma)$, and let
\[
  \theta_t^\star \in \operatorname{argmin}_{\theta\in\Theta}
  \mathcal R_t(\theta)
  \;\coloneqq\; \mathrm{MMD}^2\bigl(P_t,Q_\theta\bigr),
  \qquad
  \hat\theta_t \in \operatorname{argmin}_{\theta\in\Theta}
  \widehat{\mathcal R}_t(\theta)
  \;\coloneqq\; \mathrm{MMD}^2\bigl(\widehat P_t,Q_\theta\bigr).
\]
\noindent We assume $k$ is characteristic and the model is well-specified ($P_t=Q_{\theta_t^\star}$) or misspecified with a unique population minimizer.

\begin{theorem}[Rates for the MMD estimator of mixture parameters]\label{thm:mmd_param_rates}
Fix $t$ and suppose $\theta_t^\star$ is identifiable up to label permutation, with $Q_\theta$ twice continuously differentiable in $\theta$ and the Jacobian $J_t \coloneqq \nabla_\theta \mu_{Q_\theta}\big|_{\theta=\theta_t^\star}$ having full column rank (\emph{regular regime}). Then
\[
  \|\hat\theta_t-\theta_t^\star\| \;=\; O_p\!\big(n_t^{-1/2}\big),
\]
and $\sqrt{n_t}\,(\hat\theta_t-\theta_t^\star)$   asymptotically normal after an appropriate relabeling of components.

If, instead, $\theta_t^\star$ is \emph{singular} (e.g., at a component collision where $J_t$ loses rank), and the first nonvanishing term in the local expansion of $\theta\mapsto \mu_{Q_\theta}$ around $\theta_t^\star$ is quadratic, then
\[
  \|\hat\theta_t-\theta_t^\star\| \;=\; O_p\!\big(n_t^{-1/4}\big).
\]
The $n^{-1/4}$ rate is minimax optimal in the singularities of classical finite mixtures and extends to the MMD objective under the stated smoothness structure. Uniform statements on a finite grid $\tau$ follow by taking $n_\star=\min_i n_{t_i}$ and applying a union bound.
\end{theorem}

\begin{proof}[Proof sketch]
View $\widehat{\mathcal R}_t(\theta)$ as an M-estimation objective in a Hilbert space. In the regular regime, a second-order Taylor expansion of $\mathcal R_t$ at $\theta_t^\star$ yields strong local convexity with curvature $J_t^\top J_t$, while $\widehat{\mathcal R}_t-\mathcal R_t$ fluctuates at $n_t^{-1/2}$. Standard M-estimation theory then gives $\sqrt{n_t}$-consistency and asymptotic normality. At singular points, $J_t$ loses rank and the quadratic (Hessian) term may vanish in directions associated with collisions; the first nonzero term becomes quartic in the natural reparameterization of mixtures, leading to $n_t^{-1/4}$ rates (cf. classical finite-mixture singularity analyses). The same mechanism carries over because the MMD objective is a smooth functional of the mixture moments entering the Gaussian kernel integrals.
\end{proof}

\section{Simulations}
\label{sec:4}

% --- Simulation setup -------------------------------------------------
To benchmark finite-sample performance against state-of-the-art methods, we conduct a simulation study. Unlike many competitors, our approach explicitly prioritizes interpretability through time-varying functions
$\alpha_s(\cdot)$ for $s=1,\dots,K$. The results show that this emphasis on interpretability does not compromise accuracy: our estimator closely approximates the underlying distribution function.

Let \(\mathcal{T} = [0, 1]\). We fix a target population distribution $f$ at each \(t \in \mathcal{T} \)  given by a mixture of three Gaussian components whose means and common variance evolve linearly in time:
\begin{equation*}
  f(x,t)
  = \frac{1}{3}\sum_{s=1}^{3}
      \mathcal{N}\!\bigl(x;\,m_s(t),\,\sigma^{2}(t)\,\mathsf{Id}\bigr),
  \label{eq:mixture}
\end{equation*}
\noindent where %the time-dependent variance is
\begin{equation*}
  \sigma^{2}(t) = 1 + t,
  \qquad t \in \mathcal{T},
  \label{eq:variance}
\end{equation*}
\noindent and %the component means follow
\begin{equation}
  m_1(t) = -2 + 20\cdot t, 
  \qquad
 m_2(t) = 16\cdot t, 
  \qquad
  m_3(t) = 5 + 6\cdot t.
  \label{eq:means}
\end{equation}

\noindent For dimensions \(d \ge 2\) each mean vector \(m_s(t)\) has identical coordinates given by
\eqref{eq:means}, and the covariance operator remains isotropic as has identical coordinates given by the expressions above, and the covariance remains isotropic 
\(\Sigma(t) = \sigma^{2}(t)\,\mathsf{Id}\).
This design enables the simultaneous analysis of both multimodal and unimodal regimes over time, capturing different distributional shapes within a single simulation study. For each setting we generate \(B = 100\) independent replicates.  Performance is evaluated at
\(m = 11\) time points \(t_i \in \mathcal{T} = [0,1]\)
(equally spaced unless stated otherwise).
At each time \(t_i\) we draw \(n_t\) observations, with the sample size
chosen from
\(
n_t \in \{20,\,50,\,100,\,200,\,300, 500\}.
\)

\subsection{Competitors}

We compare our estimator with three competitor models in a low dimensional scenario $d=1$ and a large dimensional scenario $d=10$. The methods are the GAMLSS model \cite{rigby2005generalized} (used only for dimension $d=1$), a conditional-in-time KDE \cite{tsybakov2009nonparametric, chacon2018multivariate}, and a Masked Autoregressive Flow \cite{papamakarios2021normalizing}.

Although these approaches can account for changes over time, they generally do not introduce a single, interpretable function for continuous ``shifts'' in the distribution (like our $\alpha(t)$) and may be less transparent for complex, high-dimensional data. Brief overviews of each competing method are given below; all implementations use the same early-stopping rule and hyperparameter search.

\paragraph{Normalizing flows}
We study normalizing flow as an autoregressive time-series model, following \cite{NIPS2017_6c1da886}.  
Let $x=(x_1,\dots,x_I)\in\mathbb{R}^I$ and factorize its joint density as
\[
F(x)=\prod_{i=1}^{I} F(x_i\,|\,x_{<i}),
\qquad x_{<i}\equiv(x_1,\dots,x_{i-1}),
\]
\noindent where each conditional is Gaussian,
\[
F(x_i\,|\,x_{<i})=\mathcal{N}\!\bigl(x_i\mid\mu_i,\exp(2\alpha_i)\bigr), 
\qquad 
\mu_i=f_{\mu,i}(x_{<i}),\;
\alpha_i=f_{\alpha,i}(x_{<i}).
\]

Draw independent noise $u_i\stackrel{\text{i.i.d.}}\sim\mathcal{N}(0,1)$ and set
\[
x_i=\mu_i+\exp(\alpha_i)\,u_i,
\]
which defines an invertible transformation $x=f(u)$ with 
$u=(u_1,\dots,u_I)\sim\mathcal{N}(0,I)$.
The inverse recursion
\[
u_i=(x_i-\mu_i)\exp(-\alpha_i)
\]
is strictly lower-triangular, giving
\[
\left|\det\nabla_x f^{-1}(x)\right|
  =\exp\!\Bigl(-\sum_{i=1}^{I}\alpha_i\Bigr).
\]

Substituting $u=f^{-1}(x)$ and the Jacobian determinant into the change-of-variables formula yields an exact, tractable log-likelihood, confirming that this autoregressive model is precisely a normalizing flow.  The official implementation of the Masked Autoregressive Flow (MAF) proposed by \cite{NIPS2017_6c1da886} is available at \href{https://github.com/gpapamak/maf}{this GitHub repository}.

\paragraph{GAMLSS (Generalized Additive Models for Location, Scale, and Shape)}

GAMLSS \cite{Rigby2005} extends standard generalized additive models by allowing multiple parameters of a chosen distribution family (e.g., location, scale, skewness, and kurtosis) to depend on covariates—including time. Thus, one may write
\[
Y(t)\;\sim\;D\!\bigl(\mu(t),\,\sigma(t),\,\nu(t),\,\tau(t)\bigr),
\]
where \(D\) is a parametric family and each parameter (e.g., \(\mu(t)\)) is modeled via a smooth function of \(t\). Although GAMLSS accommodates smoothly time‐varying distributions, it typically assumes a single fixed family and may not readily capture abrupt or highly nonlinear shifts across multiple dimensions.

In our application, we specialize \(D\) to the Gaussian (Normal) distribution, so that
\[
  Y(t)\;\sim\;\mathcal{N}\bigl(\mu(t),\,\sigma^2(t)\bigr),
\]
with both the time‐dependent mean \(\mu(t)\) and standard deviation \(\sigma(t)\) modeled using spline smoothers. We employ the official \texttt{gamlss} R package, which supports this and a rich variety of other distributions (e.g., Box–Cox, Student’s \(t\), and other skewed or heavy‐tailed laws) and enables model terms that include linear effects, spline‐based smoothers, random effects, or spatial components. We further leverage its diagnostic tools, model‐selection criteria (GAIC, BIC), and visualization routines.

% Multivariate Kernel Density Estimator (KDE) in four lines

\paragraph{Kernel Density Estimators}
Kernel density estimators (KDEs), see \cite{silverman1986density}, are non-parametric methods that approximate probability densities by centering a smooth kernel (e.g., Gaussian) at each data point and averaging. KDEs naturally adapt to complex distributions without assuming a fixed parametric form. Temporal dependence can be introduced by conditioning on time—fitting a separate KDE at each observed time point and linearly interpolating between the resulting density estimates. While conceptually simple and intuitive, this approach can become computationally expensive as the numbers of time points and observations grow. In our experiments, we fit a conditional KDE at each time instant of the observed series and then linearly interpolate in time to obtain smooth, time-varying density estimates.

\subsection{Results on synthetic data: low dimensions}

\begin{figure}[ht!]
  \centering
    \includegraphics[width=\textwidth]{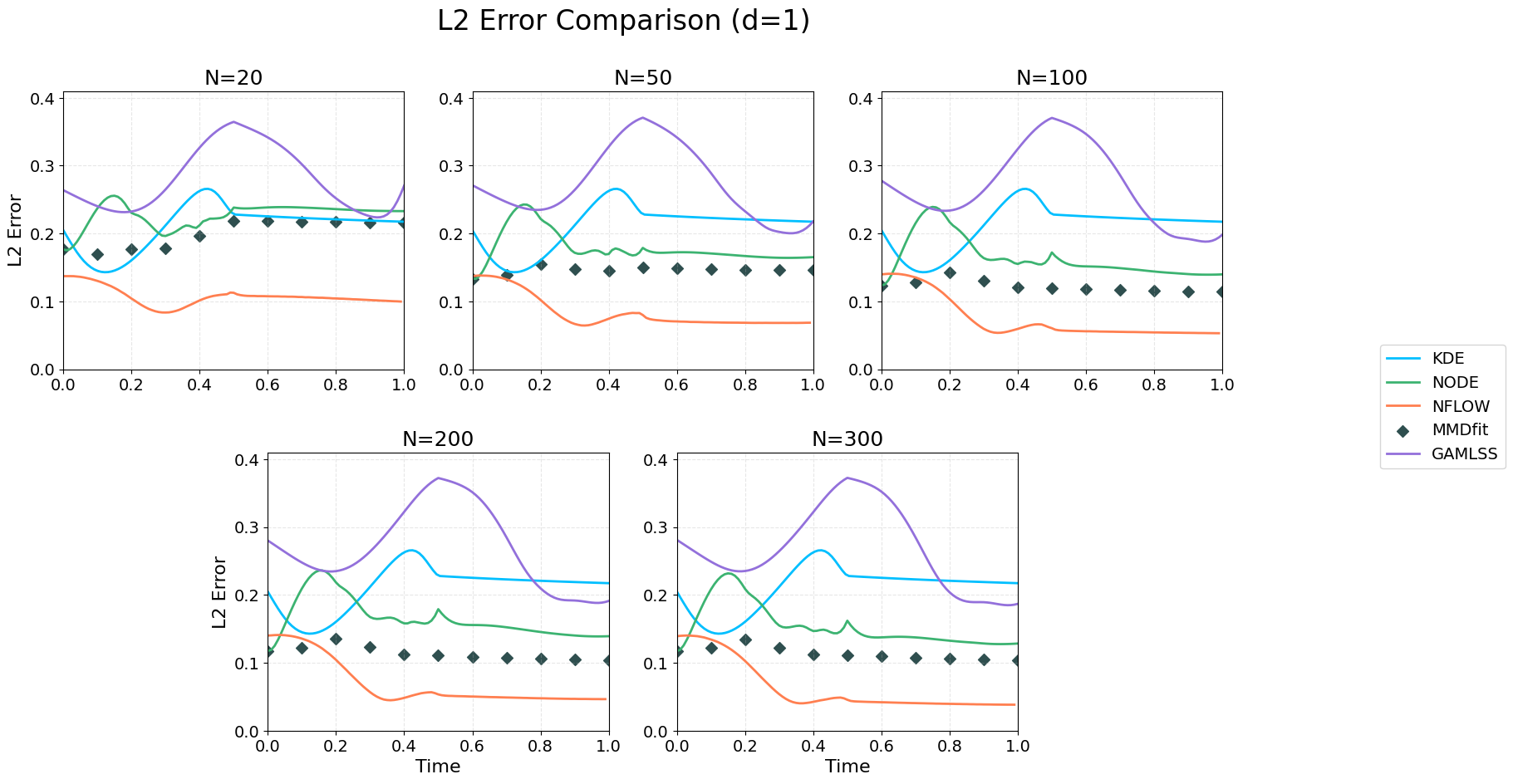} 
  \caption{Pointwise $L^2$-error over time in dimension $d=1$. We compare \textsc{NODE} (our model), \textsc{NFLOW}, \textsc{KDE} and \textsc{GAMLSS}.  Curves are averages over $B=100$ seeds, measured against the ground truth of three  Gaussians with time-dependent parameters. Errors for the discrete-time fitting are displayed as \textsc{MMDfit}.}
  \label{fig:l2errsd1}
\end{figure}

In dimension $d=1$, \cref{fig:l2errsd1} shows that our model quantitatively competitive with respect to the  alternatives. The NFLOW competitor may achieve slightly lower $L^2$-error, reflecting its higher expressivity under strong regularity, but our goal is to remain competitive while providing interpretable time-varying  distribution with clinically meaningful mixture weights. We therefore accept a small loss in expressivity in exchange for transparency.

\subsection{Results on synthetic data: high dimensions}
\begin{figure}[ht!]
  \centering
    \includegraphics[width=\textwidth]{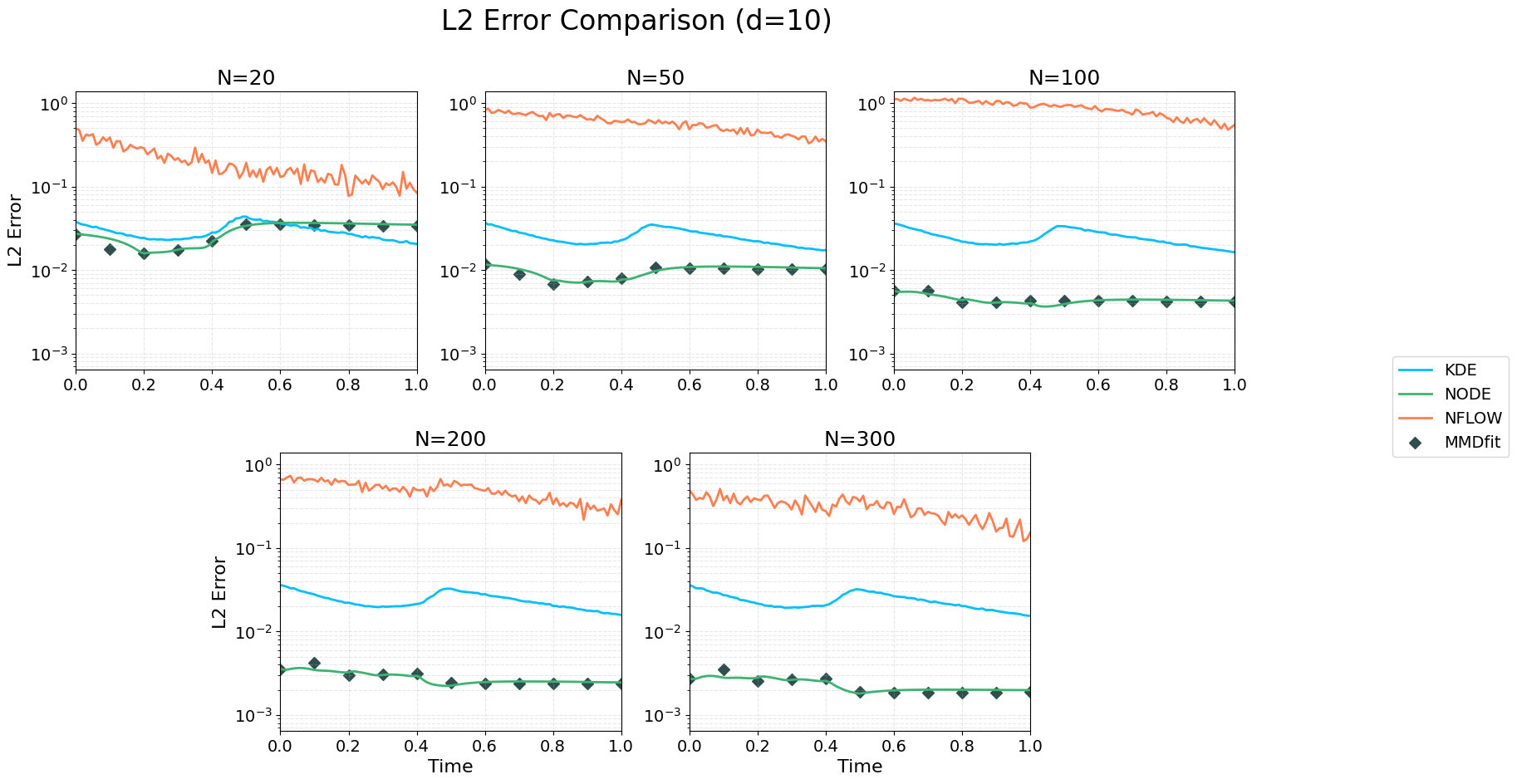} 
\caption{Pointwise $L^2$-error over time. We compare \textsc{NFLOW}, \textsc{KDE}, and \textsc{NODE} (our model). Curves are averages over $B=100$ simulations. Errors for the discrete-time fitting are displayed as \textsc{MMDfit}.}
  \label{fig:l2errsd10}
\end{figure}

In dimension $d = 10$, \cref{fig:l2errsd10}  shows that our model  outperforms the alternatives. In this setting, the method with the worst performance is KDE, which is widely recognized to be highly sensitive to dimensionality \cite{tsybakov2009nonparametric}.

\section{Digital health diabetes case study}
\label{sec:5}
From a distributional data analysis perspective, this study aims to demonstrate the explainability advantages of our novel probabilistic framework for detecting distributional shifts in longitudinal clinical data.

\subsection{Data description and scientific question}

Our study is motivated by the Juvenile Diabetes Research Foundation (JDRF) Continuous Glucose Monitoring randomised controlled trial \cite{doi:10.1056/NEJMoa0805017,juvenile2009effect}.%
\footnote{Publicly available at \url{https://public.jaeb.org/datasets/diabetes}.}  
A total of $451$ participants (adults and children) with type~1 diabetes mellitus (T1DM) were randomised to either a \emph{treatment} arm---receiving continuous glucose monitoring (CGM) data and education---or a \emph{control} arm relying on standard self-monitoring without CGM feedback.

For data-quality reasons, we restrict our analysis to the $443$ individuals with a maximum monitoring period of $26$~weeks. Descriptive statistics are reported in the Supplementary Material. Our primary clinical question is whether continuous glucose monitor (CGM) use (\emph{treatment}) versus usual care (\emph{control}) leads to different individual longitudinal glucose trajectories over the course of the trial. The continuous-time nature of our model is ideally suited to address this question because it allows us to characterize how the entire glucose distribution evolves over the full follow-up period.

%A secondary goal is to use the first-month data to build a predictive model for each patient’s glycaemic profile during the final study week.

\subsection{Modeling the marginal density of glucose time series}

Let $Y_i(t)$ denote the CGM reading for participant $i$th at continuous time $t\in \mathcal{T}$. In practice, for a total number of $n$ participants, we focus on a time--discrete process with weekly observations $t\in\mathcal{T}_i$.
Within the interval $(t-1,t]$ we collect $n_{it}$ CGM readings.
Our goal is to estimate the time-varying distribution function
\[
  F_{i}(x,t)\;=\;\mathbb{P}\bigl(Y_i(t)\le x\bigr).
\]

\noindent For computational and comparison purposes we assume that, for all $i = 1, \dots, n$, the observed data can be approximated enough well by the mixture density model
\[
f_{i}(x, t) = \sum_{s=1}^{K} \alpha_{is}(t)\, \mathcal{N}\!\bigl( m_{s}, \Sigma_{s}\bigr),
\]
\noindent where $\mathcal{N}(m_s, \Sigma_s)$ denotes a Gaussian distribution with mean $m_s$ and scalar variance $\Sigma_s$, appropriate for modeling the univariate biomarker. To ensure comparability, we impose the constraint that the Gaussian parameters $(m_s, \Sigma_s)$ of the mixture components are shared globally across all individuals; only the time-dependent mixing weights $\alpha_{is}(t)$ vary by participant.

\paragraph{Choice of the number of components.}
We set $K=3$ to encode clinically meaningful phenotypes of glycaemic control (good, intermediate, poor). The same three components are used for all participants to enable cross-subject comparability; accordingly, $(m_s,\Sigma_s)_{s=1}^3$ are shared across individuals, whereas only the time-varying mixture weights $\alpha_{is}(t)$ are subject-specific. Interpreting the trajectories of $\alpha_{is}(t)$ within these three reference groups yields a clinically transparent summary of distributional dynamics.

\begin{figure}[h!]
    \centering
    \includegraphics[width=0.32\textwidth]{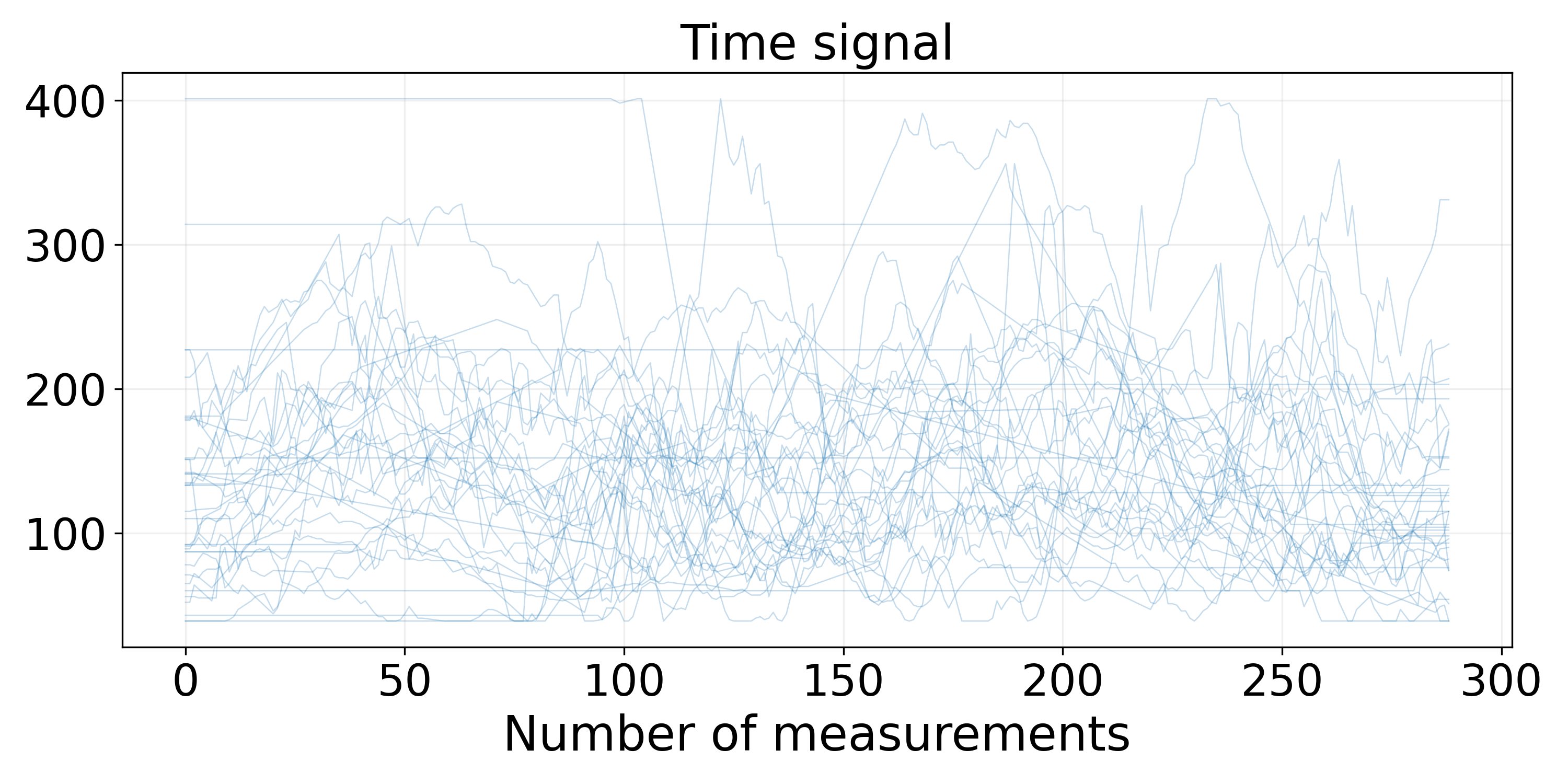}
    \includegraphics[width=0.32\textwidth]{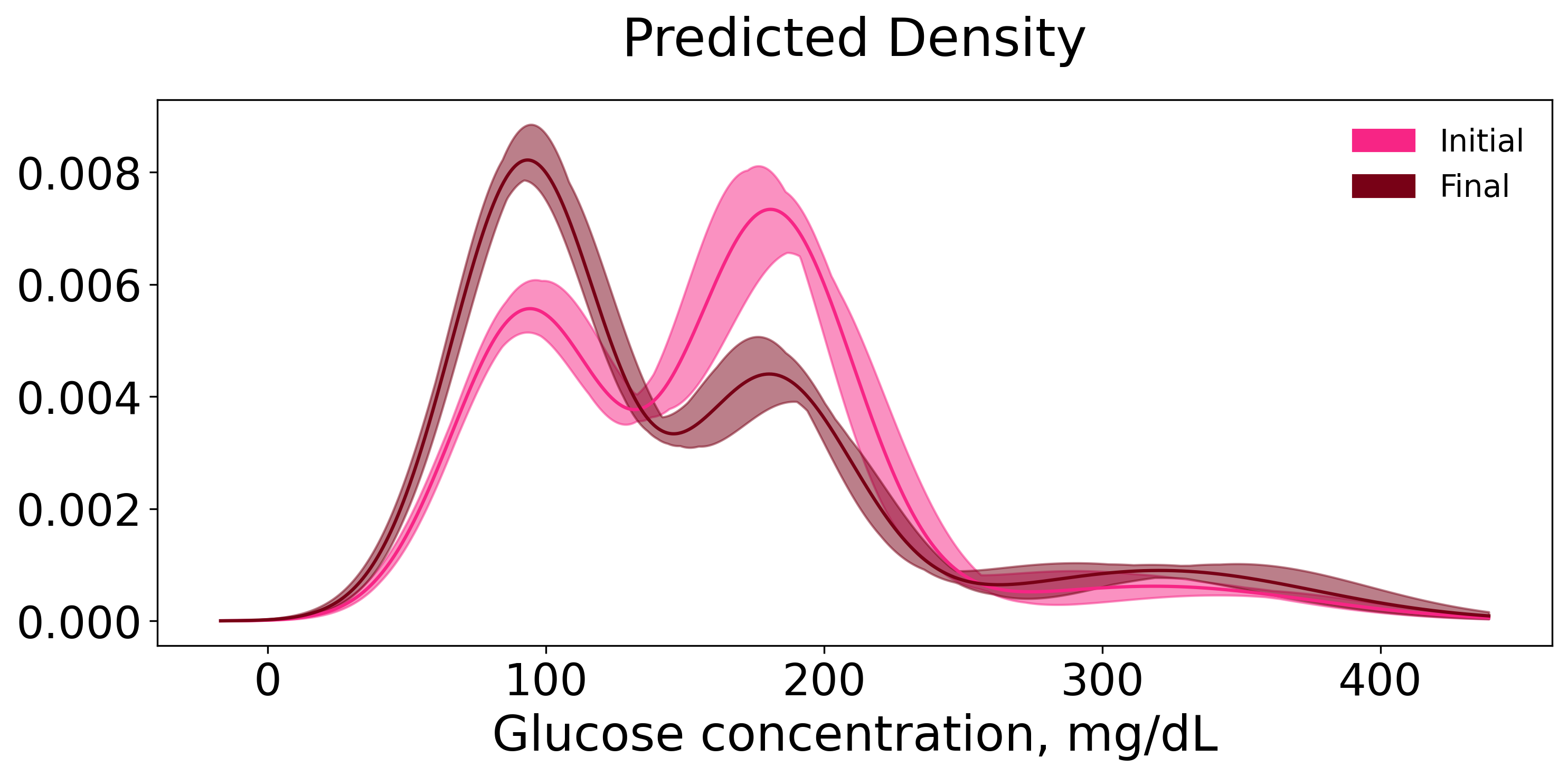}\includegraphics[width=0.32\textwidth]{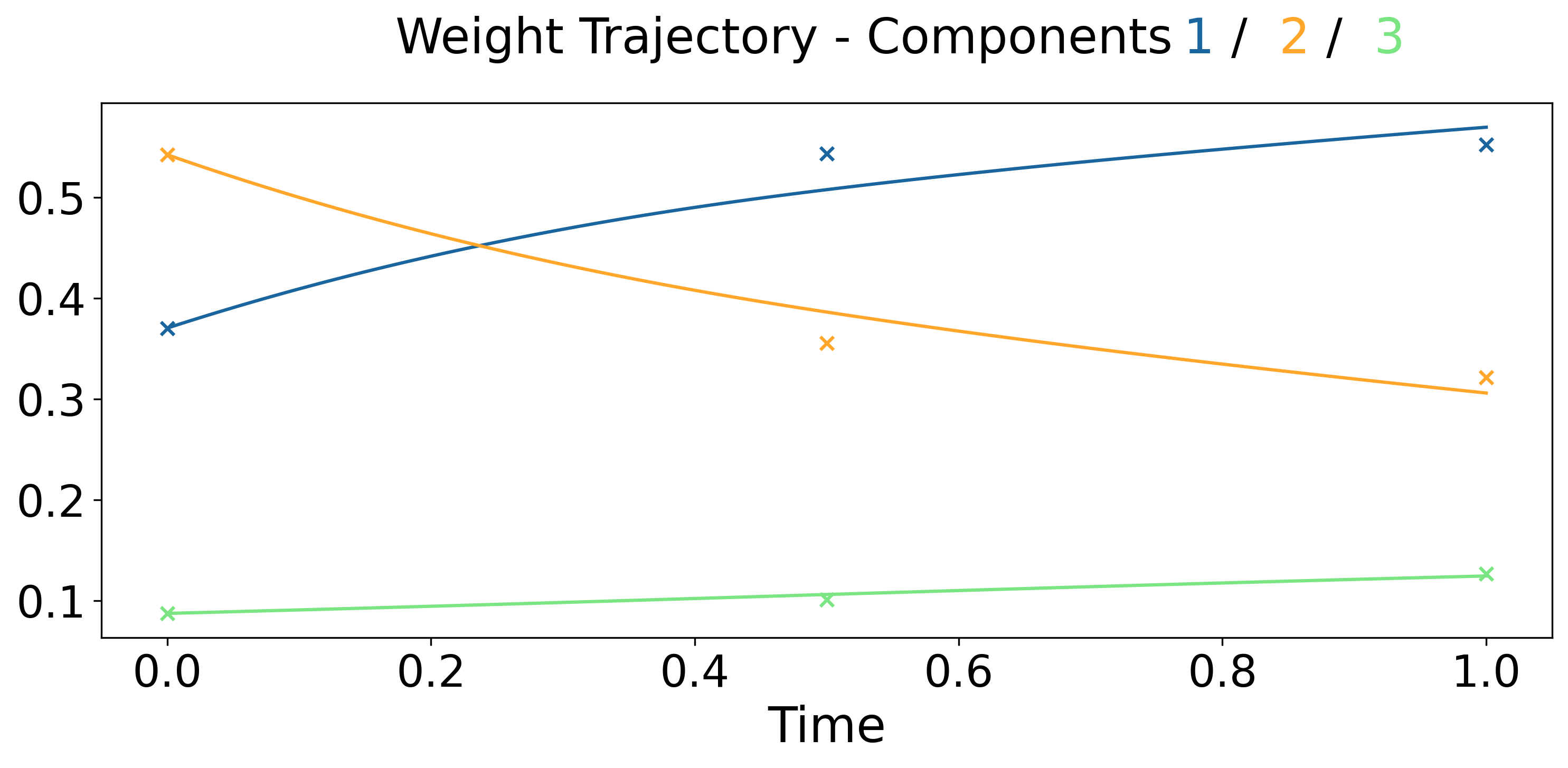}
    \\[1ex]
      \includegraphics[width=0.32\textwidth]{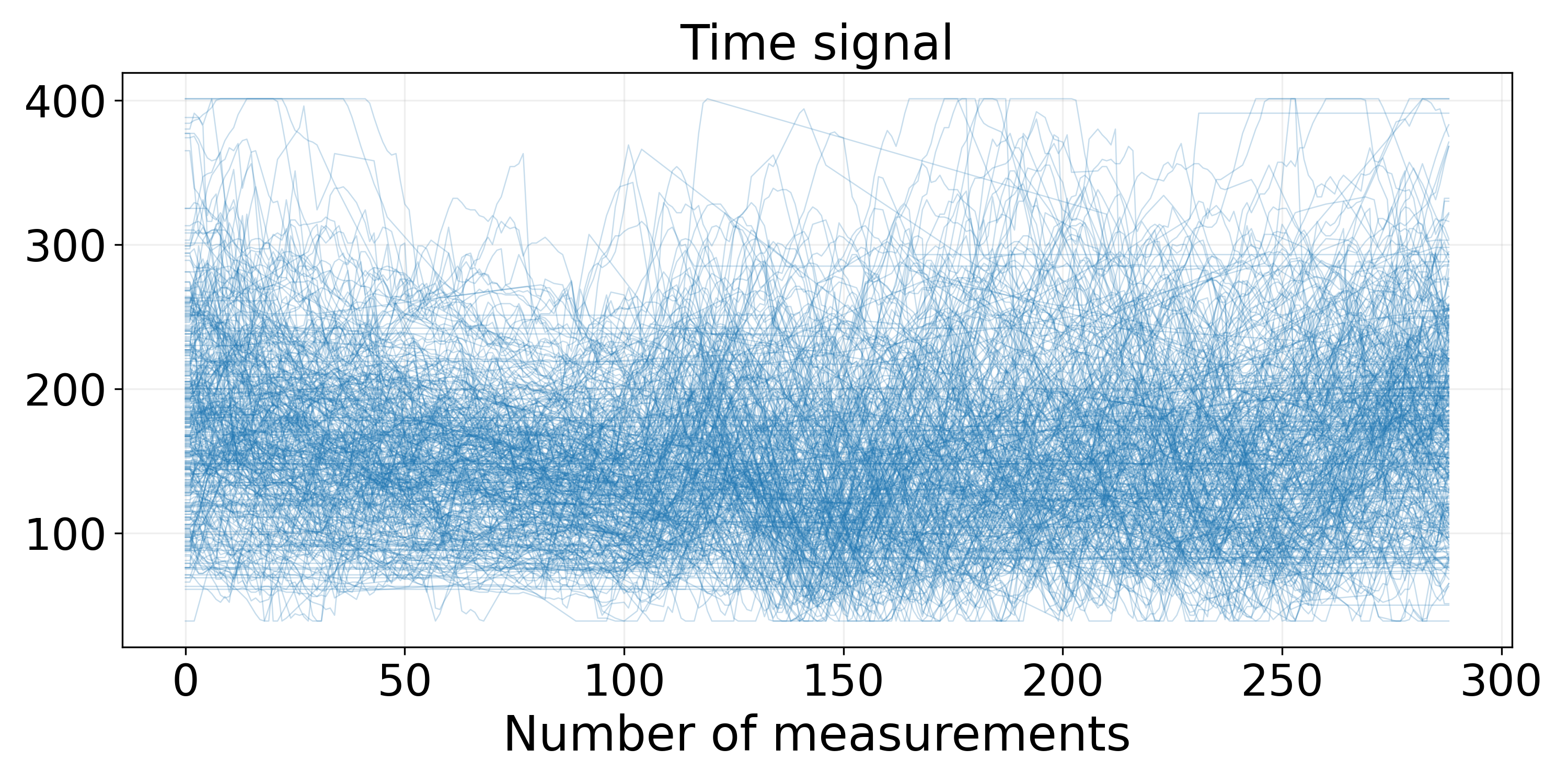}\includegraphics[width=0.32\textwidth]{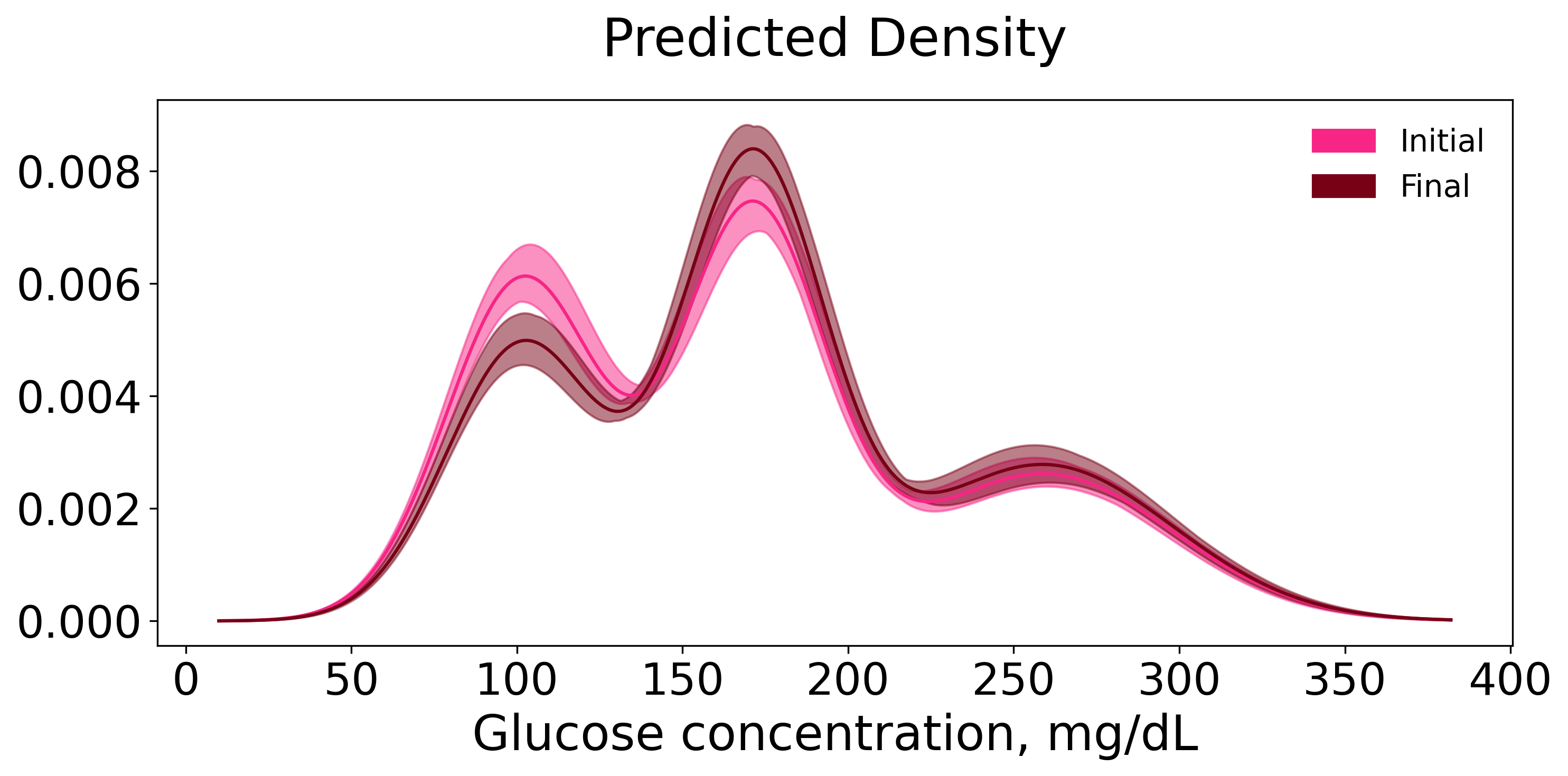}
      \includegraphics[width=0.32\textwidth]{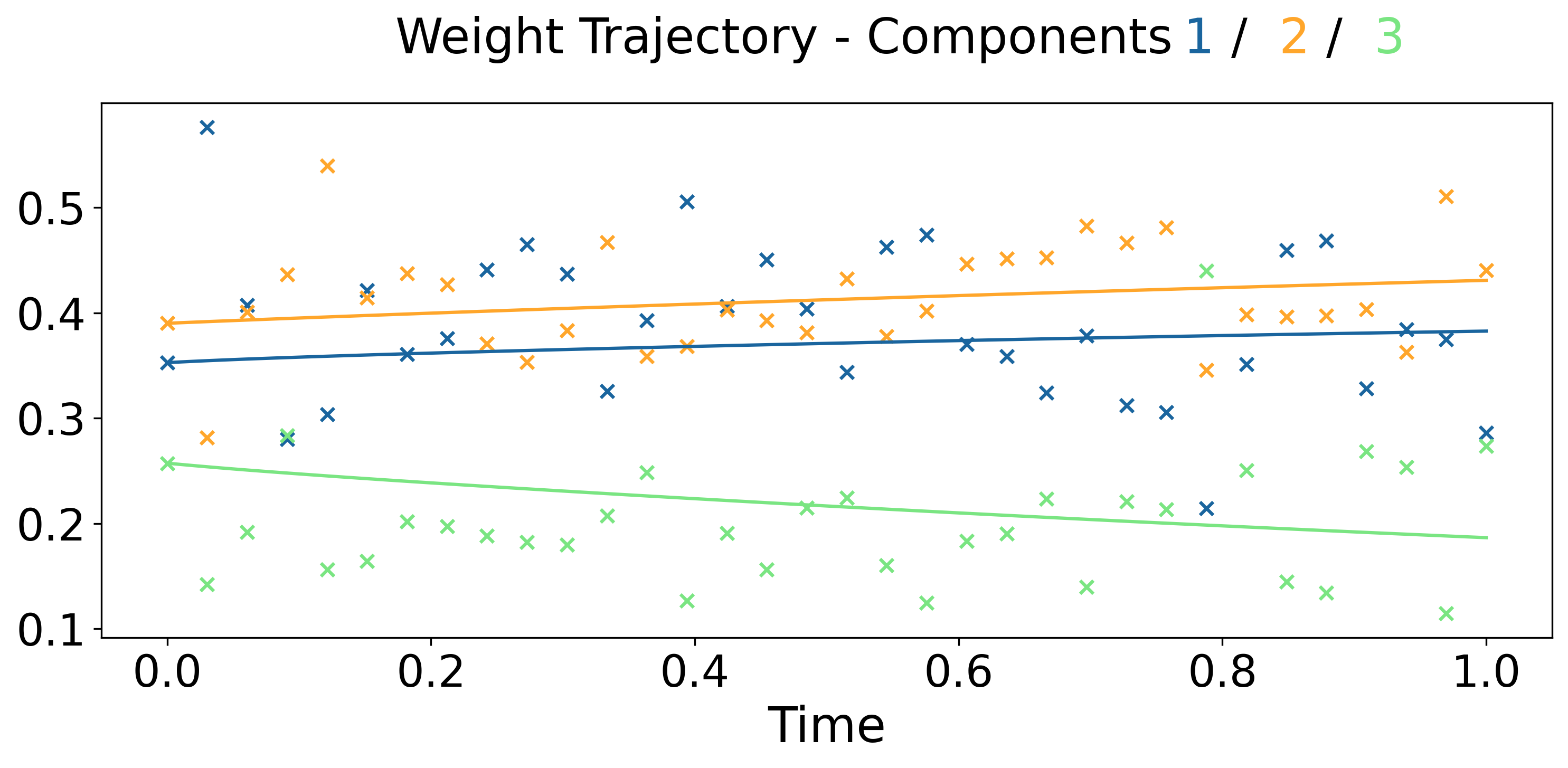}
  \\[1ex]
    \includegraphics[width=0.32\textwidth]{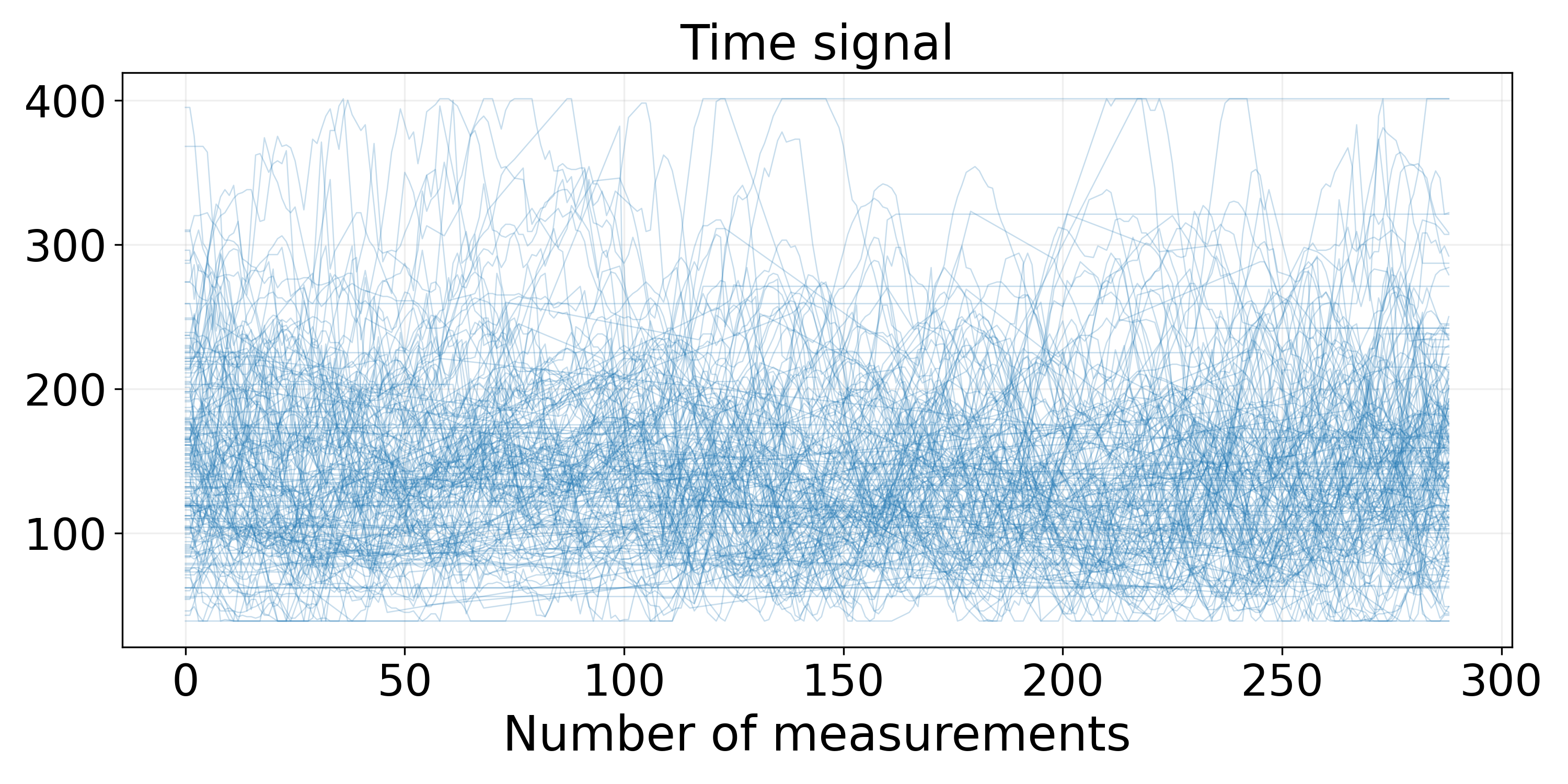}
    \includegraphics[width=0.32\textwidth]{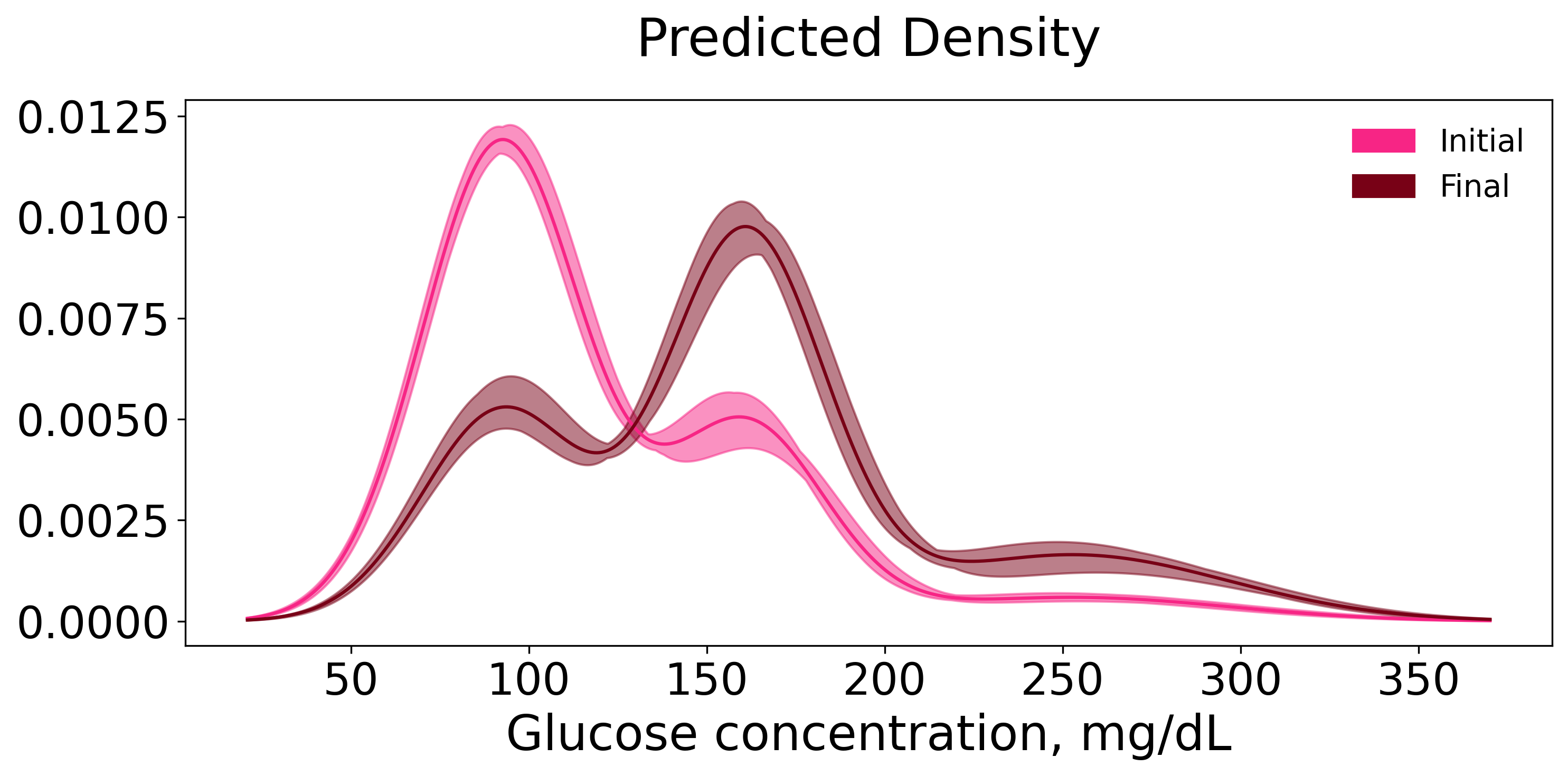}
    \includegraphics[width=0.32\textwidth]{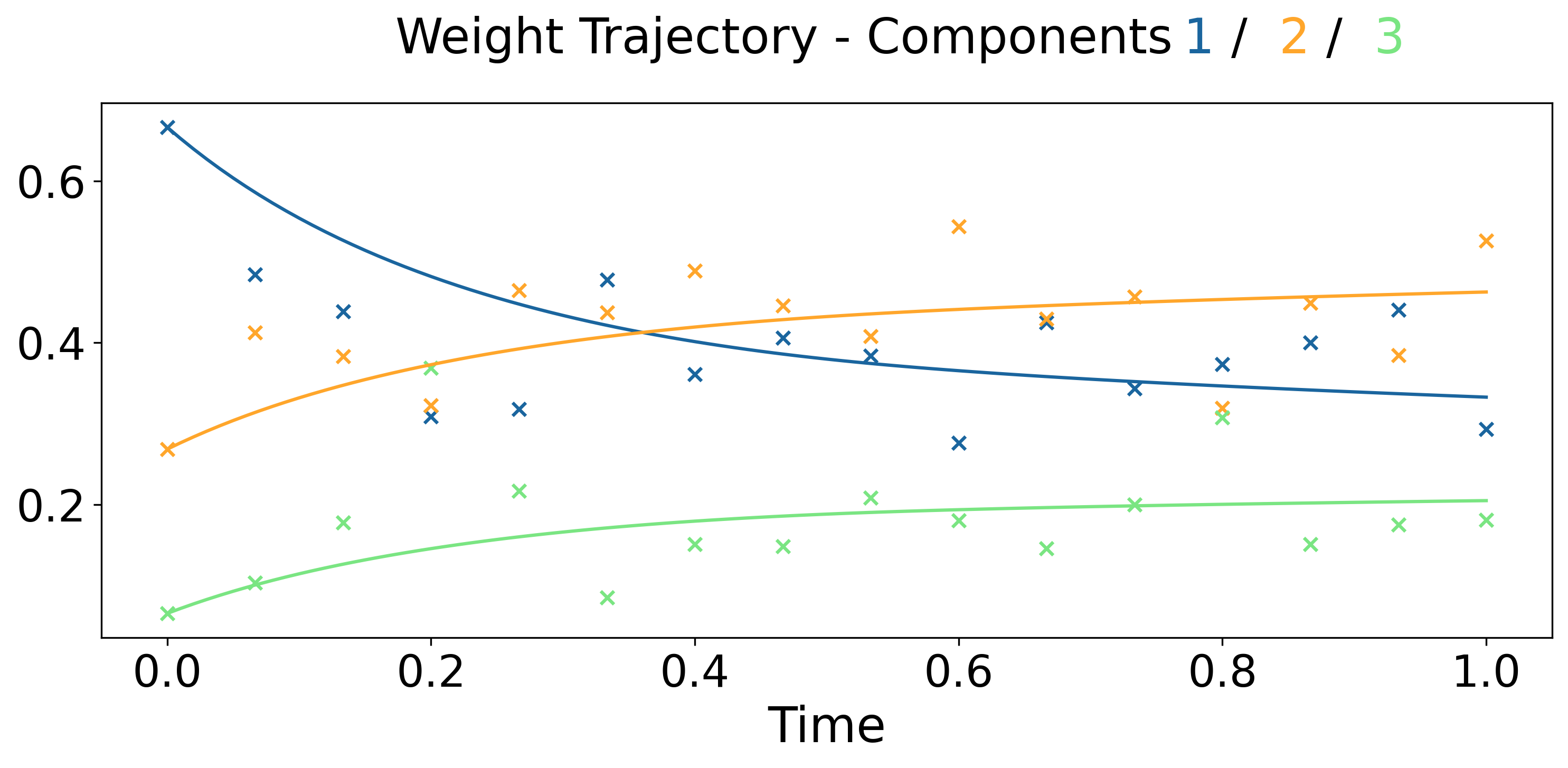}

 \caption{Three representative subjects (IDs 13, 62, and 377, one one each row). Left: the patient’s full CGM time series. Middle: initial (pink) and final (dark red) MMD-fitted densities with 95\% bootstrap confidence bands. Right: weight trajectories $\alpha_{s}(t)$ from the neural ODE (Step~2); endpoints match the middle-column fits.}
    \label{fig:trajexample}
\end{figure}

 %using 100 bags to construct 80\% boostrap bands, with the average densities highlighted; and the weight trajectories predicted by the neural ODEs alongside the average fitted discrete values.

\begin{figure}[h!]
    \centering
    \begin{tabular}{ccc}
        \includegraphics[width=0.3\textwidth]{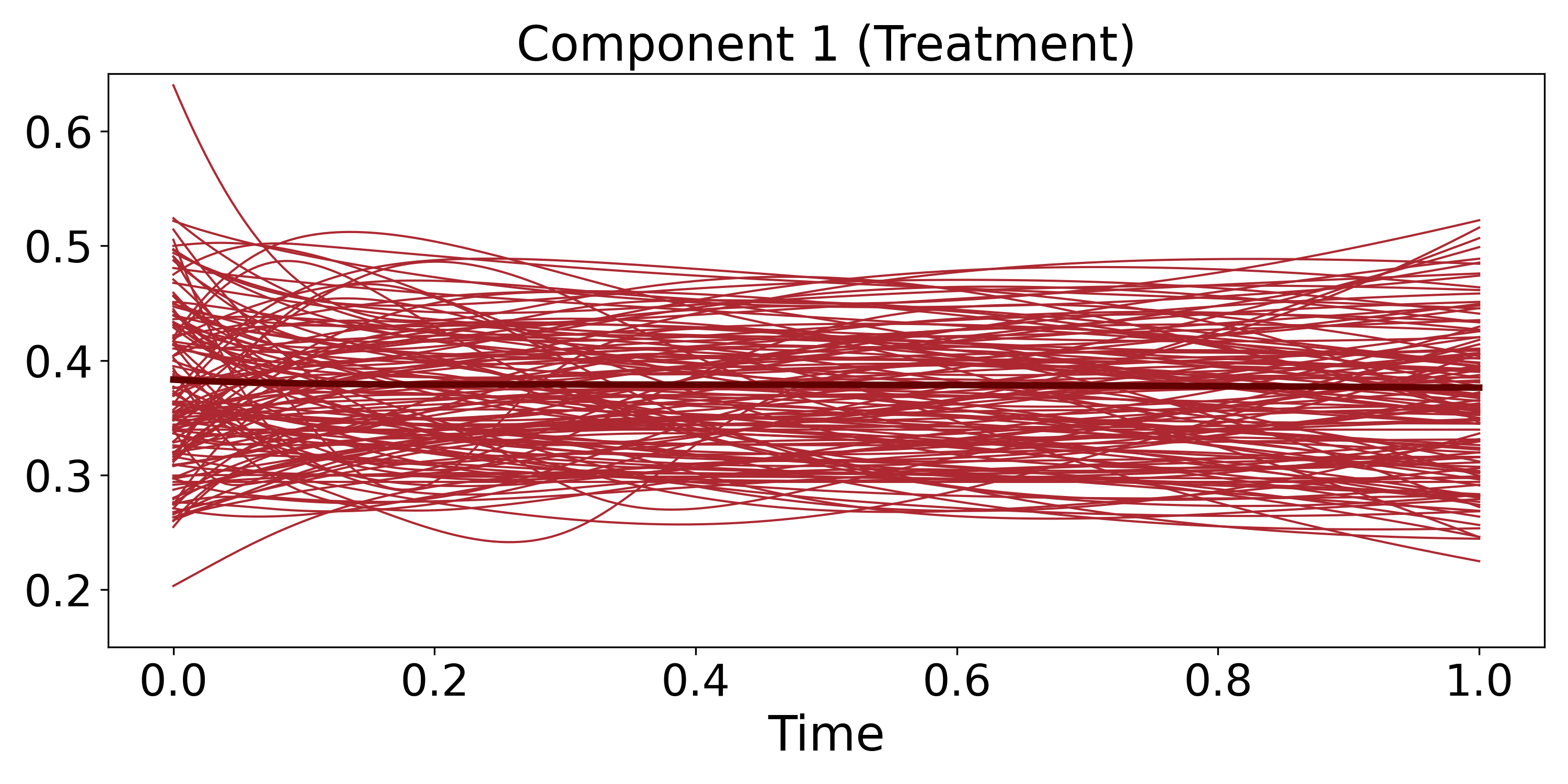} &
        \includegraphics[width=0.3\textwidth]{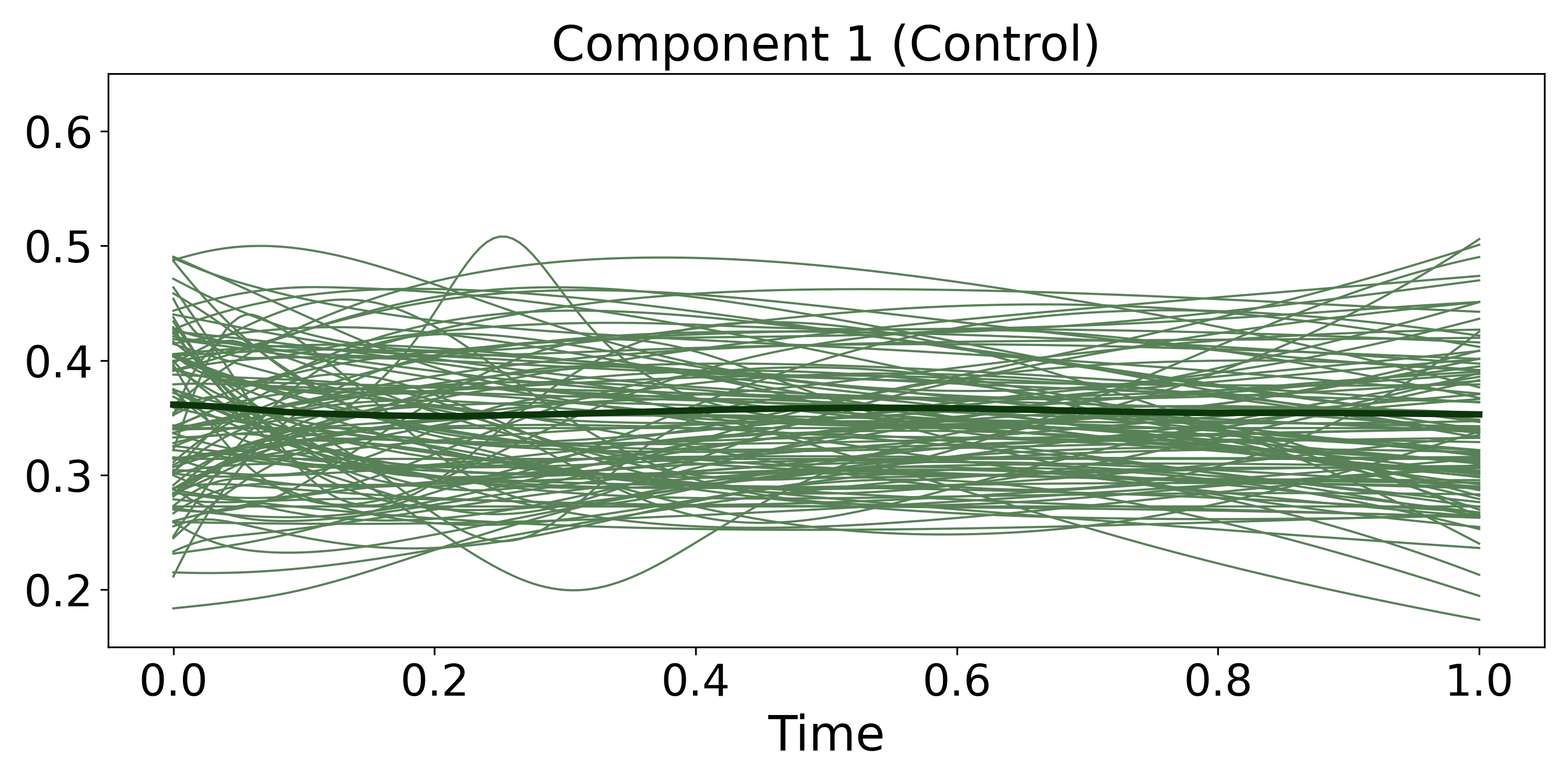} &
        \includegraphics[width=0.3\textwidth]{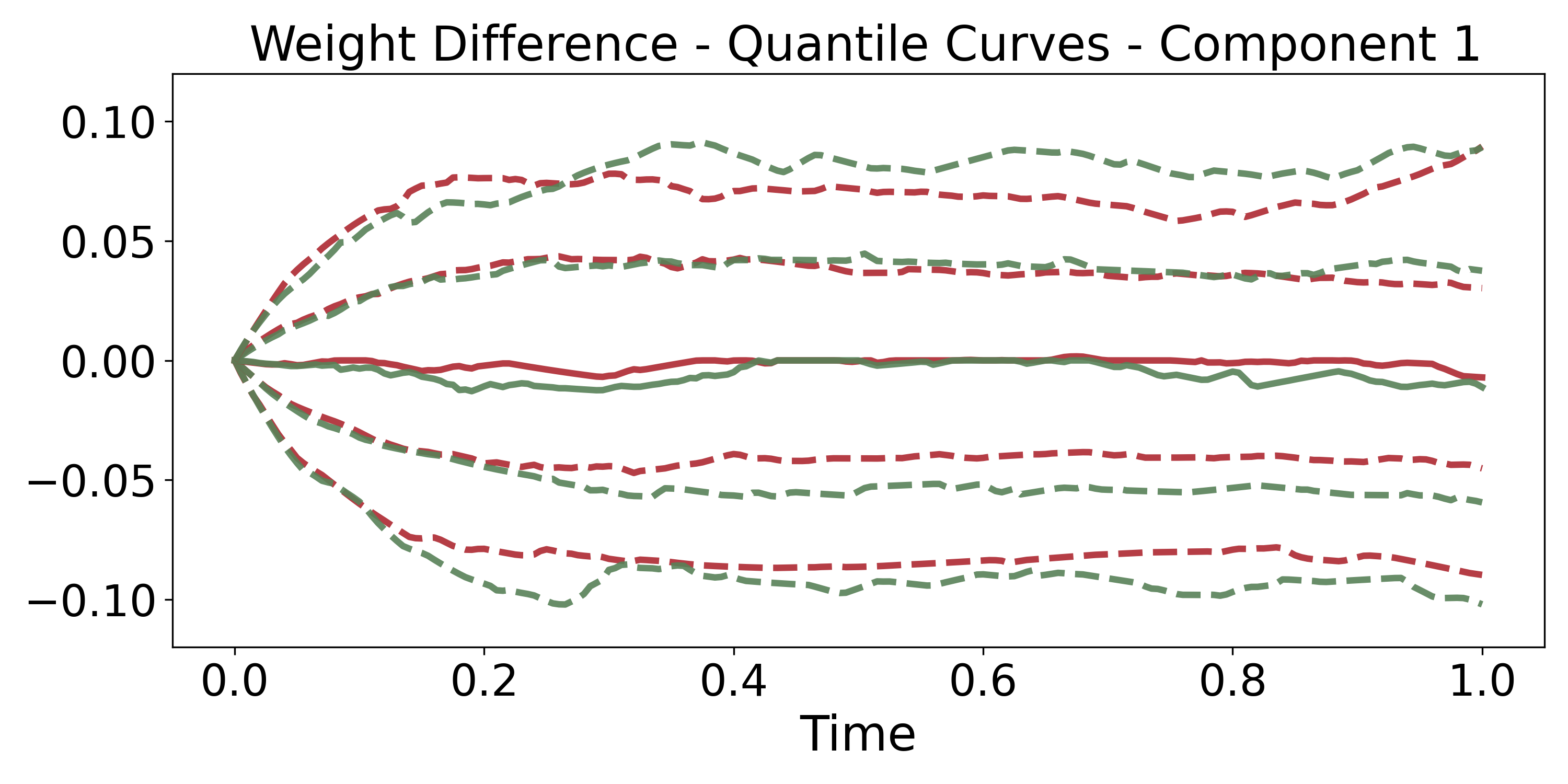} \\
        \includegraphics[width=0.3\textwidth]{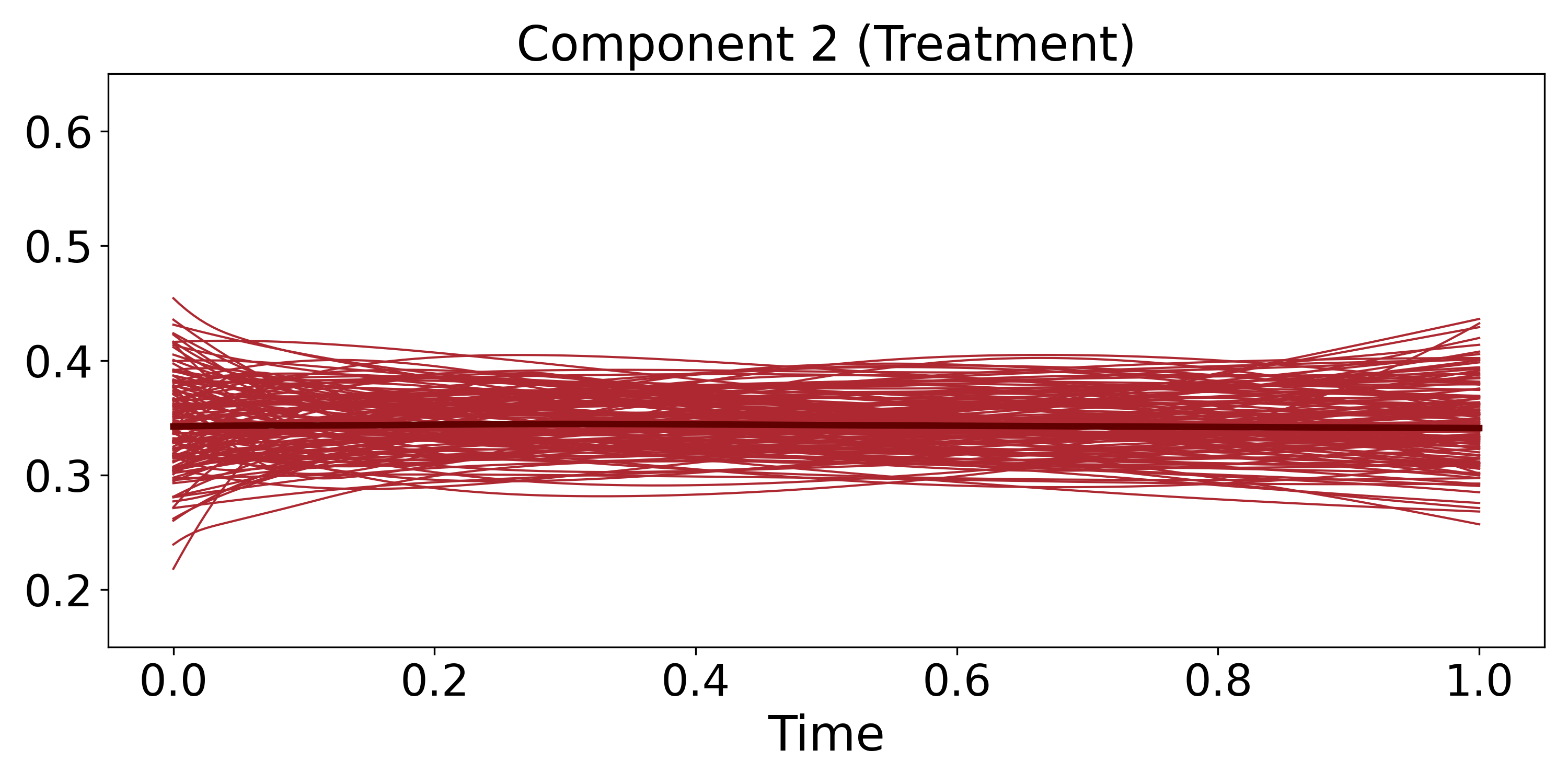} &
        \includegraphics[width=0.3\textwidth]{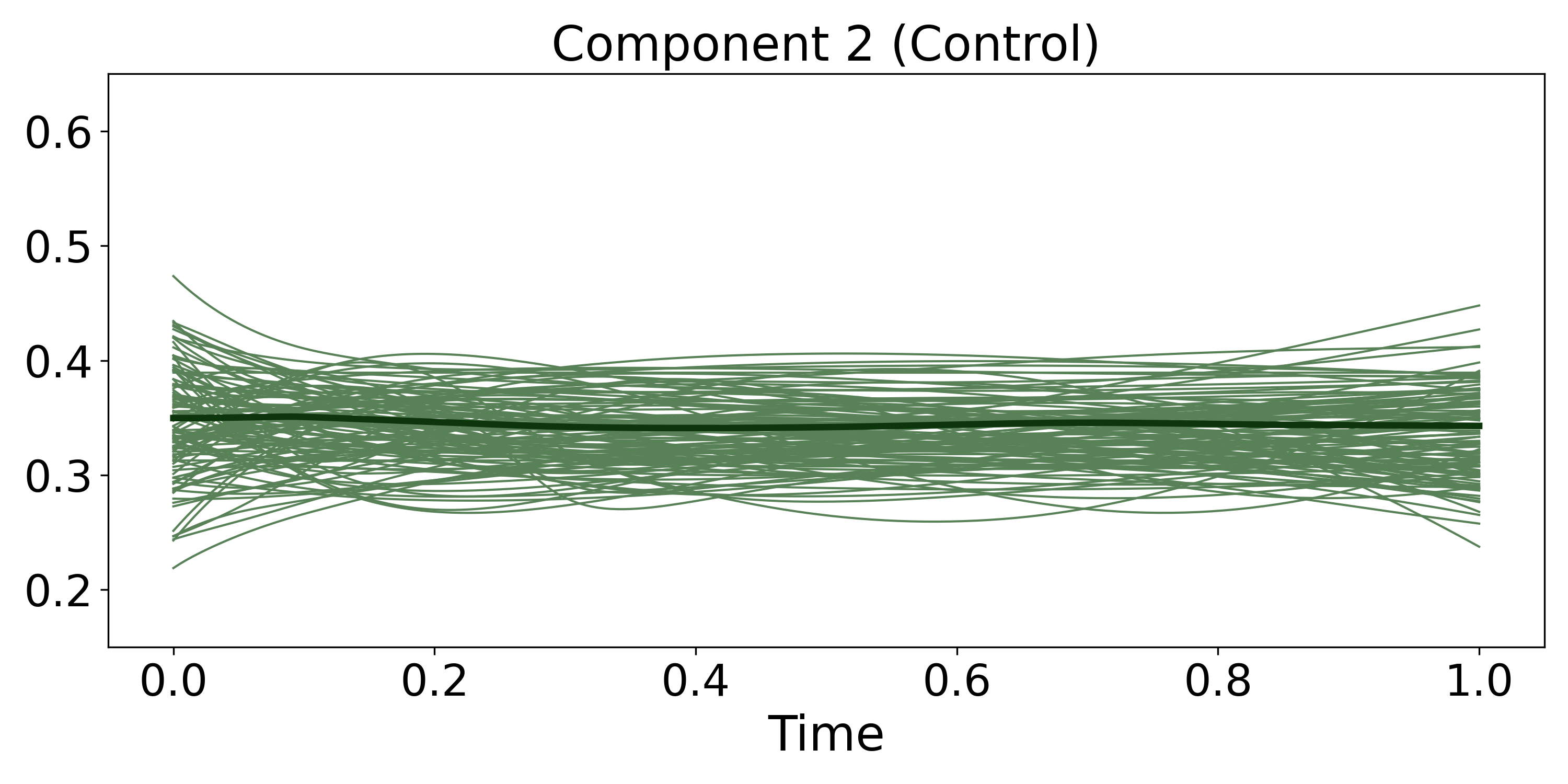} &
        \includegraphics[width=0.3\textwidth]{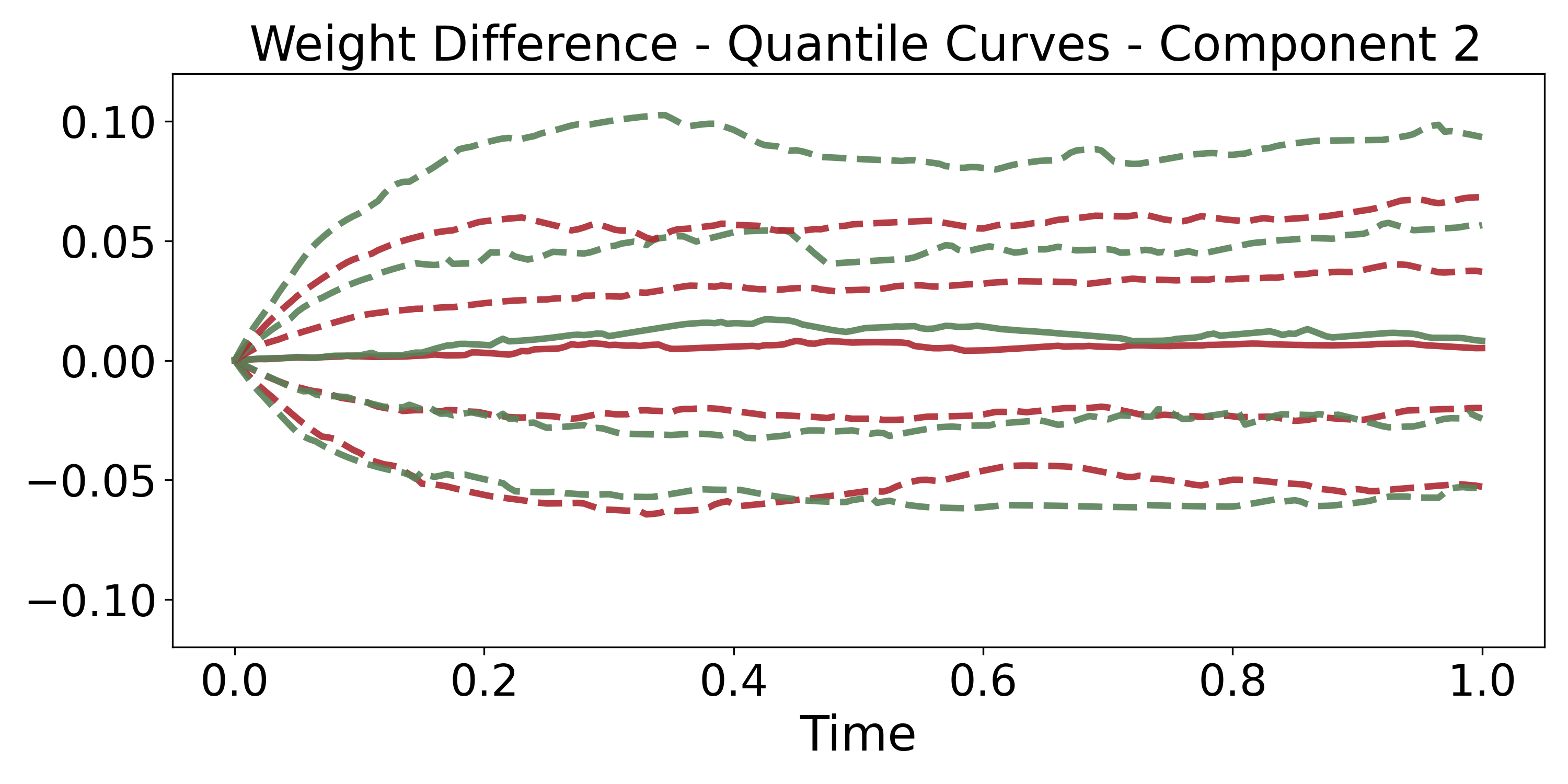} \\
        \includegraphics[width=0.3\textwidth]{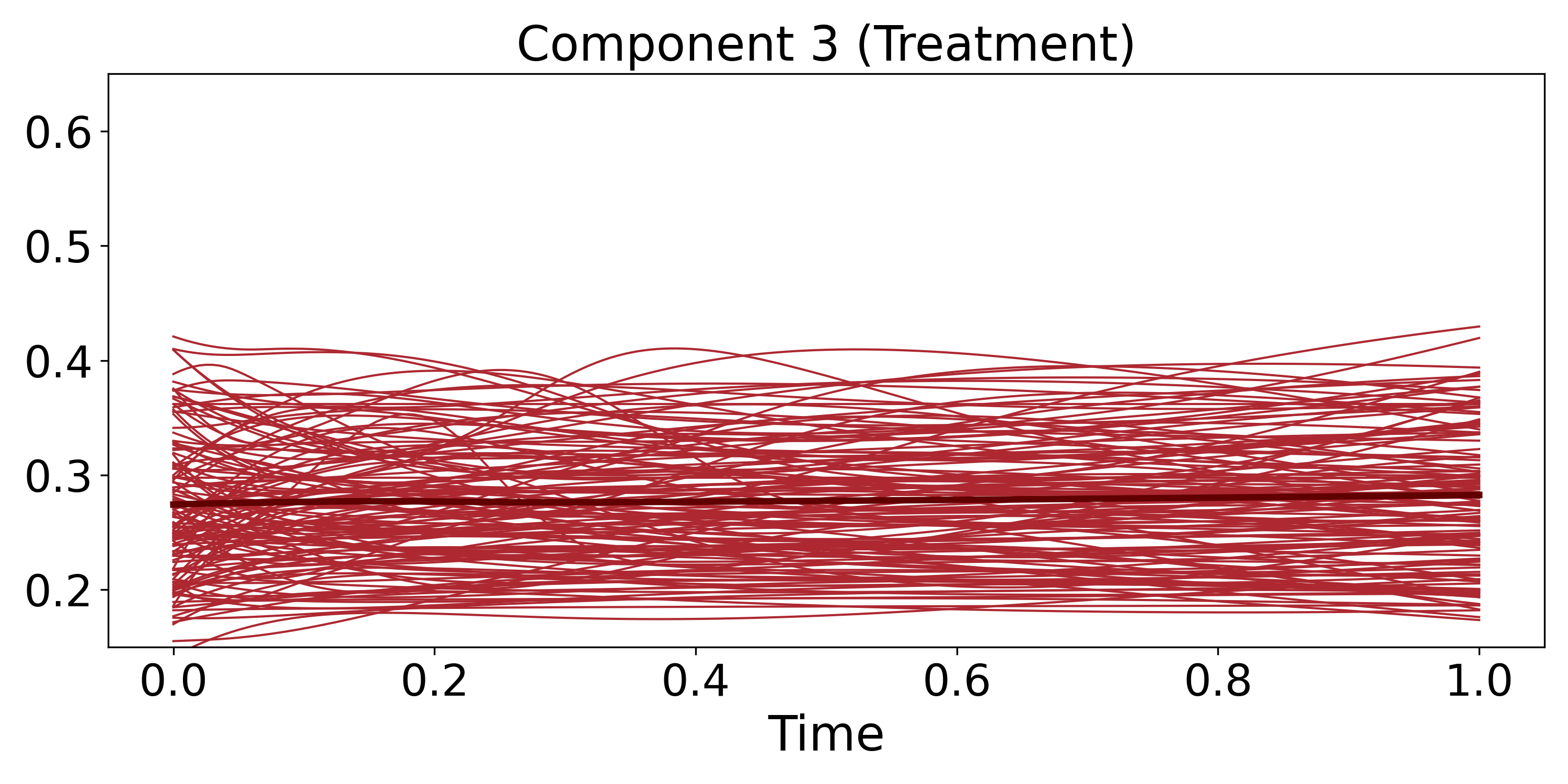} &
        \includegraphics[width=0.3\textwidth]{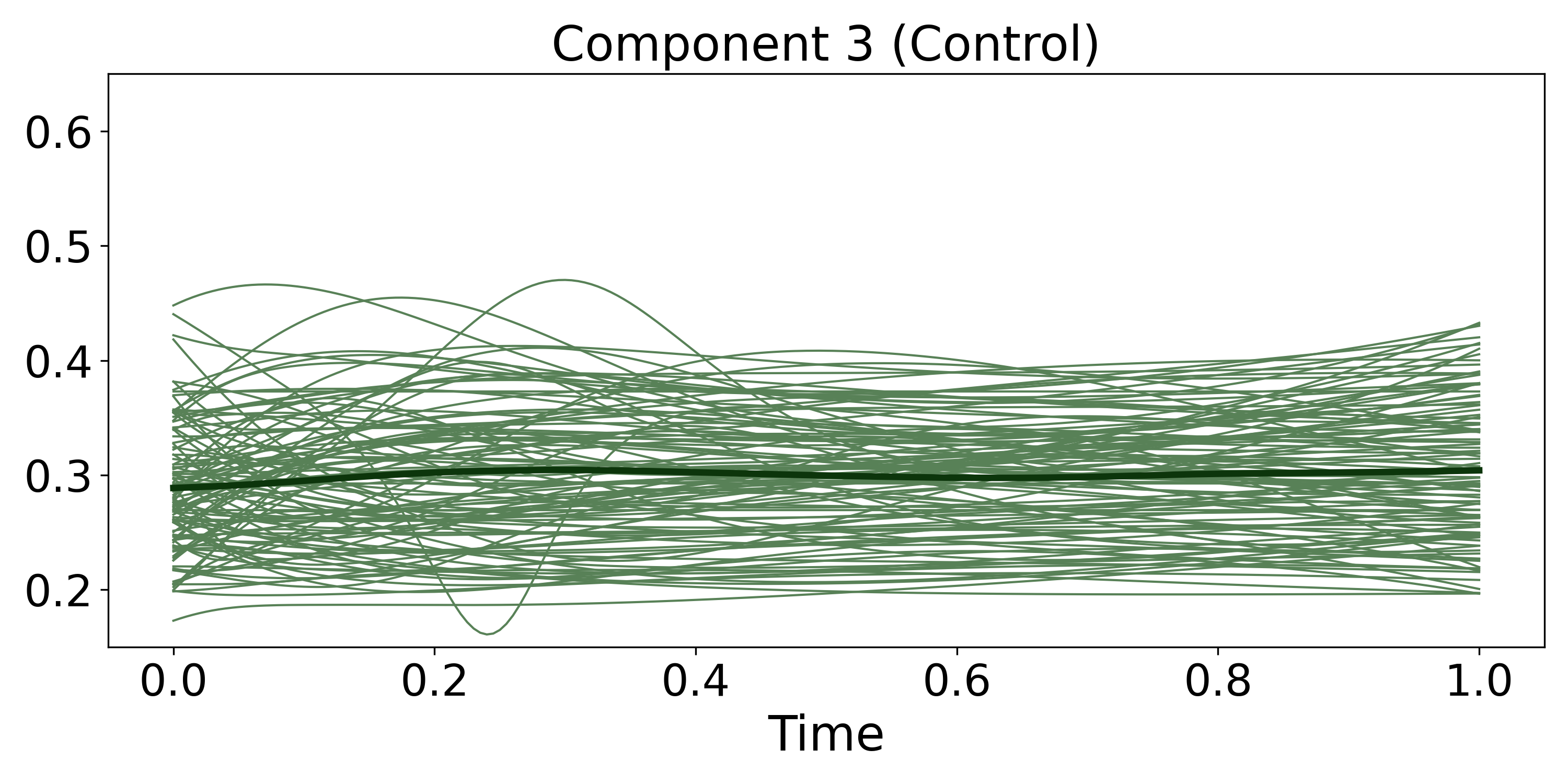} &
        \includegraphics[width=0.3\textwidth]{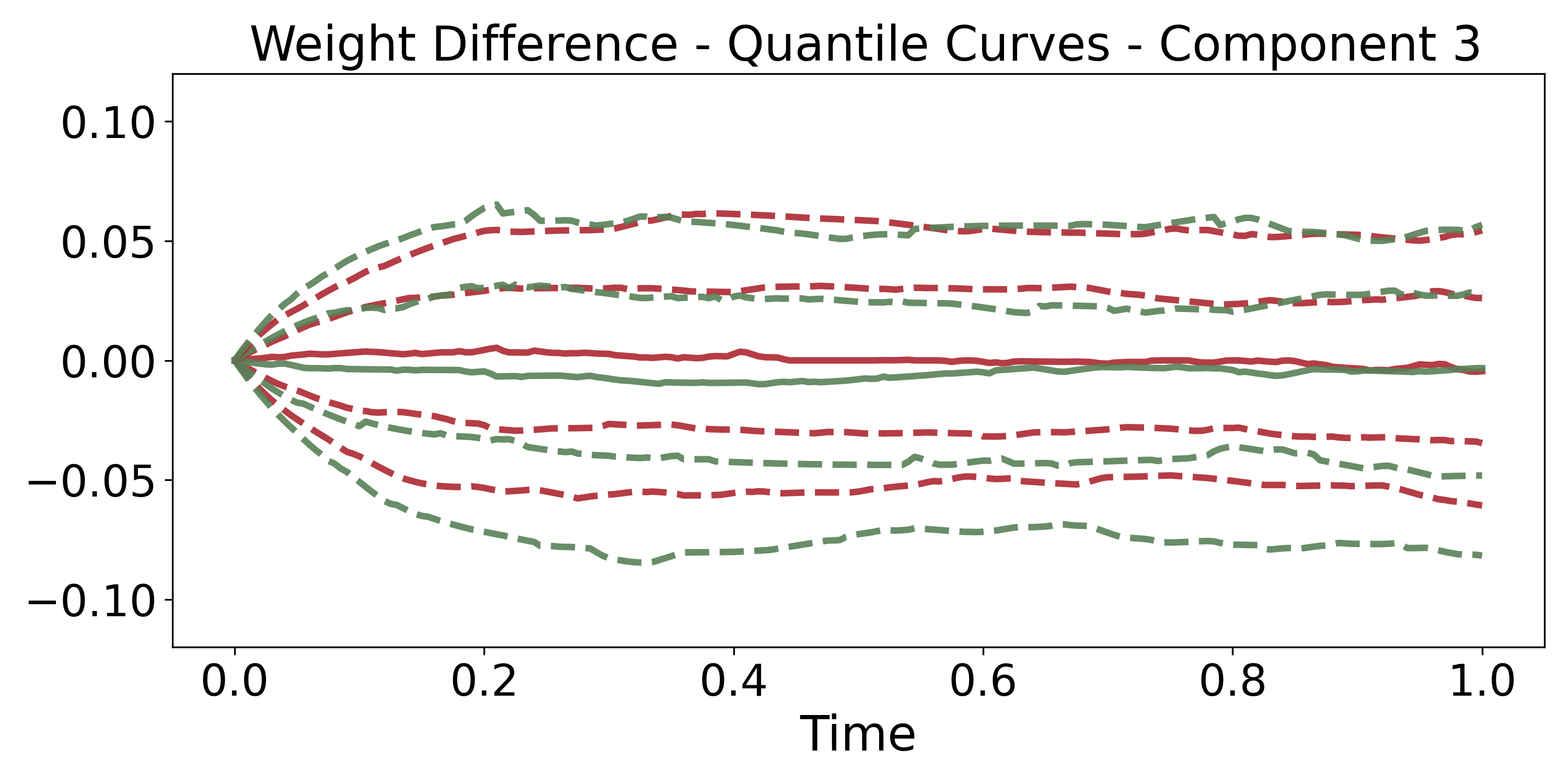} \\
    \end{tabular}
       \caption{Component-wise weight dynamics by arm (components $s = 1, 2, 3$ in rows from top to bottom). Columns 1-2: individual curves $\alpha_s(t)$ (thin) and mean (thick) for treatment (left) and control (right) groups. Column~3: $(0.1, 0.25, 0.5, 0.75, 0.9)$-quantile trajectories of the process $Z_{is}$ defined in \eqref{eq:Z_is} across subjects for each of the two groups and by arm.}
   \label{fig:weights_by_arm}
\end{figure}

{\footnotesize
\begin{table}[ht]
  \centering
  \caption{Calibrated global statistics for univariate  and bivariate data.}
  \begin{tabular}{l|ccc}
    \toprule
    \multicolumn{4}{l}{\textbf{Univariate ($d=1$): Fitted global means and variances}} \\
    \midrule
    Parameter & \(s=1\) & \(s=2\) & \(s=3\) \\
    \midrule
    \(\mu_s\)        & \(97.81\)  & \(167.84\) & \(259.37\) \\
    \(\sigma_s^2\)   & \(537.72\) & \(501.04\) & \(1737.04\) \\
    \midrule
    \multicolumn{4}{l}{\textbf{Bivariate ($d=2$): Fitted global means and covariance matrices}} \\
    \midrule
    Parameter & \(s=1\) & \(s=2\) & \(s=3\) \\
    \midrule
    \(\boldsymbol{\mu}_s\) &
      \((97.81,\,0.58)\) &
      \((167.66,\,0.21)\) &
      \((258.87,\,-0.059)\) \\
    \(\boldsymbol{\Sigma}_s\) &
      \(\begin{bmatrix}537.75 & -27.01\\[-2pt] -27.01 & 124.26\end{bmatrix}\) &
      \(\begin{bmatrix}501.01 & 1.41\\[-2pt] 1.41 & 63.62\end{bmatrix}\) &
      \(\begin{bmatrix}1737.04 & -18.21\\[-2pt] -18.21 & 208.75\end{bmatrix}\) \\
    \bottomrule
  \end{tabular}
  \label{tab:global-statsc2}
\end{table}}

\Cref{tab:global-statsc2} reports the global mean and variance of each Gaussian component.  
From a clinical perspective, these parameters were selected to represent three clearly differentiated glycaemic states:
(i) well-controlled diabetes condition, (ii) suboptimal diabetes control, and (iii) poorly controlled diabetes.  
By computing, at every time, the proportion of participants falling within these three archetypes \cite{cutler1994archetypal} (or phenotypes),  
we obtain a dynamic, standardized index that summarizes the longitudinal evolution of glycaemic profiles for every subject in the trial.

\Cref{fig:trajexample} shows the three representative day-to-day CGM trajectories together with their estimated marginal densities, highlighting the intrinsic complexity of the continuous-time series. In the left panel we observe substantial intra-day variability, suggesting that an analysis at the day level is advisable. To make subjects with different visit schedules comparable, time was rescaled to the unit interval $[0,1]$. We then focus on the subject-specific weight functions $\alpha_{is}(t)$.

For the three selected participants, the first subject shows an improvement (density shifts left and weight $\alpha_{i1}$ increases),  
the second remains relatively stable, and the third worsens (density shifts right and weight $\alpha_{i1}$ decreases).  
Importantly, some patients deteriorate over time even in the treatment arm, consistent with previous findings that only a subset of this cohort responds to the intervention. We also observe that certain participants exhibit apparent improvements in CGM metrics in the control group; however, these changes may reflect unrecorded external interventions---such as additional insulin dosing---rather than the effect of the study treatment. Moreover, despite randomization, baseline characteristics are not perfectly balanced between control and treatment groups (e.g. the treatment arm began with better glycaemic control) \cite{matabuena2024multilevel, juvenile2010effectiveness}.

\Cref{fig:weights_by_arm} contrasts the estimated mixture weights for the treatment and control arms.  
Participants in the treatment group devote a larger share of time to phenotype~1 across the entire follow-up, reflecting superior glycaemic control at all assessment points, including baseline.  To investigate individual temporal dynamics, we examined the centered trajectories.
\begin{equation}\label{eq:Z_is}
  Z_{is}(t) = \alpha_{is}(t) - \alpha_{is}(0), \hspace{1cm} s = 1,2,3 \text{ and } i=1\dots, n.
\end{equation}
 For each arm, we plot the selected quantiles of the $Z_{is}(\cdot)$ process. For phenotype 1, the quantile curves are nearly identical---once good control is reached, further gains are limited. For phenotype 3 (the worst glycaemic control), roughly 10\% of control participants spend at least 5\% of their time in this state. Because these participants start from a poorer baseline, they exhibit larger absolute reductions--improvements that are inherently easier to achieve than those seen in the treatment arm.
 We summarize the main results below.
 
i) Roughly 10\% of the treated participants increase their time in phenotype 1 within the first month.
ii) Only a minority of patients, both improving and worsening, show time--varying changes in weights \(\alpha_{is}\); improvements, when present, typically require at least two weeks and then stabilize or reverse after four months in the absence of intervention.
iii) Continuous‑time mixed‑density models capture dynamic real‑time fluctuations that static or discretized analyses miss.
iv) Weight trajectories \(\alpha_{is}\) provide a clinically interpretable metric to monitor individual response in CGM‑based trials.

\subsection{Bivariate analysis}
We extend our univariate density analysis of clinical CGM data to the bivariate setting.  The ten‐day time windows considered here are identical to those in the main paper.

For each participant \(i\) and each ten‐day interval \(t\), we build the bivariate sample
\[
  (G_{it\ell}, S_{it\ell}),\quad \ell=1,\dots,n_{it},
\]
where \(G_{it\ell}\) is the raw glucose reading, and
\[
  S_{it\ell}
  = \frac{G_{it\ell} - G_{it,(\ell-1)}}{\Delta t}
\]
is its finite‐difference rate of change (first temporal derivative), with \(\Delta t\) denoting the CGM sampling interval.

\paragraph{Bivariate mixture representation.}
Within each window, we model the joint density of \((G,S)\) by the dynamic mixture
\[
  f_{i}(g,s;t)
  = \sum_{j=1}^{K}
      \alpha_{ij}(t)\,
      \mathcal{N}\bigl(\boldsymbol{\mu}_j,\boldsymbol{\Sigma}_j\bigr)(g,s),
\]
 where \(\boldsymbol{\mu}_j\in\mathbb{R}^2\) and \(\boldsymbol{\Sigma}_j\in\mathbb{R}^{2\times 2}\) are shared across all participants, and the weight trajectories \(\alpha_{ij}(t)\) capture subject‐specific temporal dynamics. Again, we fix $F=3$ and fit the global means ${\boldsymbol{\mu}_s}_{s=1}^3$ and covariance matrices ${\boldsymbol{\Sigma}_s}_{s=1}^3$ with the same three representative patients, see \cref{tab:global-statsc2}.

\begin{figure}[h!]
  \centering
  \includegraphics[width=0.32\textwidth]{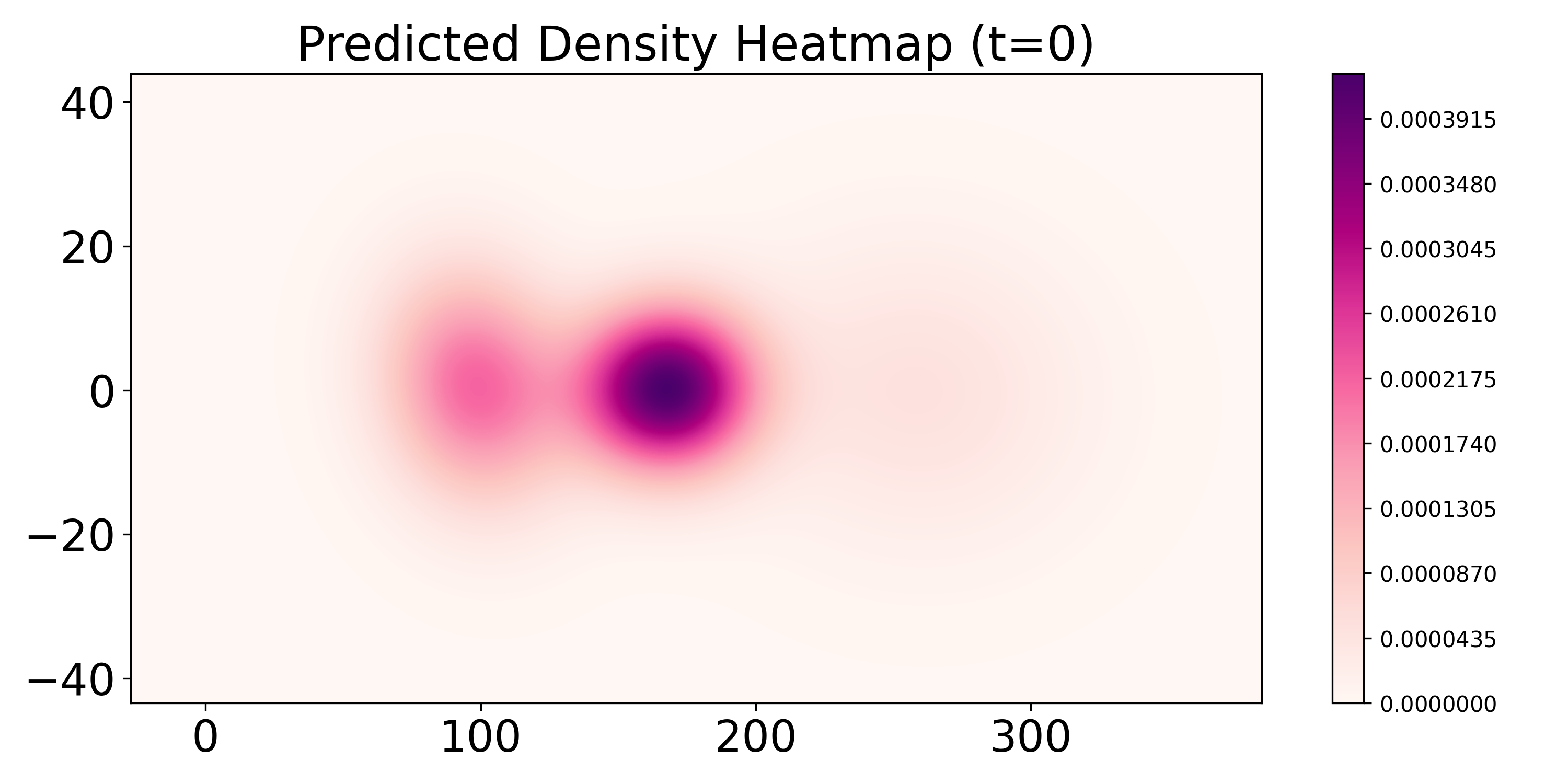}
  \includegraphics[width=0.32\textwidth]{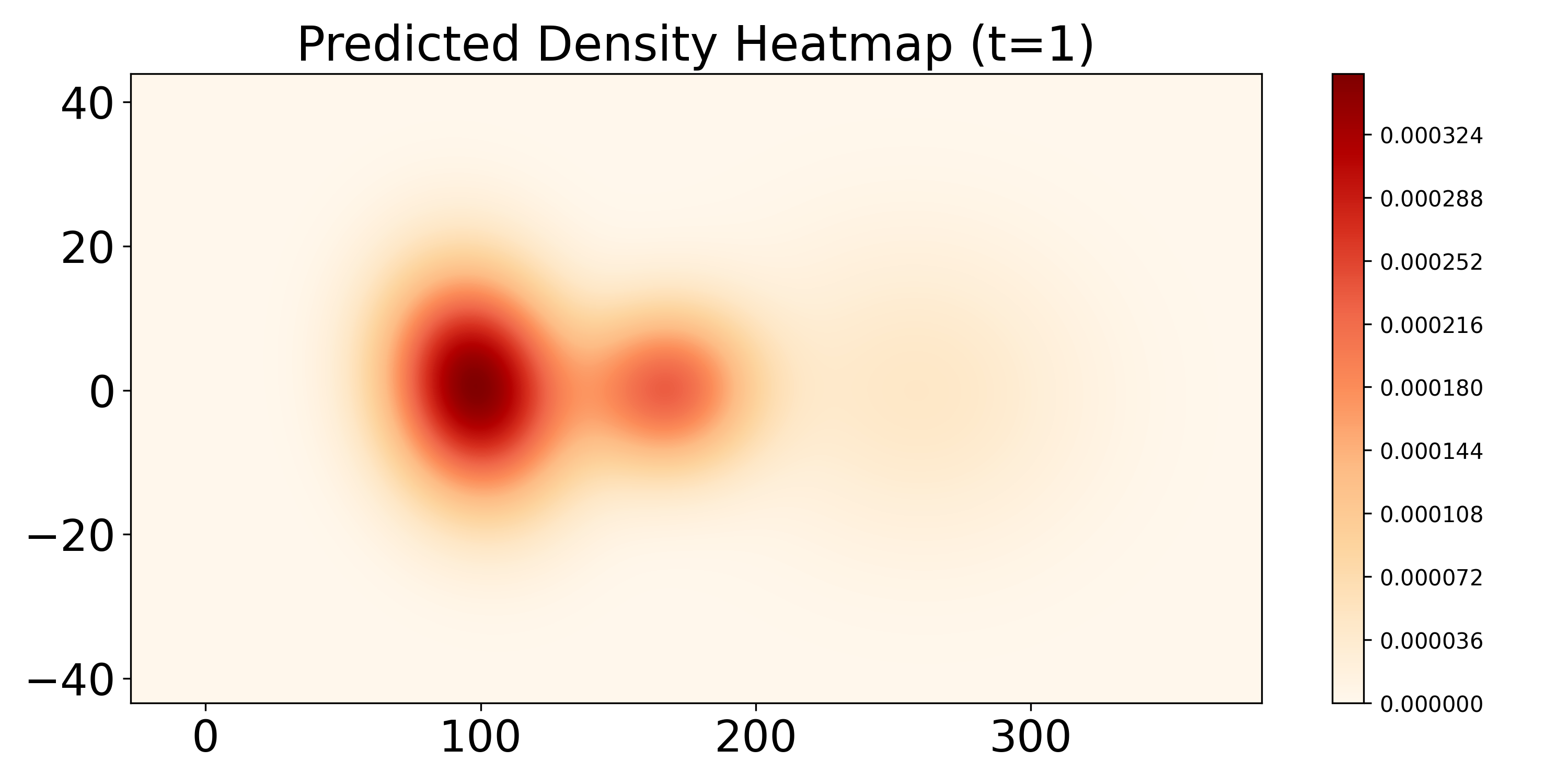}
  \includegraphics[width=0.32\textwidth]{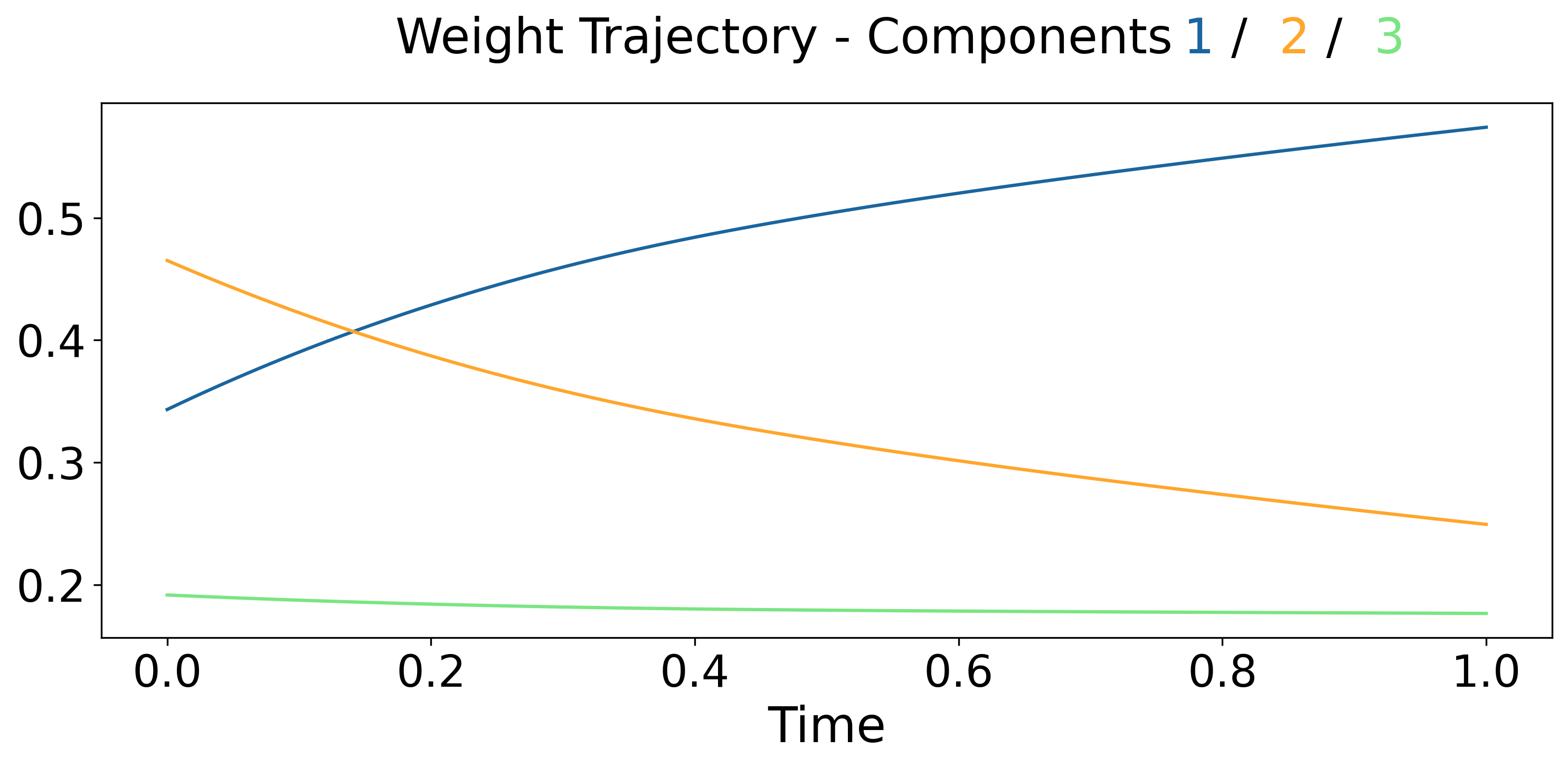}
  \\[1ex]
  \includegraphics[width=0.32\textwidth]{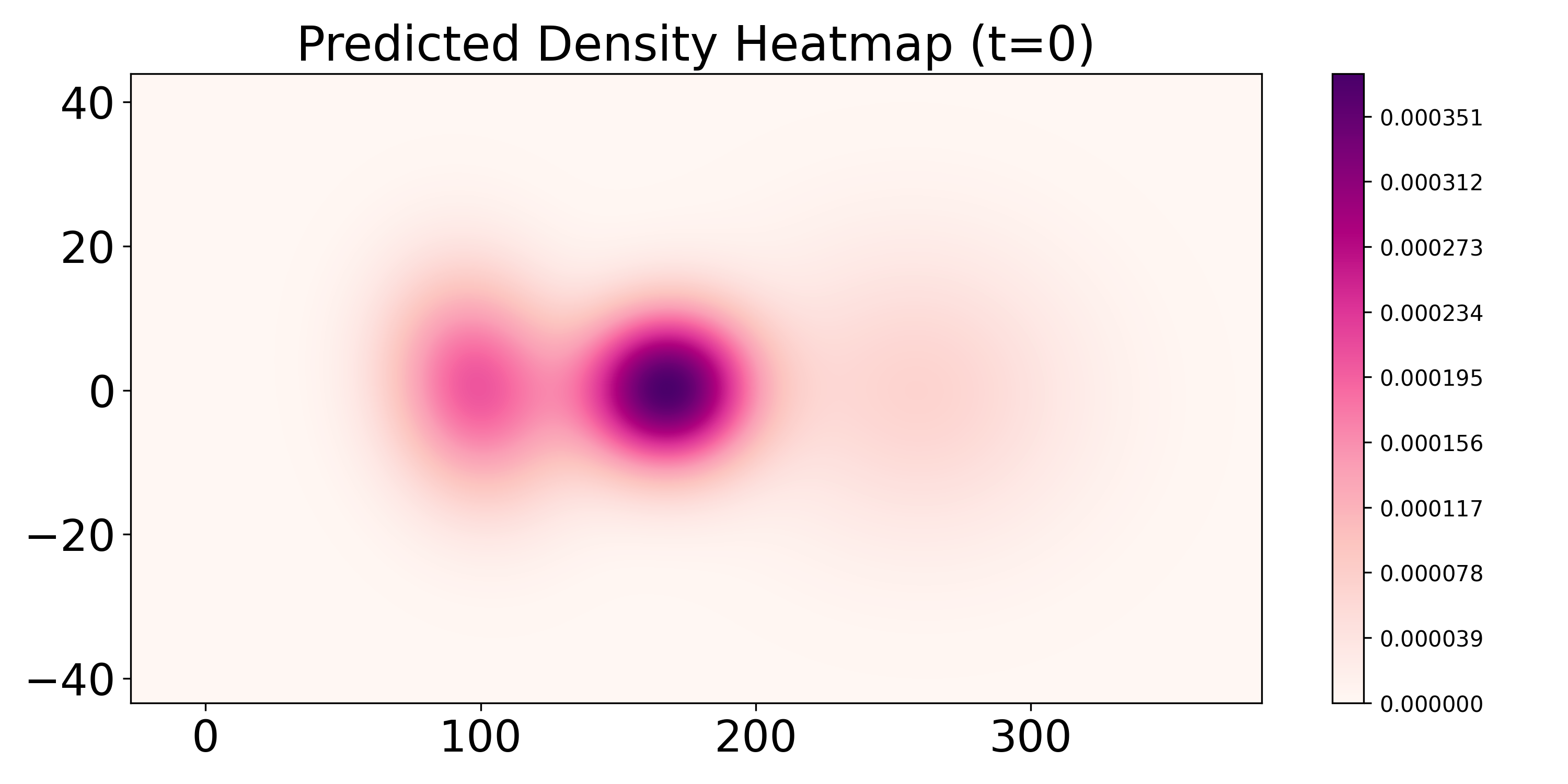}
  \includegraphics[width=0.32\textwidth]{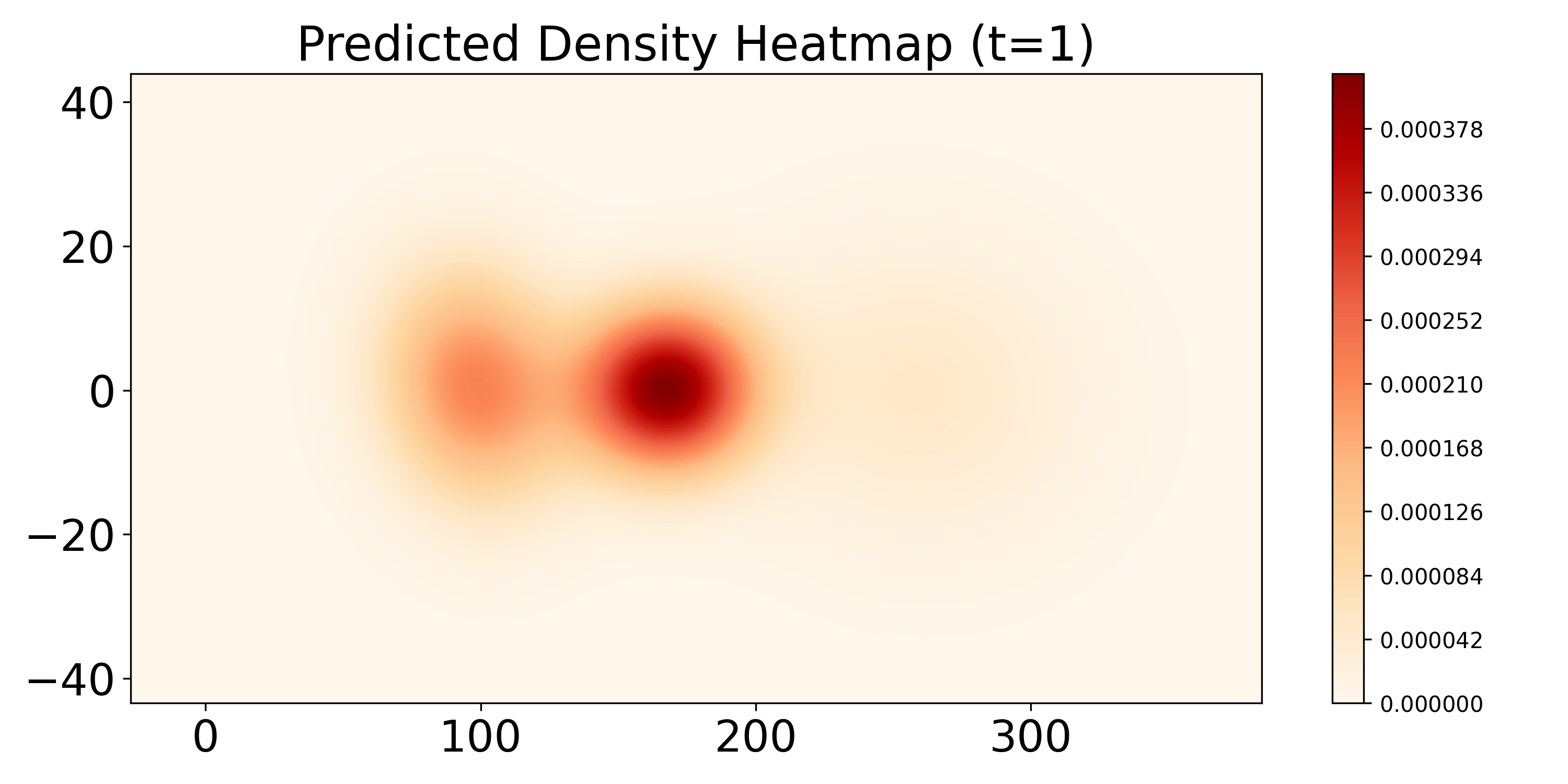}
  \includegraphics[width=0.32\textwidth]{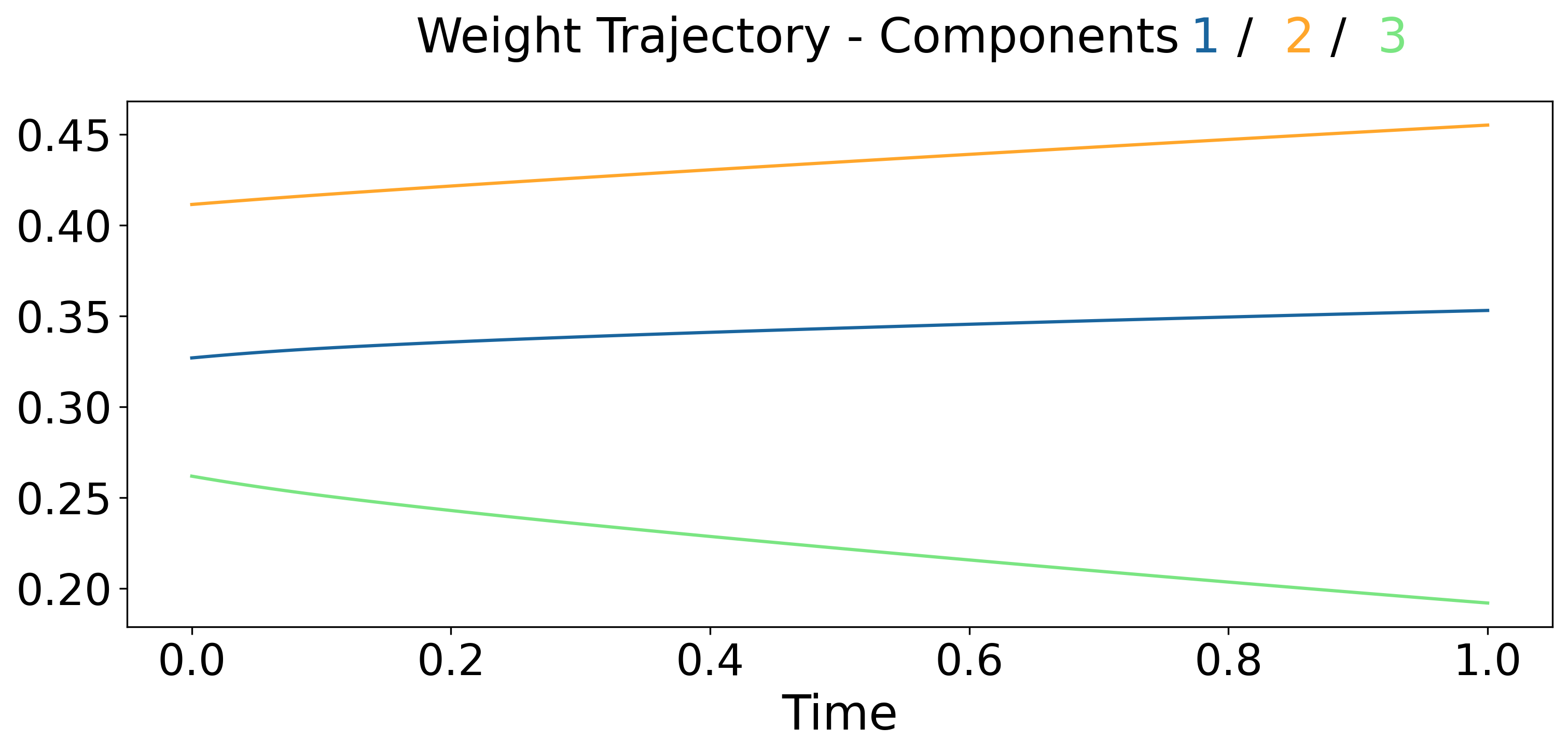}
  \\[1ex]
  \includegraphics[width=0.32\textwidth]{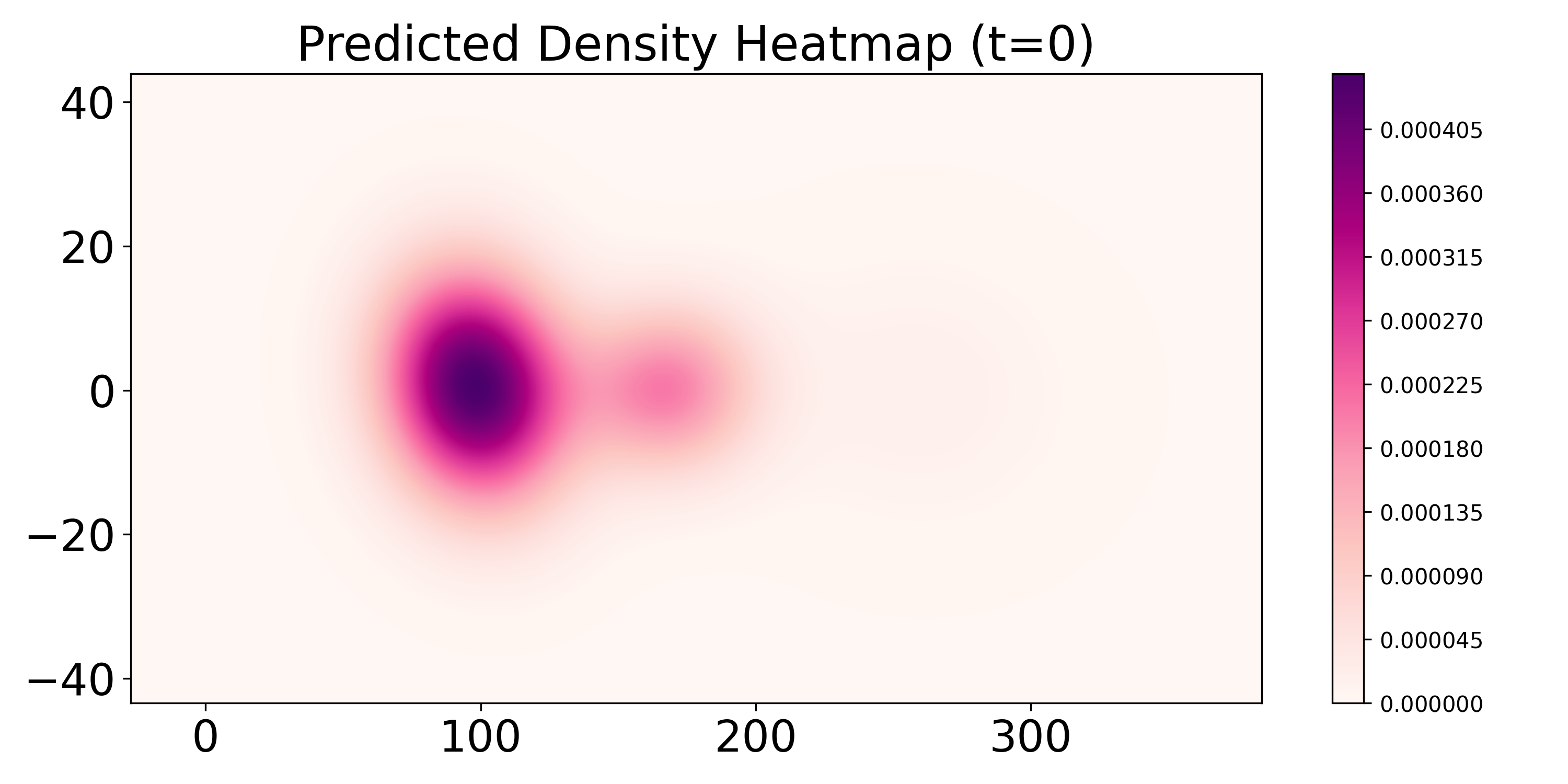}
  \includegraphics[width=0.32\textwidth]{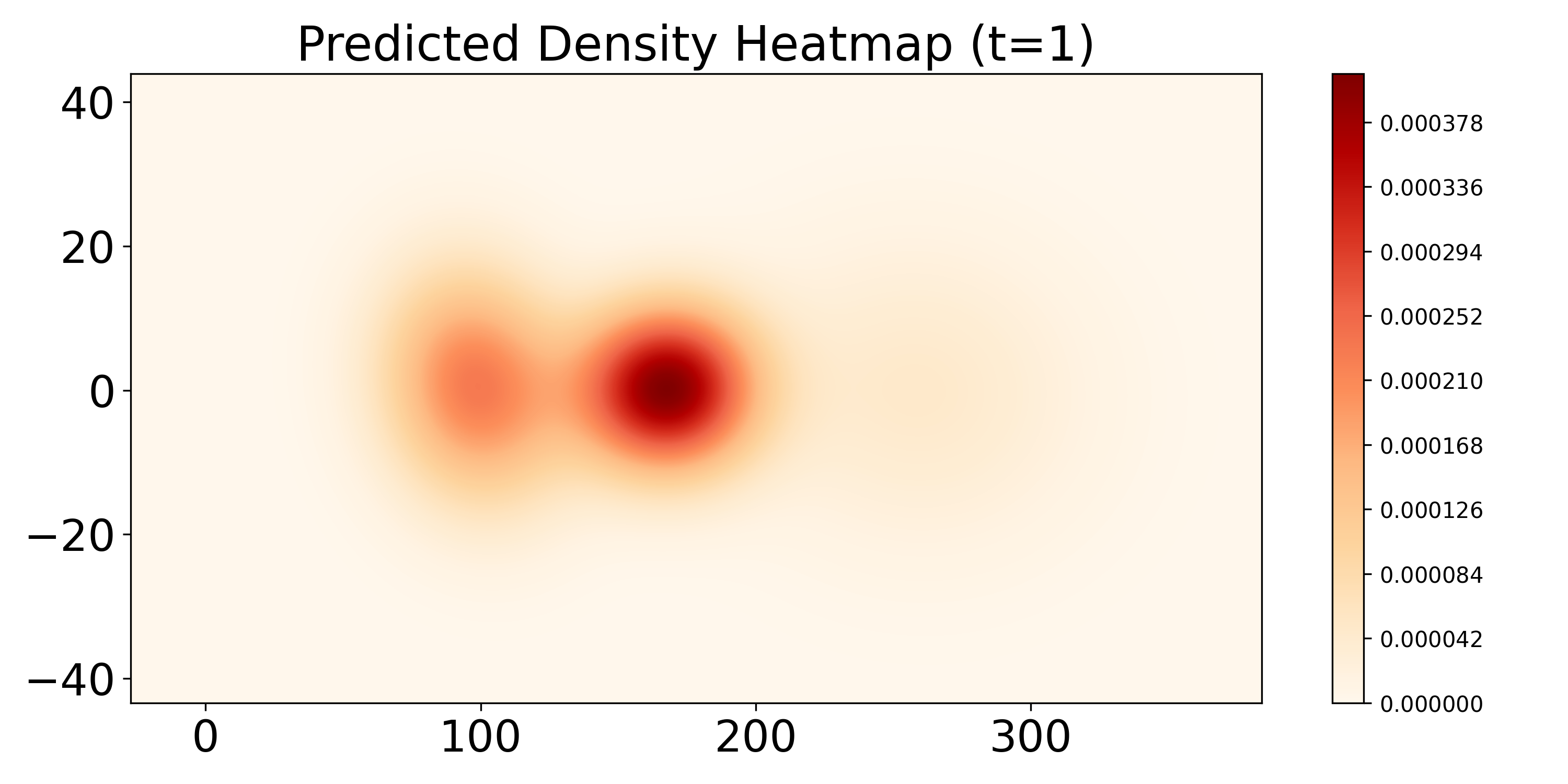}
  \includegraphics[width=0.32\textwidth]{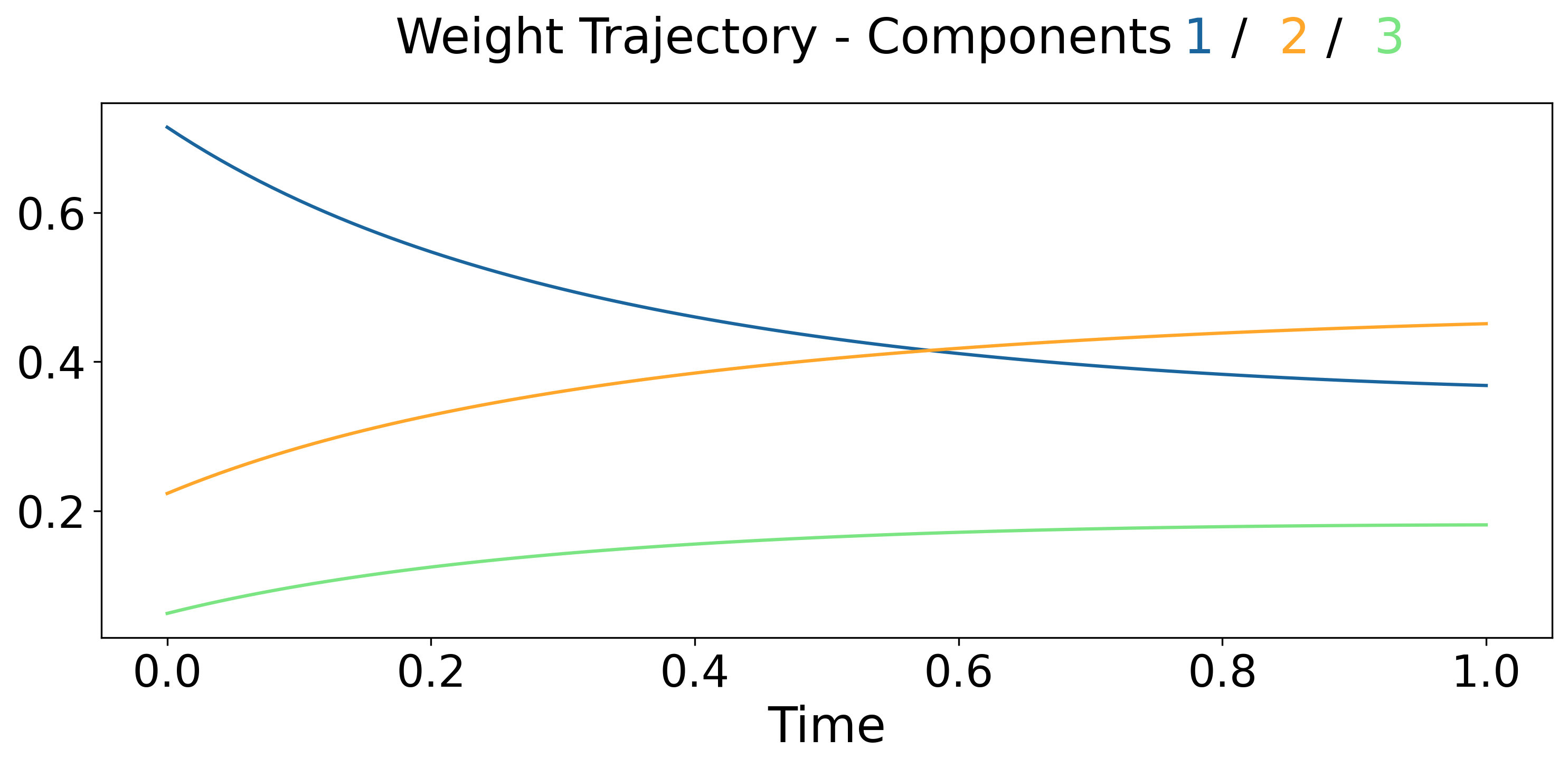}
  \caption{Three representative subjects (IDs 13, 62, and 377; one per row). Left: initial fitted bivariate density; Middle: final fitted bivariate density; Right: weight trajectories $\alpha_{s}(t)$ from the neural ODE model, whose endpoints match the middle-column fits.}
  \label{fig:traj2dexample}
\end{figure}

\begin{figure}[htbp]
  \centering
  \begin{tabular}{ccc}
    \includegraphics[width=0.3\textwidth]{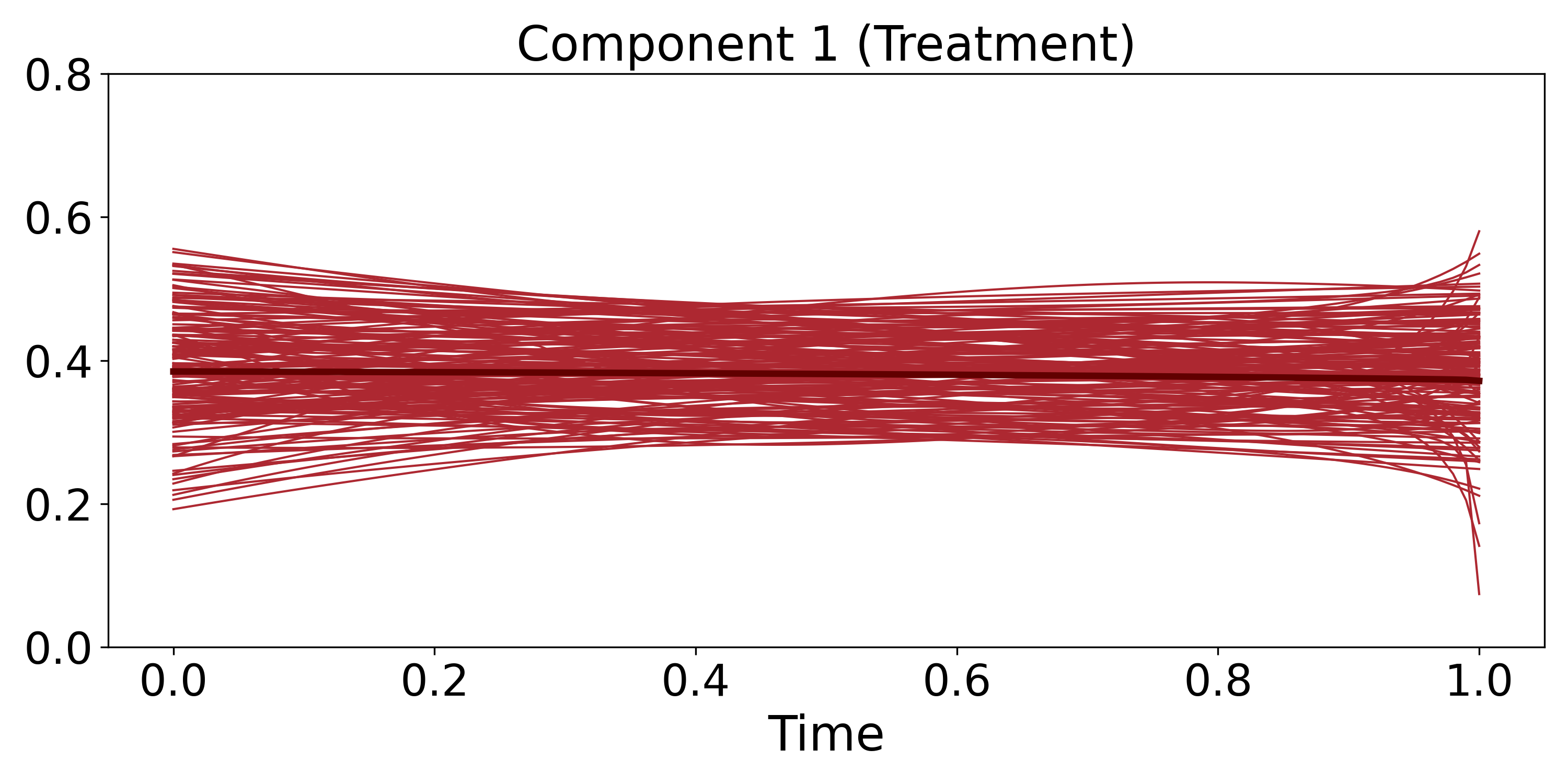} &
    \includegraphics[width=0.3\textwidth]{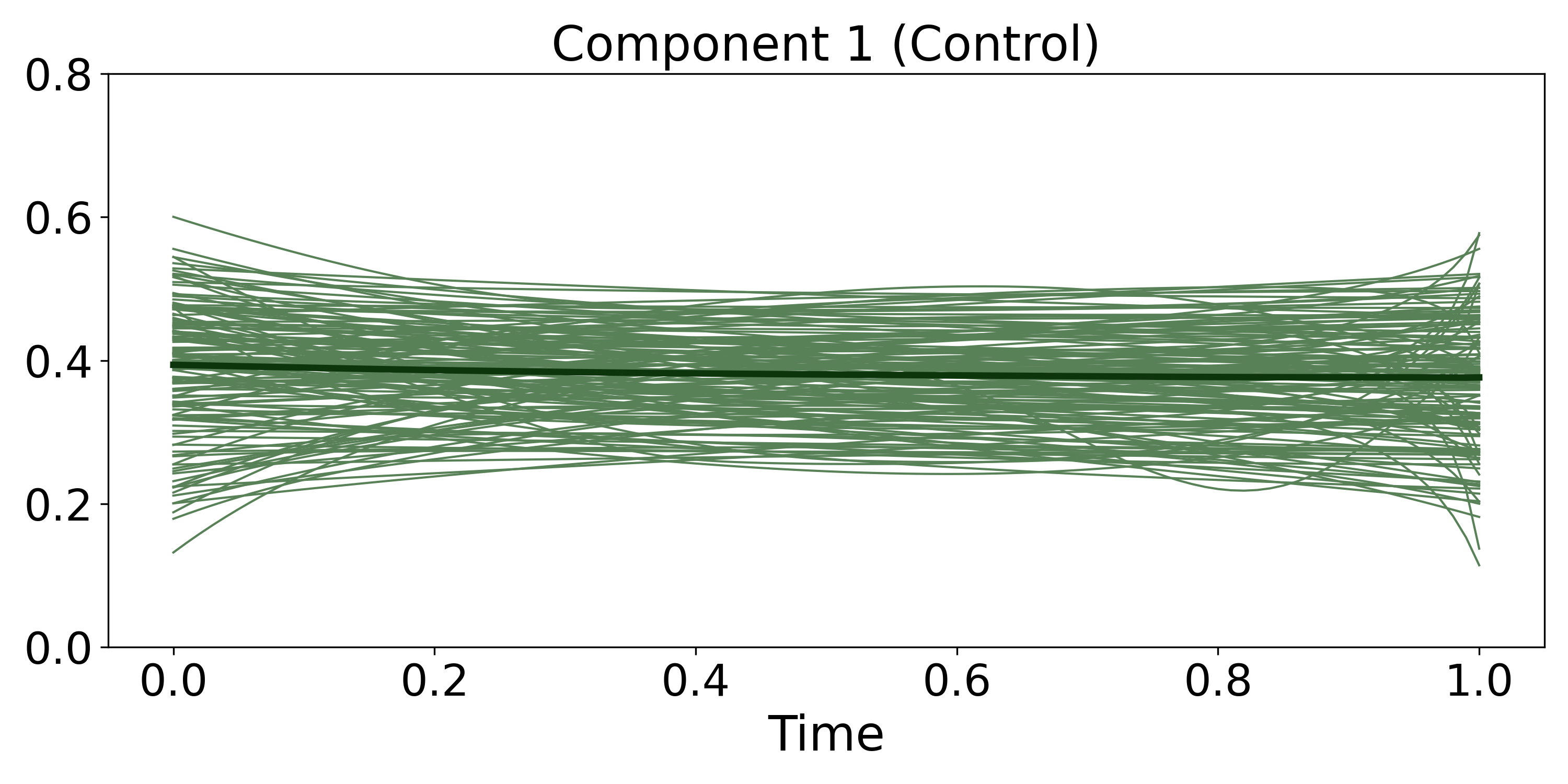} &
    \includegraphics[width=0.3\textwidth]{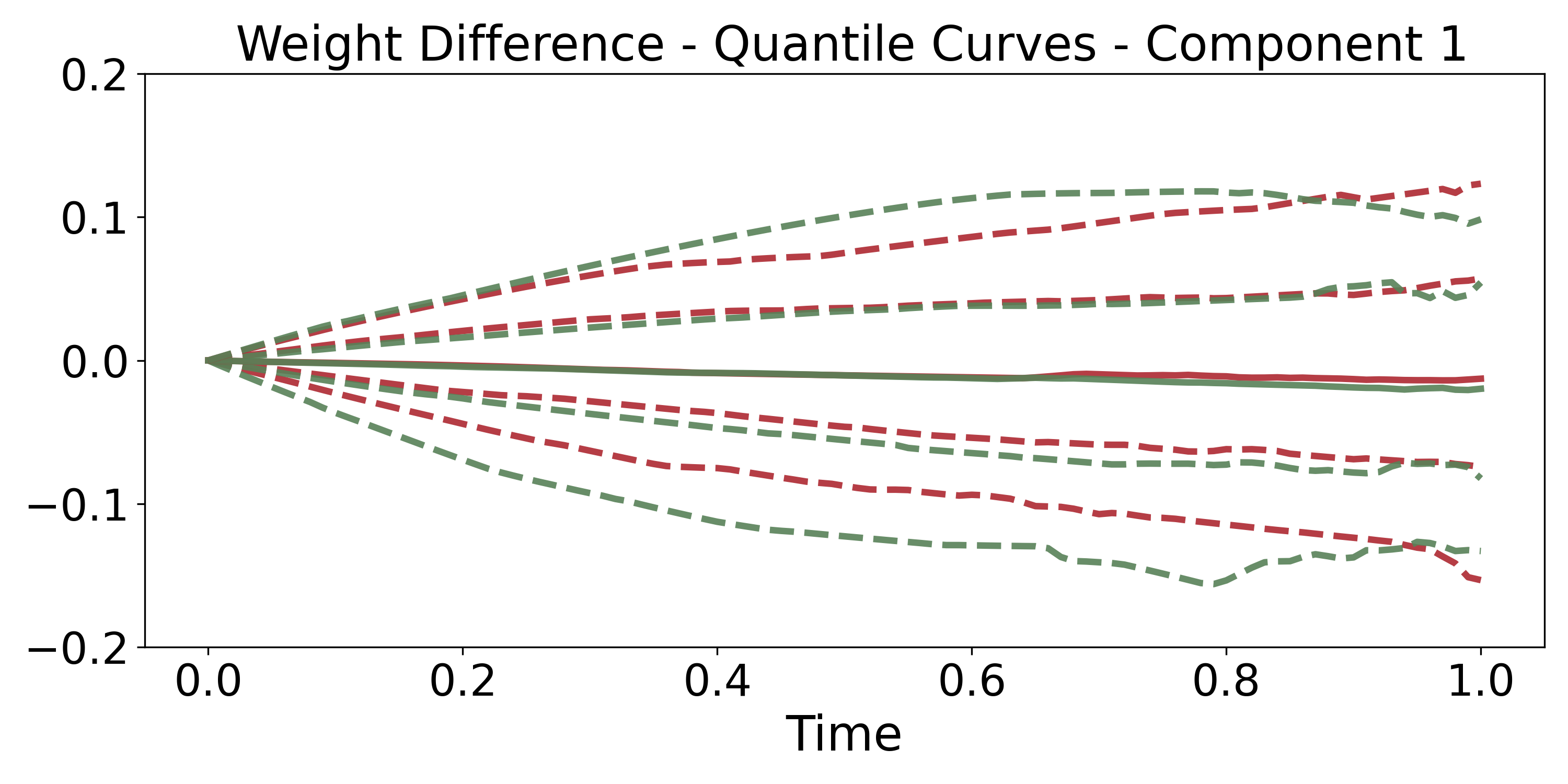} \\[1ex]
    \includegraphics[width=0.3\textwidth]{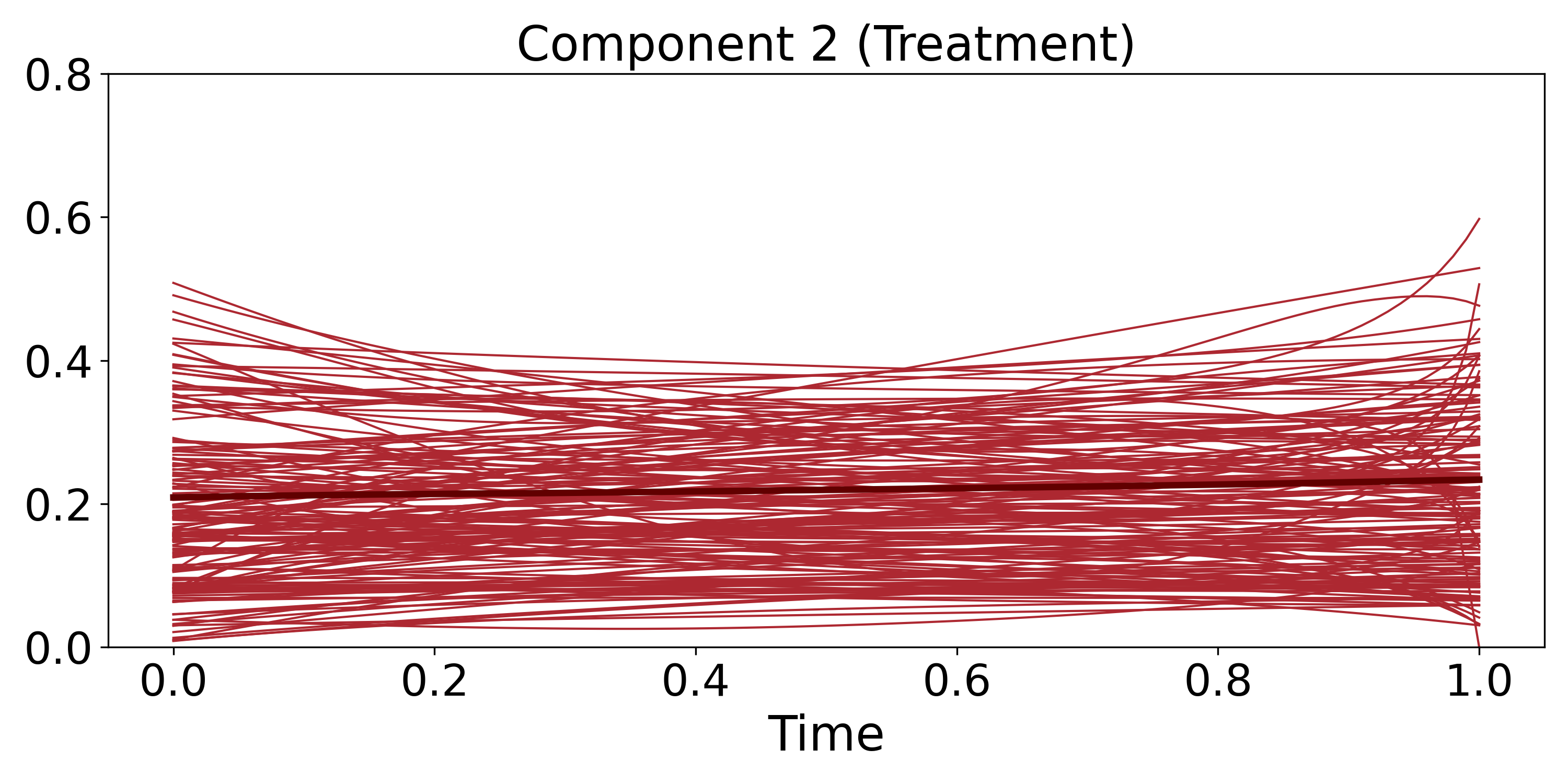} &
    \includegraphics[width=0.3\textwidth]{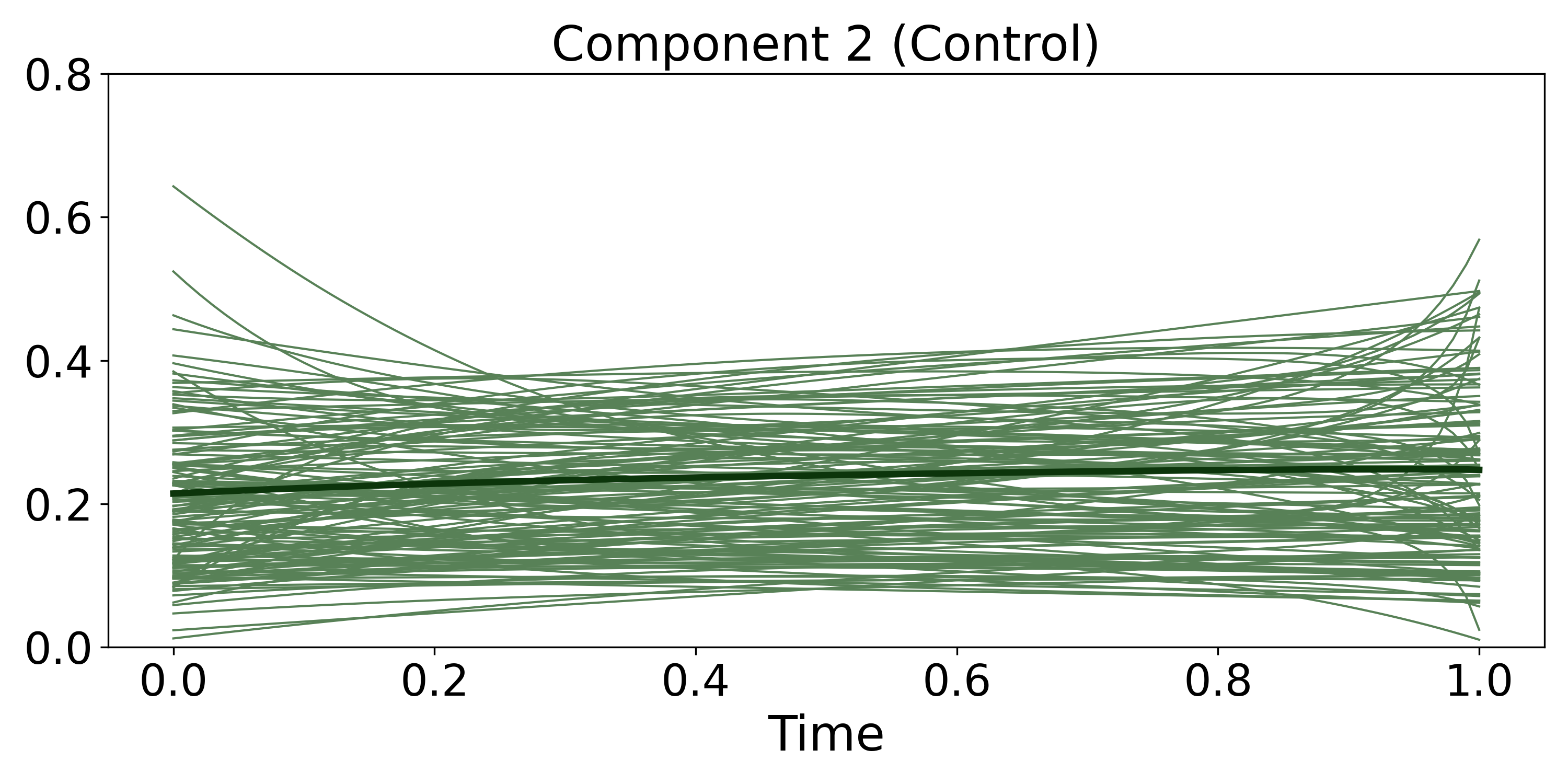} &
    \includegraphics[width=0.3\textwidth]{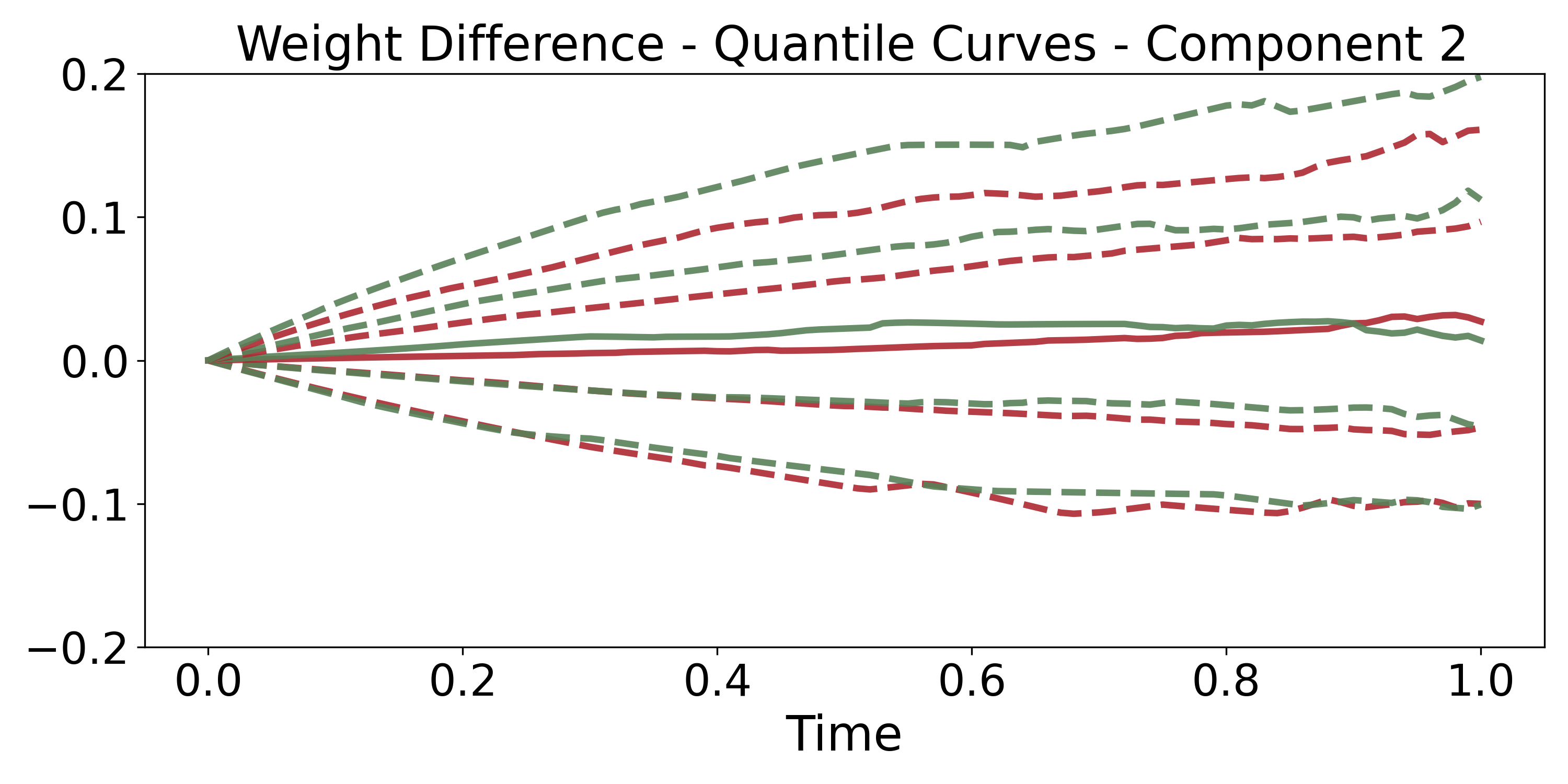} \\[1ex]
    \includegraphics[width=0.3\textwidth]{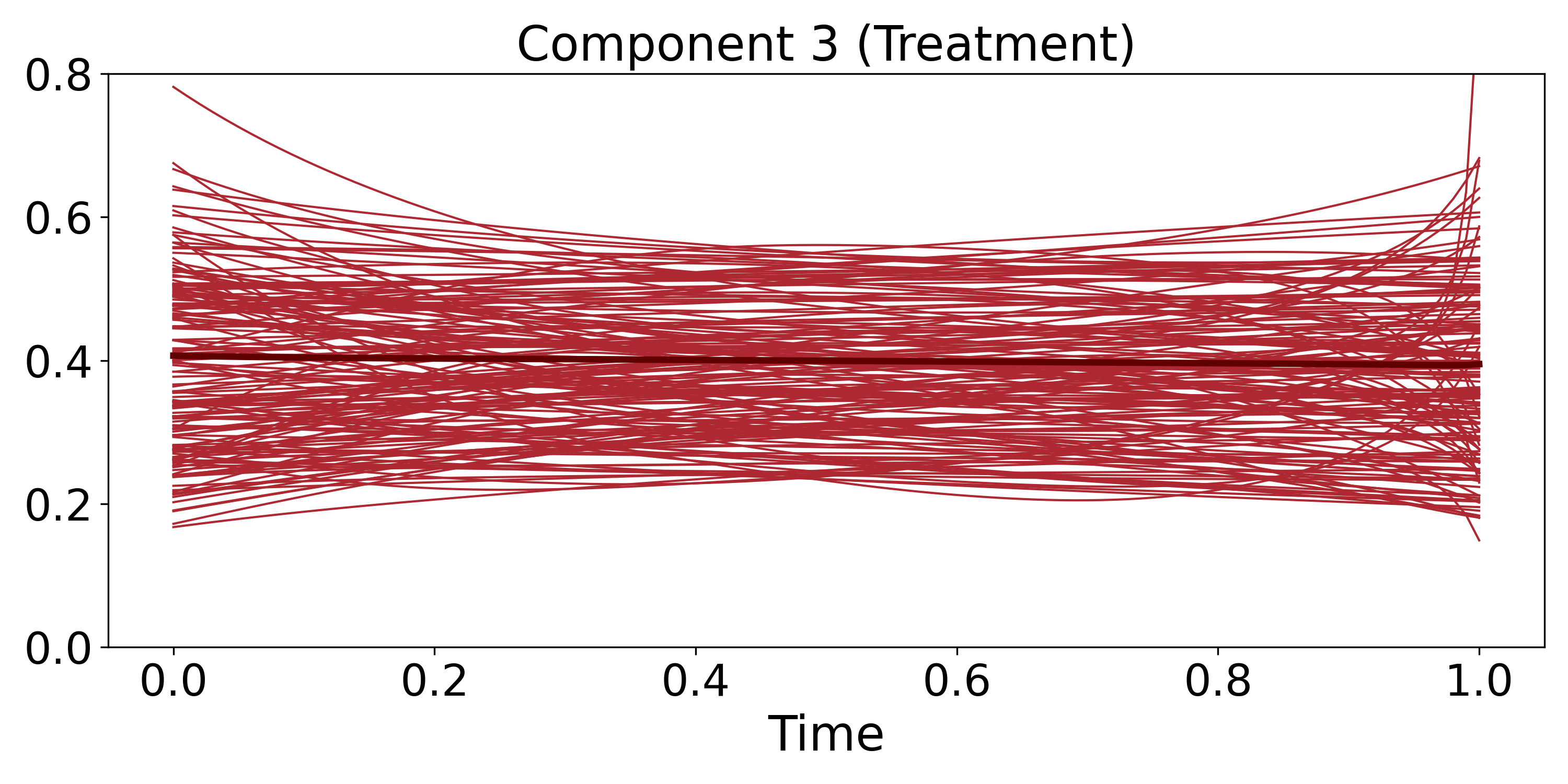} &
    \includegraphics[width=0.3\textwidth]{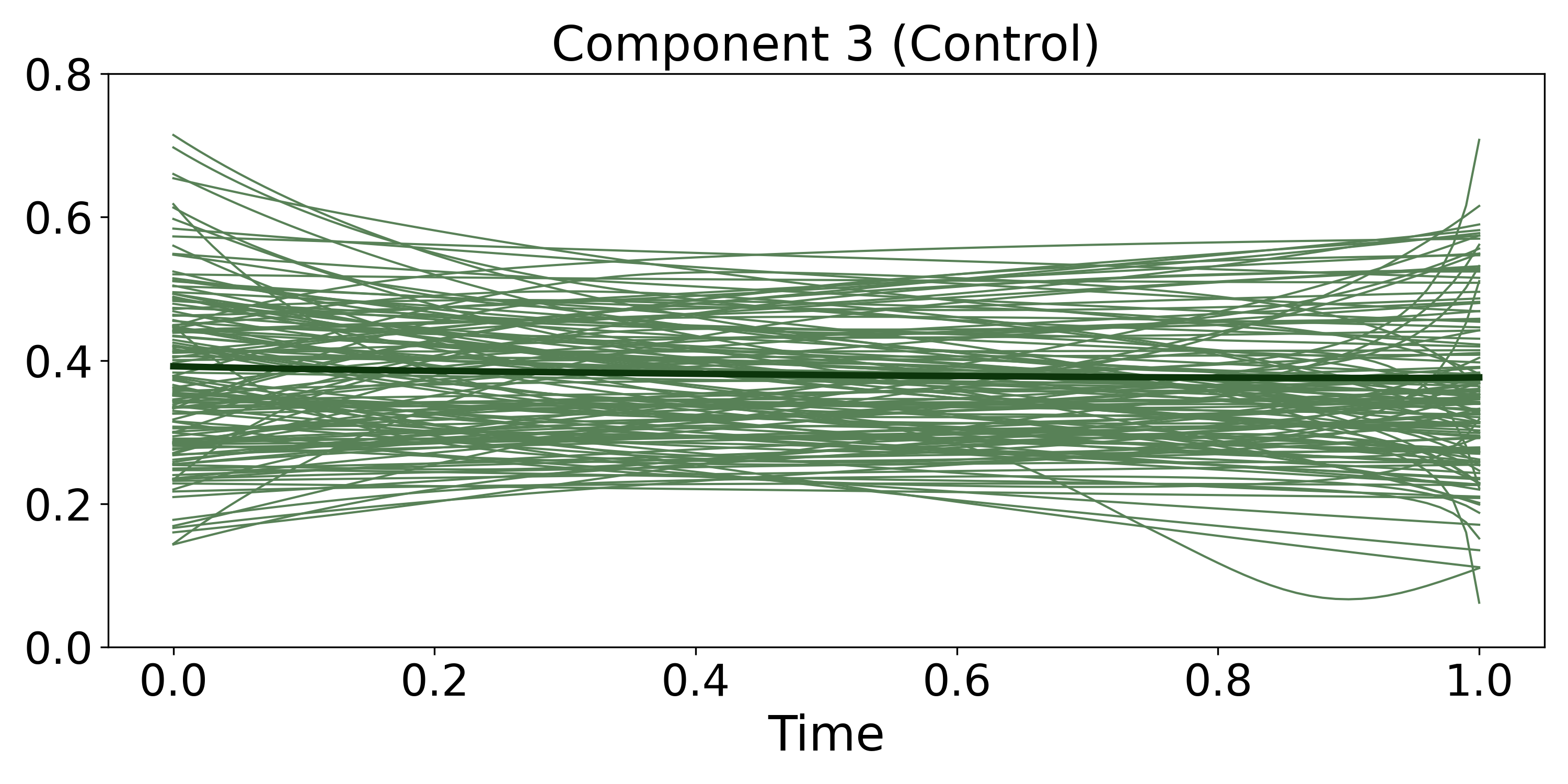} &
    \includegraphics[width=0.3\textwidth]{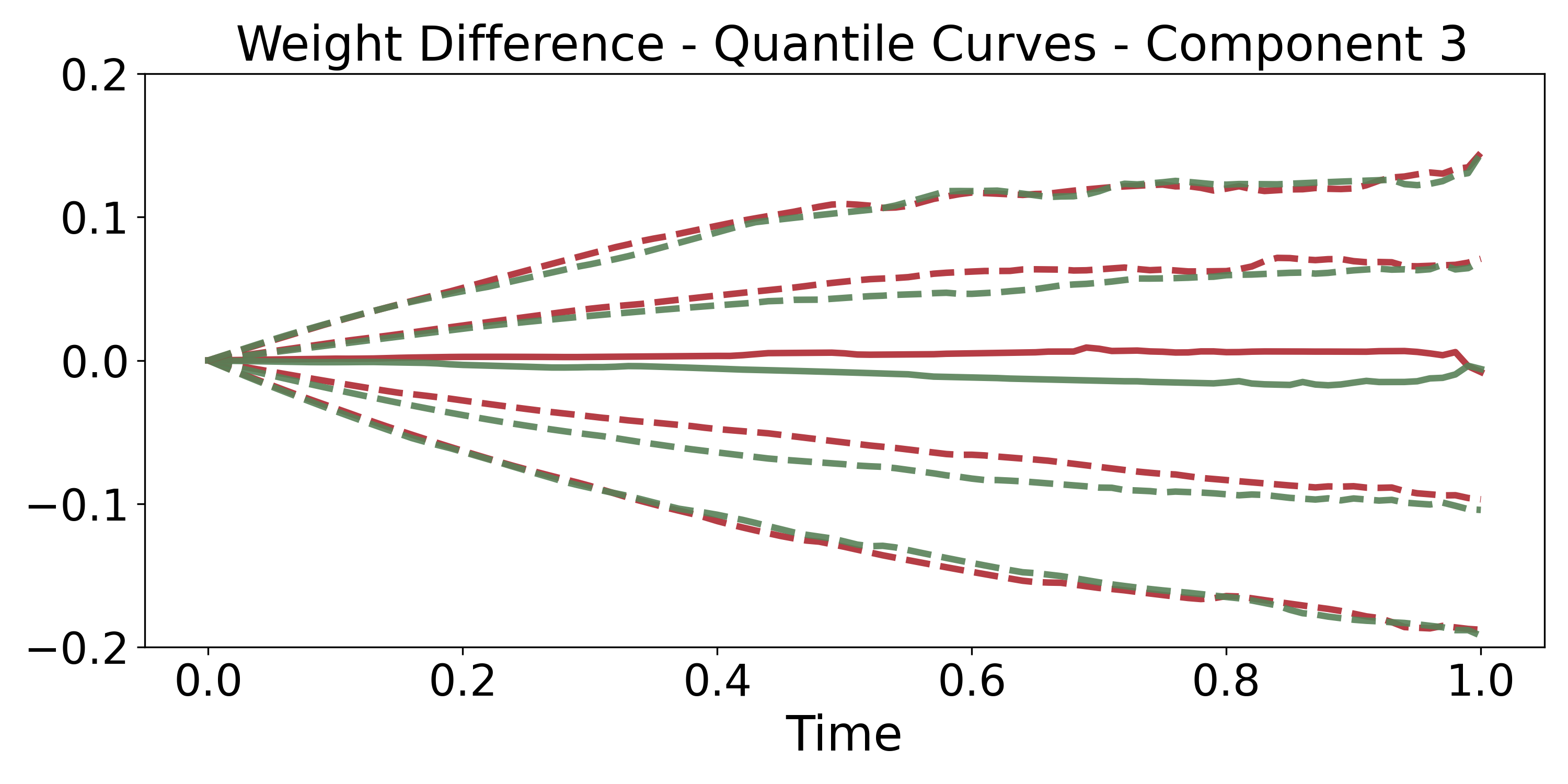}
  \end{tabular}
\caption{Component-wise weight dynamics on the full medical dataset after extending to a bivariate analysis by including the temporal derivative of glucose. Columns 1–2: individual curves $\alpha_s(t)$ (thin) and mean (thick) for treatment (left) and control (right). Column~3: $(0.1, 0.25, 0.5, 0.75, 0.9)$-quantile trajectories of $Z_{is}=\alpha_{is}(t)-\alpha_{is}(0)$ across subjects.}
  \label{fig:med-trajsd2}
\end{figure}

\Cref{fig:med-trajsd2} compares the observed versus predicted values of \(Z_i\); points lie close to the identity line, underscoring the benefit of incorporating temporal dynamics via the mixture weights.

The interpretation of the process $Z_{ij}(t) = \alpha_{ij}(t) - \alpha_{ij}(0)$ and the resulting conclusions are analogous to those in the main article.

\section{Discussion}
\label{sec:6}

We introduce an interpretable framework for estimating the dynamics of time-indexed probability distributions. In a comprehensive simulation study, the proposed method achieves a performance comparable to or exceeding state-of-the-art approaches, such as normalizing flows, while offering faster computation and new insights into temporal evolution. Theoretical guarantees further substantiate these empirical findings. In a clinical case study using CGM data, our approach reveals temporal patterns that conventional statistical methods and standard machine learning techniques fail to capture.

As future work, we plan to: (i) develop distributed, online variants of the algorithm to enable large‑scale, real‑time analysis (e.g., motivated by the need to analyze massive cohorts as \emph{All of Us} \cite{all2019all}); (ii) integrate our dynamic density estimates into a model‑free conformal prediction framework with finite‑sample validity for metric‑space–valued objects \cite{lugosi2024uncertainty}; (iii) extend the methods to vector‑valued or functional biomarkers in richly multivariate longitudinal datasets (e.g., UCI studies with $>50$ concurrent biomarkers); (iv) design new control systems driven by CGM insulin pumps based on this framework and Neural ODE control theory \cite{e_proposal_2017}; and (v) generalize the models to time series whose units of interest lie in a separable Hilbert space \cite{dubey2020functional}.

\section{Computational Details}

All experiments were executed on Ubuntu 20.04 LTS using eight NVIDIA GeForce RTX 3080 GPUs (10 GB each) on nodes with dual Intel Xeon Gold 6226R CPUs (2 × 32 cores @ 2.9 GHz), 384 GB RAM, and a 3.8 TB local SSD. The software environment was built with the following.
\begin{itemize}
  \item Python 3.12.9 (Conda)
  \item CUDA 12.9 (Driver 575.51.03)
  \item PyTorch 2.2.0 (with \texttt{torchdiffeq} 0.2.3, \texttt{torchcde} 0.2.5)
  \item NumPy 1.26.4, SciPy 1.14.1, scikit-learn 1.6.1
\end{itemize}

\section{Proofs}

\begin{proof}[Proof of \cref{prop:uniform_shared_dictionary}]
Fix $\varepsilon>0$. By uniform tightness, choose $R<\infty$ such that $\int_{\|x\|>R} f(x,t)\,\mathrm{d}x<\varepsilon/4$ for all $t$, and write $B_R=\{x\in\mathbb R^d:\|x\|\le R\}$.

Let $\varphi_\sigma$ be a Gaussian mollifier. Using the approximate-identity representation and the change of variables $h=\sigma u$,
\begin{align*}
\|f(\cdot,t)-f(\cdot,t)\ast\varphi_\sigma\|_{L^1}
&=\Big\|\int_{\mathbb R^d}\big(f(\cdot,t)-f(\cdot-h,t)\big)\,\varphi_\sigma(h)\,\mathrm{d}h\Big\|_{L^1}
\\
&\le \int \varphi_1(u)\,\|f(\cdot,t)-f(\cdot-\sigma u,t)\|_{L^1}\,\mathrm{d}u.
\end{align*}
Taking $\sup_t$ and using uniform translation–equicontinuity plus dominated convergence, we get
$\sup_{t}\|f(\cdot,t)-f(\cdot,t)\ast\varphi_\sigma\|_{L^1}<\varepsilon/4$ for all $\sigma$ small enough. Partition $B_R$ into hypercubes $\{C_s\}_{s=1}^K$ of mesh size $h$ and let $\mu_s$ be the centre of $C_s$. For each $t$, let
\[
\tilde\alpha_s(t)=\int_{C_s} f(y,t)\,\mathrm{d}y,\qquad
Z_t=\sum_{s=1}^K \tilde\alpha_s(t)=\int_{B_R} f(y,t)\,\mathrm{d}y\in[1-\varepsilon/4,\,1],
\]
and define $\alpha_s(t)=\tilde\alpha_s(t)/Z_t$, so that $\alpha(t)\in\Delta^{K-1}$. Writing $g_t=f(\cdot,t)\ast\varphi_\sigma$, we have
\[
\begin{aligned}
\Big\|g_t-\sum_{s=1}^K \tilde\alpha_s(t)\,\varphi_\sigma(\cdot-\mu_s)\Big\|_{L^1}
&\le \Big\|\int_{B_R^c} f(y,t)\,\varphi_\sigma(\cdot-y)\,\mathrm{d}y\Big\|_{L^1} \\
&\quad + \Big\|\sum_{s=1}^K \int_{C_s} f(y,t)\,\big(\varphi_\sigma(\cdot-y)-\varphi_\sigma(\cdot-\mu_s)\big)\,\mathrm{d}y\Big\|_{L^1}.
\end{aligned}
\]
The first term equals $1-Z_t\le \varepsilon/4$. Since $\varphi_\sigma$ is Lipschitz in $L^1$ with
$\|\varphi_\sigma(\cdot-y)-\varphi_\sigma(\cdot-\mu_s)\|_{L^1}
\le \|\nabla\varphi_\sigma\|_{L^1}\,\|y-\mu_s\|\le C_d\,h/\sigma$, we obtain
\[
\sup_t\Big\|g_t-\sum_{s=1}^K \tilde\alpha_s(t)\,\varphi_\sigma(\cdot-\mu_s)\Big\|_{L^1}
\le \frac{\varepsilon}{4}+C_d\,\frac{h}{\sigma}\,\sup_t\!\int_{B_R}\! f(y,t)\,\mathrm{d}y
\le \frac{\varepsilon}{4}+C_d\,\frac{h}{\sigma}.
\]
Moreover,
\[
\Big\|\sum_{s=1}^K \tilde\alpha_s(t)\,\varphi_\sigma(\cdot-\mu_s)
-\sum_{s=1}^K \alpha_s(t)\,\varphi_\sigma(\cdot-\mu_s)\Big\|_{L^1}
= |Z_t-1| \;\le\; \varepsilon/4.
\]
Finally, by the triangle inequality,
\[
\bigg\|\,f(\cdot,t)-\sum_{s=1}^K \alpha_s(t)\,\varphi_\sigma(\cdot-\mu_s)\,\bigg\|_{L^1}
\le \underbrace{\|f-g_t\|_{L^1}}_{\le \varepsilon/4}
+ \underbrace{\Big\|g_t-\sum_{s=1}^K \tilde\alpha_s(t)\,\varphi_\sigma(\cdot-\mu_s)\Big\|_{L^1}}_{\le \varepsilon/4+C_d h/\sigma}
+ \underbrace{\Big\|\sum \tilde\alpha_s\varphi_\sigma-\sum \alpha_s\varphi_\sigma\Big\|_{L^1}}_{\le \varepsilon/4}.
\]
Choosing $h\le \sigma\,\varepsilon/(4C_d)$ gives the claim.

If, in addition, $t\mapsto f(\cdot,t)$ is uniformly $L^1$–continuous, then
$t\mapsto \tilde\alpha_s(t)=\int_{C_s} f(y,t)\,\mathrm{d}y$ is continuous for each $s$. Since $Z_t=\sum_s \tilde\alpha_s(t)\in[1-\varepsilon/4,1]$ is continuous and bounded away from $0$, the normalized weights $\alpha_s(t)=\tilde\alpha_s(t)/Z_t$ are continuous. Hence $t\mapsto\alpha(t)$ is continuous.
\end{proof}

\begin{proof}[Proof of \cref{thm:MMD_rate}]
 \textbf{Step 1: Pointwise concentration.}
Since $k(x,\cdot)\in\mathcal H$ is bounded by $1$, the centred feature map
\[
  Z_{t,i}
  = k(X_{t,i},\cdot)
    - \mathbb{E}\bigl[k(X_{t,i},\cdot)\bigr]
\]
is sub‐Gaussian in $\mathcal H$. By Pinelis’ inequality
\[
  \|F^{\theta}_{t,n_t}-F^{\theta}_t\|_{\mathcal H}
  = O_p(n_t^{-1/2}).
\]
In regime (S), local singularity of order $n_t^{-1/2}$ yields $O_p(n_t^{-1/4})$. For any bounded kernel with $\sup_x k(x,x)\le K$ we have the sharp inequality
$\mathrm{MMD}(P,Q)\le\sqrt{2K}\,\|P-Q\|_{\mathrm{TV}}$, 
so minimax bounds in total variation automatically carry over to MMD.  We must note that, when two mixture components become indistinguishable,
the optimal rate in TV (and Hellinger) deteriorates to $n^{-1/4}$.  

\textbf{Step 2: Uniform control.}
Cover $[0,1]$ by an $n_*^{-1}$‐net of size $O(n_*)$. Lipschitz continuity of $F^{\theta}_t$ in
$\mathcal H$ implies a sub‐Gaussian process indexed by this net. Applying Massart’s maximal
inequality and a Borell–TIS bound gives the
uniform $O_p(\cdot)$ rates above.   
\end{proof}

%====================================================================

%\begin{assumption}[Basic Bounds]\label{ass:basic}
%There exist constants $L,\delta,M_1,m_2,M_2>0$ such that, for each %component $s=1,\dots,K$ and every $t\in[0,1]$:
%\begin{enumerate}[nosep]
  %\item The map $\theta\mapsto f_{\theta}(x,t)$ is $L$–Lipschitz %in~$\theta$.
  %\item The mixing weight satisfies $\alpha_s(t)\in[\delta,\,1-\delta]$.
  %\item The component mean satisfies $|m_s(t)|\le M_1$.
  %\item The covariance matrix satisfies
  %  \[
     % m_2\,I_d \;\preceq\; \Sigma_s(t)\;\preceq\; M_2\,I_d.
 %   \]
%\end{enumerate}
%\end{assumption}

%==============================================================
%  Uniform Weak Continuity and Uniform Donsker Theorem
%==============================================================

\begin{proof}[Proof of \cref{prop:comparison}]
We analyze the three terms in the MMD\(^2\) decomposition separately, in the joint space \(\mathbb{R}^{d(m+1)}\):

\medskip
\noindent
\textbf{(1) Mixture–Mixture Term:}
\[
\sum_{s=1}^K \sum_{\ell=1}^K \alpha_s \alpha_\ell \int \int k(x, y)\, \mathcal{N}(x \mid m_s, \Sigma_s)\, \mathcal{N}(y \mid m_\ell, \Sigma_\ell)\, dx\,\mathrm{d}y
\]
This expression has a closed-form formula due to the product of Gaussians and the Gaussian kernel. Each pair \((k, \ell)\) requires:
\begin{itemize}
    \item Computing a determinant and inverse of a \(d(m+1) \times d(m+1)\) matrix, each of cost \( \mathcal{O}((d(m+1))^3) \).
    \item Evaluating a quadratic form of cost \( \mathcal{O}((d(m+1))^2) \), which is negligible compared to the matrix inverse.
\end{itemize}
There are \(K^2\) such terms.

Thus, the total cost is:
\[
\mathcal{O}(K^2 (d(m+1))^3).
\]

\medskip
\noindent
\textbf{(2) Sample–Mixture Term:}
\[
\sum_{j=1}^n \sum_{s=1}^K \alpha_s\, \int k(X_j, y)\, \mathcal{N}(y \mid m_s, \Sigma_s)\,\mathrm{d}y = \sum_{j,s} \alpha_s\, \mathcal{N}(X_j \mid m_s, \Sigma_s + \Gamma)
\]
Each evaluation is a multivariate normal PDF at a single point:
\begin{itemize}
    \item Cost of computing the density: \( \mathcal{O}((d(m+1))^2) \).
\end{itemize}
There are \(nK\) such evaluations.

Thus, the total cost is:
\[
\mathcal{O}(nK (d(m+1))^2).
\]

\medskip
\noindent
\textbf{(3) Sample–Sample Term:}
\[
\frac{1}{n^2} \sum_{j=1}^n \sum_{r=1}^n k(X_j, X_r)
\]
Each kernel evaluation involves computing a squared Euclidean norm in \(\mathbb{R}^{d(m+1)}\):
\begin{itemize}
    \item Cost per evaluation: \( \mathcal{O}(d(m+1)) \)
\end{itemize}
There are \(n^2\) such terms.

Thus, the total cost is:
\[
\mathcal{O}(n^2 d(m+1)).
\]

\medskip
\noindent
 Adding up the three costs yields the total complexity:
\[
\mathcal{O}\left( K^2 (d(m+1))^3 + nK (d(m+1))^2 + n^2 d(m+1) \right).
\]
\end{proof}

\clearpage

\bibliography{bibliography.bib}

\begin{thebibliography}{57}
\providecommand{\natexlab}[1]{#1}
\providecommand{\url}[1]{\texttt{#1}}
\expandafter\ifx\csname urlstyle\endcsname\relax
  \providecommand{\doi}[1]{doi: #1}\else
  \providecommand{\doi}{doi: \begingroup \urlstyle{rm}\Url}\fi

\bibitem[Alquier and Gerber(2024)]{alquier2024mmd}
P.~Alquier and M.~Gerber.
\newblock Universal robust regression via maximum mean discrepancy.
\newblock \emph{Biometrika}, 111\penalty0 (1):\penalty0 71--92, 2024.

\bibitem[Alquier et~al.(2023)Alquier, Ch{\'e}rief-Abdellatif, Derumigny, and
  Fermanian]{alquier2023copulammd}
P.~Alquier, B.-E. Ch{\'e}rief-Abdellatif, A.~Derumigny, and J.-D. Fermanian.
\newblock Estimation of copulas via maximum mean discrepancy.
\newblock \emph{JASA}, 118\penalty0 (543):\penalty0 1997--2012, 2023.

\bibitem[{\'A}lvarez-L{\'o}pez et~al.(2024){\'A}lvarez-L{\'o}pez, Geshkovski,
  and Ruiz-Balet]{alvarezlopez2025}
Antonio {\'A}lvarez-L{\'o}pez, Borjan Geshkovski, and Dom{\`e}nec Ruiz-Balet.
\newblock Constructive approximate transport maps with normalizing flows.
\newblock \emph{Applied Mathematics \& Optimization}, 2024.

\bibitem[Battelino et~al.(2023)Battelino, Alexander, Amiel, Arreaza-Rubin,
  Beck, Bergenstal, Buckingham, Carroll, Ceriello, Chow,
  et~al.]{battelino2023continuous}
Tadej Battelino, Charles~M Alexander, Stephanie~A Amiel, Guillermo
  Arreaza-Rubin, Roy~W Beck, Richard~M Bergenstal, Bruce~A Buckingham, James
  Carroll, Antonio Ceriello, Elaine Chow, et~al.
\newblock Continuous glucose monitoring and metrics for clinical trials: an
  international consensus statement.
\newblock \emph{The lancet Diabetes \& endocrinology}, 11\penalty0
  (1):\penalty0 42--57, 2023.

\bibitem[Bengio et~al.(2017)Bengio, Goodfellow, Courville,
  et~al.]{bengio2017deep}
Yoshua Bengio, Ian Goodfellow, Aaron Courville, et~al.
\newblock \emph{Deep learning}, volume~1.
\newblock MIT press Cambridge, MA, USA, 2017.

\bibitem[Bertsimas et~al.(2020)Bertsimas, Pauphilet, and
  Van~Parys]{bertsimas2020sparse}
Dimitris Bertsimas, Jean Pauphilet, and Bart Van~Parys.
\newblock Sparse regression.
\newblock \emph{Statistical Science}, 35\penalty0 (4):\penalty0 555--578, 2020.

\bibitem[Chac{\'o}n and Duong(2018)]{chacon2018multivariate}
Jos{\'e}~E Chac{\'o}n and Tarn Duong.
\newblock \emph{Multivariate kernel smoothing and its applications}.
\newblock CRC Press, 2018.

\bibitem[Chen et~al.(2018)Chen, Rubanova, Bettencourt, and
  Duvenaud]{chen_neural_2018}
Ricky T.~Q. Chen, Yulia Rubanova, Jesse Bettencourt, and David~K Duvenaud.
\newblock Neural {Ordinary} {Differential} {Equations}.
\newblock In \emph{Advances in {Neural} {Information} {Processing} {Systems}},
  volume~31. Curran Associates, Inc., 2018.

\bibitem[Ch{\'e}rief-Abdellatif and
  Alquier(2022)]{CheriefAbdellatifAlquier2022}
B.-E. Ch{\'e}rief-Abdellatif and P.~Alquier.
\newblock Finite-sample properties of parametric mmd estimation: Robustness to
  misspecification and dependence.
\newblock \emph{Bernoulli}, 28\penalty0 (1):\penalty0 181--213, 2022.

\bibitem[Cutler and Breiman(1994)]{cutler1994archetypal}
Adele Cutler and Leo Breiman.
\newblock Archetypal analysis.
\newblock \emph{Technometrics}, 36\penalty0 (4):\penalty0 338--347, 1994.

\bibitem[Dubey and M{\"u}ller(2020)]{dubey2020functional}
Paromita Dubey and Hans-Georg M{\"u}ller.
\newblock Functional models for time-varying random objects.
\newblock \emph{Journal of the Royal Statistical Society Series B: Statistical
  Methodology}, 82\penalty0 (2):\penalty0 275--327, 2020.

\bibitem[E(2017)]{e_proposal_2017}
Weinan E.
\newblock A {Proposal} on {Machine} {Learning} via {Dynamical} {Systems}.
\newblock \emph{Communications in Mathematics and Statistics}, 5\penalty0
  (1):\penalty0 1--11, March 2017.
\newblock ISSN 2194-671X.
\newblock \doi{10.1007/s40304-017-0103-z}.

\bibitem[Gao et~al.(2021)Gao, Liu, Zhang, Han, Liu, Niu, and
  Sugiyama]{gao2021mmd}
Ruize Gao, Feng Liu, Jingfeng Zhang, Bo~Han, Tongliang Liu, Gang Niu, and
  Masashi Sugiyama.
\newblock Maximum mean discrepancy test is aware of adversarial attacks.
\newblock In Marina Meila and Tong Zhang, editors, \emph{Proceedings of the
  38th International Conference on Machine Learning (ICML 2021)}, pages
  3564--3575. ML Research Press, July 2021.

\bibitem[Garreau et~al.(2017)Garreau, Jitkrittum, and
  Kanagawa]{garreau2017large}
Damien Garreau, Wittawat Jitkrittum, and Motonobu Kanagawa.
\newblock Large sample analysis of the median heuristic.
\newblock \emph{arXiv preprint arXiv:1707.07269}, 2017.

\bibitem[Ghosal and Matabuena(2024)]{ghosal2024multivariate}
Rahul Ghosal and Marcos Matabuena.
\newblock Multivariate scalar on multidimensional distribution regression with
  application to modeling the association between physical activity and
  cognitive functions.
\newblock \emph{Biometrical Journal}, 66\penalty0 (7):\penalty0 e202400042,
  2024.

\bibitem[Ghosal et~al.(2023)Ghosal, Varma, Volfson, Hillel, Urbanek, Hausdorff,
  Watts, and Zipunnikov]{ghosal2023distributional}
Rahul Ghosal, Vijay~R Varma, Dmitri Volfson, Inbar Hillel, Jacek Urbanek,
  Jeffrey~M Hausdorff, Amber Watts, and Vadim Zipunnikov.
\newblock Distributional data analysis via quantile functions and its
  application to modeling digital biomarkers of gait in alzheimer’s disease.
\newblock \emph{Biostatistics}, 24\penalty0 (3):\penalty0 539--561, 2023.

\bibitem[Ghosal et~al.(2025)Ghosal, Ghosh, Schrack, and
  Zipunnikov]{ghosal2025distributional}
Rahul Ghosal, Sujit~K Ghosh, Jennifer~A Schrack, and Vadim Zipunnikov.
\newblock Distributional outcome regression via quantile functions and its
  application to modelling continuously monitored heart rate and physical
  activity.
\newblock \emph{Journal of the American Statistical Association}, \penalty0
  (just-accepted):\penalty0 1--20, 2025.

\bibitem[Gretton et~al.(2012)Gretton, Borgwardt, Rasch, Sch{\"o}lkopf, and
  Smola]{gretton2012kernel}
Arthur Gretton, Karsten~M Borgwardt, Malte~J Rasch, Bernhard Sch{\"o}lkopf, and
  Alexander Smola.
\newblock A kernel two-sample test.
\newblock \emph{The Journal of Machine Learning Research}, 13\penalty0
  (1):\penalty0 723--773, 2012.

\bibitem[Group(2009)]{juvenile2009effect}
Juvenile Diabetes Research Foundation Continuous Glucose Monitoring~Study
  Group.
\newblock The effect of continuous glucose monitoring in well-controlled type 1
  diabetes.
\newblock \emph{Diabetes care}, 32\penalty0 (8):\penalty0 1378--1383, 2009.

\bibitem[Group et~al.(2010)]{juvenile2010effectiveness}
Juvenile Diabetes Research Foundation Continuous Glucose Monitoring~Study Group
  et~al.
\newblock Effectiveness of continuous glucose monitoring in a clinical care
  environment: evidence from the juvenile diabetes research foundation
  continuous glucose monitoring (jdrf-cgm) trial.
\newblock \emph{Diabetes Care}, 33\penalty0 (1):\penalty0 17--22, 2010.

\bibitem[Haber and Ruthotto(2017)]{Haber_2018}
Eldad Haber and Lars Ruthotto.
\newblock Stable architectures for deep neural networks.
\newblock \emph{Inverse Problems}, 34\penalty0 (1):\penalty0 014004, dec 2017.
\newblock \doi{10.1088/1361-6420/aa9a90}.

\bibitem[Jain(2010)]{jain2010data}
Anil~K Jain.
\newblock Data clustering: 50 years beyond k-means.
\newblock \emph{Pattern recognition letters}, 31\penalty0 (8):\penalty0
  651--666, 2010.

\bibitem[Jia and Benson(2019)]{NEURIPS2019_njumpsdes}
Junteng Jia and Austin~R Benson.
\newblock Neural jump stochastic differential equations.
\newblock In H.~Wallach, H.~Larochelle, A.~Beygelzimer, F.~d\textquotesingle
  Alch\'{e}-Buc, E.~Fox, and R.~Garnett, editors, \emph{Advances in Neural
  Information Processing Systems}, volume~32. Curran Associates, Inc., 2019.

\bibitem[Katta et~al.(2024)Katta, Parikh, Rudin, and
  Volfovsky]{katta2024interpretable}
Srikar Katta, Harsh Parikh, Cynthia Rudin, and Alexander Volfovsky.
\newblock Interpretable causal inference for analyzing wearable, sensor, and
  distributional data.
\newblock In \emph{International Conference on Artificial Intelligence and
  Statistics}, pages 3340--3348. PMLR, 2024.

\bibitem[Kidger et~al.(2020)Kidger, Morrill, Foster, and
  Lyons]{kidger2020neural}
Patrick Kidger, James Morrill, James Foster, and Terry Lyons.
\newblock Neural controlled differential equations for irregular time series.
\newblock In H.~Larochelle, M.~Ranzato, R.~Hadsell, M.F. Balcan, and H.~Lin,
  editors, \emph{Advances in Neural Information Processing Systems}, volume~33,
  pages 6696--6707. Curran Associates, Inc., 2020.

\bibitem[Kosorok(2007)]{kosorok2007introduction}
Michael~R Kosorok.
\newblock \emph{Introduction to empirical processes and semiparametric
  inference}.
\newblock Springer Science \& Business Media, 2007.

\bibitem[Lugosi and Matabuena(2024)]{lugosi2024uncertainty}
G{\'a}bor Lugosi and Marcos Matabuena.
\newblock Uncertainty quantification in metric spaces.
\newblock \emph{arXiv preprint arXiv:2405.05110}, 2024.

\bibitem[Marzouk et~al.(2024)Marzouk, Ren, Wang, and Zech]{Marzouk2024}
Youssef Marzouk, Zhi Ren, Sven Wang, and Jakob Zech.
\newblock Distribution learning via neural differential equations: a
  nonparametric statistical perspective.
\newblock \emph{J. Mach. Learn. Res.}, 25\penalty0 (1), January 2024.
\newblock ISSN 1532-4435.

\bibitem[Massaroli et~al.(2020)Massaroli, Poli, Park, Yamashita, and
  Asama]{NEURIPS2020_massaroli}
Stefano Massaroli, Michael Poli, Jinkyoo Park, Atsushi Yamashita, and Hajime
  Asama.
\newblock Dissecting neural odes.
\newblock In H.~Larochelle, M.~Ranzato, R.~Hadsell, M.F. Balcan, and H.~Lin,
  editors, \emph{Advances in Neural Information Processing Systems}, volume~33,
  pages 3952--3963. Curran Associates, Inc., 2020.

\bibitem[Matabuena and Crainiceanu(2024)]{matabuena2024multilevel}
Marcos Matabuena and Ciprian~M Crainiceanu.
\newblock Multilevel functional distributional models with application to
  continuous glucose monitoring in diabetes clinical trials.
\newblock \emph{arXiv preprint arXiv:2403.10514}, 2024.

\bibitem[Matabuena and Petersen(2023)]{matabuena2023distributional}
Marcos Matabuena and Alexander Petersen.
\newblock Distributional data analysis of accelerometer data from the nhanes
  database using nonparametric survey regression models.
\newblock \emph{Journal of the Royal Statistical Society Series C: Applied
  Statistics}, 72\penalty0 (2):\penalty0 294--313, 2023.

\bibitem[Matabuena et~al.(2021{\natexlab{a}})Matabuena, Petersen, Vidal, and
  Gude]{matabuena2020glucodensities}
Marcos Matabuena, Alexander Petersen, Juan~C Vidal, and Francisco Gude.
\newblock Glucodensities: a new representation of glucose profiles using
  distributional data analysis.
\newblock \emph{Statistical methods in medical research}, 2021{\natexlab{a}}.

\bibitem[Matabuena et~al.(2021{\natexlab{b}})Matabuena, Petersen, Vidal, and
  Gude]{matabuena2021glucodensities}
Marcos Matabuena, Alexander Petersen, Juan~C Vidal, and Francisco Gude.
\newblock Glucodensities: A new representation of glucose profiles using
  distributional data analysis.
\newblock \emph{Statistical methods in medical research}, 30\penalty0
  (6):\penalty0 1445--1464, 2021{\natexlab{b}}.

\bibitem[Matabuena et~al.(2022)Matabuena, F{\'e}lix, Hammouri, Mota, and del
  Pozo~Cruz]{matabuena2022physical}
Marcos Matabuena, Paulo F{\'e}lix, Ziad Akram~Ali Hammouri, Jorge Mota, and
  Borja del Pozo~Cruz.
\newblock Physical activity phenotypes and mortality in older adults: a novel
  distributional data analysis of accelerometry in the nhanes.
\newblock \emph{Aging Clinical and Experimental Research}, 34\penalty0
  (12):\penalty0 3107--3114, 2022.

\bibitem[Matabuena et~al.(2024)Matabuena, Ghosal, Aguilar, Wagner, Merino,
  Castro, Zipunnikov, Onnela, and Gude]{matabuena2024glucodensity}
Marcos Matabuena, Rahul Ghosal, Javier~Enrique Aguilar, Robert Wagner,
  Carmen~Fern{\'a}ndez Merino, Juan~S{\'a}nchez Castro, Vadim Zipunnikov,
  Jukka-Pekka Onnela, and Francisco Gude.
\newblock Glucodensity functional profiles outperform traditional continuous
  glucose monitoring metrics.
\newblock \emph{arXiv preprint arXiv:2410.00912}, 2024.

\bibitem[Mesk{\'o} and Topol(2023)]{mesko2023imperative}
Bertalan Mesk{\'o} and Eric~J Topol.
\newblock The imperative for regulatory oversight of large language models (or
  generative ai) in healthcare.
\newblock \emph{NPJ digital medicine}, 6\penalty0 (1):\penalty0 120, 2023.

\bibitem[Muandet et~al.(2017)Muandet, Fukumizu, Sriperumbudur, Sch{\"o}lkopf,
  et~al.]{muandet2017kernel}
Krikamol Muandet, Kenji Fukumizu, Bharath Sriperumbudur, Bernhard
  Sch{\"o}lkopf, et~al.
\newblock Kernel mean embedding of distributions: A review and beyond.
\newblock \emph{Foundations and Trends{\textregistered} in Machine Learning},
  10\penalty0 (1-2):\penalty0 1--141, 2017.

\bibitem[null null(2008)]{doi:10.1056/NEJMoa0805017}
null null.
\newblock Continuous glucose monitoring and intensive treatment of type 1
  diabetes.
\newblock \emph{New England Journal of Medicine}, 359\penalty0 (14):\penalty0
  1464--1476, 2008.
\newblock \doi{10.1056/NEJMoa0805017}.
\newblock PMID: 18779236.

\bibitem[of~Us~Research Program~Investigators(2019)]{all2019all}
All of~Us~Research Program~Investigators.
\newblock The “all of us” research program.
\newblock \emph{New England Journal of Medicine}, 381\penalty0 (7):\penalty0
  668--676, 2019.

\bibitem[Papamakarios et~al.(2017)Papamakarios, Pavlakou, and
  Murray]{NIPS2017_6c1da886}
George Papamakarios, Theo Pavlakou, and Iain Murray.
\newblock Masked autoregressive flow for density estimation.
\newblock In \emph{Advances in Neural Information Processing Systems},
  volume~30. Curran Associates, Inc., 2017.

\bibitem[Papamakarios et~al.(2021)Papamakarios, Nalisnick, Rezende, Mohamed,
  and Lakshminarayanan]{papamakarios2021normalizing}
George Papamakarios, Eric Nalisnick, Danilo~Jimenez Rezende, Shakir Mohamed,
  and Balaji Lakshminarayanan.
\newblock Normalizing flows for probabilistic modeling and inference.
\newblock \emph{Journal of Machine Learning Research}, 22\penalty0
  (57):\penalty0 1--64, 2021.

\bibitem[Park et~al.(2025)Park, Kok, and Gaynanova]{park2025beyond}
Junyoung Park, Neo Kok, and Irina Gaynanova.
\newblock Beyond fixed thresholds: optimizing summaries of wearable device data
  via piecewise linearization of quantile functions.
\newblock \emph{arXiv preprint arXiv:2501.11777}, 2025.

\bibitem[Qian et~al.(2021)Qian, Zame, Fleuren, Elbers, and van~der
  Schaar]{qian2021integrating}
Zhaozhi Qian, William Zame, Lucas Fleuren, Paul Elbers, and Mihaela van~der
  Schaar.
\newblock Integrating expert odes into neural odes: pharmacology and disease
  progression.
\newblock \emph{Advances in Neural Information Processing Systems},
  34:\penalty0 11364--11383, 2021.

\bibitem[Rigby and Stasinopoulos(2005{\natexlab{a}})]{rigby2005generalized}
Robert~A Rigby and D~Mikis Stasinopoulos.
\newblock Generalized additive models for location, scale and shape.
\newblock \emph{Journal of the Royal Statistical Society Series C: Applied
  Statistics}, 54\penalty0 (3):\penalty0 507--554, 2005{\natexlab{a}}.

\bibitem[Rigby and Stasinopoulos(2005{\natexlab{b}})]{Rigby2005}
Robert~A Rigby and Dimitris~M Stasinopoulos.
\newblock Generalized additive models for location, scale and shape.
\newblock \emph{Journal of the Royal Statistical Society: Series C (Applied
  Statistics)}, 54\penalty0 (3):\penalty0 507--554, 2005{\natexlab{b}}.

\bibitem[Rubanova et~al.(2019)Rubanova, Chen, and
  Duvenaud]{NEURIPS2019_rubanova}
Yulia Rubanova, Ricky T.~Q. Chen, and David~K Duvenaud.
\newblock Latent ordinary differential equations for irregularly-sampled time
  series.
\newblock In \emph{Advances in Neural Information Processing Systems},
  volume~32. Curran Associates, Inc., 2019.

\bibitem[Sejdinovic et~al.(2013)Sejdinovic, Sriperumbudur, Gretton, and
  Fukumizu]{sejdinovic2013equivalence}
Dino Sejdinovic, Bharath Sriperumbudur, Arthur Gretton, and Kenji Fukumizu.
\newblock Equivalence of distance-based and {RKHS}-based statistics in
  hypothesis testing.
\newblock \emph{The Annals of Statistics}, pages 2263--2291, 2013.

\bibitem[Serfling(2009)]{serfling2009approximation}
Robert~J Serfling.
\newblock \emph{Approximation theorems of mathematical statistics}.
\newblock John Wiley \& Sons, 2009.

\bibitem[Silverman(1986)]{silverman1986density}
Bernard~W Silverman.
\newblock \emph{Density estimation for statistics and data analysis},
  volume~26.
\newblock CRC press, 1986.

\bibitem[Silverman(2018)]{silverman2018density}
Bernard~W Silverman.
\newblock \emph{Density estimation for statistics and data analysis}.
\newblock Routledge, 2018.

\bibitem[Sriperumbudur et~al.(2011)Sriperumbudur, Fukumizu, and
  Lanckriet]{sriperumbudur2011universality}
Bharath~K Sriperumbudur, Kenji Fukumizu, and Gert~RG Lanckriet.
\newblock Universality, characteristic kernels and rkhs embedding of measures.
\newblock \emph{Journal of Machine Learning Research}, 12\penalty0 (7), 2011.

\bibitem[Szab{\'o} et~al.(2016)Szab{\'o}, Sriperumbudur, P{\'o}czos, and
  Gretton]{szabo2016learning}
Zolt{\'a}n Szab{\'o}, Bharath~K Sriperumbudur, Barnab{\'a}s P{\'o}czos, and
  Arthur Gretton.
\newblock Learning theory for distribution regression.
\newblock \emph{Journal of Machine Learning Research}, 17\penalty0
  (152):\penalty0 1--40, 2016.

\bibitem[Tsybakov and Tsybakov(2009)]{tsybakov2009nonparametric}
Alexandre~B Tsybakov and Alexandre~B Tsybakov.
\newblock Nonparametric estimators.
\newblock \emph{Introduction to Nonparametric Estimation}, pages 1--76, 2009.

\bibitem[Tucker et~al.(2023)Tucker, Wu, and M{\"u}ller]{tucker2023variable}
Danielle~C Tucker, Yichao Wu, and Hans-Georg M{\"u}ller.
\newblock Variable selection for global fr{\'e}chet regression.
\newblock \emph{Journal of the American Statistical Association}, 118\penalty0
  (542):\penalty0 1023--1037, 2023.

\bibitem[Wang and et~al.(2024)]{wang2024timemixer}
S.~Wang and et~al.
\newblock Timemixer: Decomposable multiscale mixing for time series
  forecasting.
\newblock In \emph{ICLR}, 2024.

\bibitem[Wiener(1932)]{wiener1932}
Norbert Wiener.
\newblock Tauberian theorems.
\newblock \emph{Annals of Mathematics}, 33\penalty0 (1):\penalty0 1--100, 1932.
\newblock ISSN 0003486X, 19398980.

\bibitem[Wu and et~al.(2023)]{wu2023timesnet}
H.~Wu and et~al.
\newblock Timesnet: Temporal 2d-variation modeling for general time series
  analysis.
\newblock In \emph{ICLR}, 2023.

\end{thebibliography}

\end{document}